\pdfoutput=1

\documentclass[11pt]{article}

\usepackage[final]{acl}

\usepackage{times}
\usepackage{latexsym}
\usepackage{amsfonts}
\usepackage{amsmath}
\usepackage{stmaryrd}
\usepackage{bbm}
\usepackage{tcolorbox,listings}
\usepackage{multirow}
\usepackage{array}
\usepackage{booktabs}
\usepackage{mathtools}
\usepackage{graphicx}
\usepackage{caption}
\usepackage{subcaption}

\usepackage[T1]{fontenc}

\usepackage[utf8]{inputenc}

\usepackage{microtype}
\usepackage{cprotect}

\usepackage{inconsolata}
\usepackage{tikz}
\usetikzlibrary{positioning}  
\usetikzlibrary{shapes.geometric,fit}
\usetikzlibrary{trees}

\usepackage{graphicx}

\newcommand{\ConceptSimMath}{\mathord{\text{\footnotesize\texttt{ConceptSim}}}}

\DeclareMathOperator*{\argmax}{arg\,max}

\title{Unveiling Decision-Making in LLMs for Text Classification : Extraction of influential and interpretable concepts with Sparse Autoencoders}

\author{
 \textbf{Mathis Le Bail\textsuperscript{1}},
 \textbf{Jérémie Dentan\textsuperscript{1}},
 \textbf{Davide Buscaldi\textsuperscript{1,2}},
 \textbf{Sonia Vanier\textsuperscript{1}}
\\
\\
 \textsuperscript{1}LIX (École Polytechnique, IP Paris, CNRS)
\\
 \textsuperscript{2}LIPN (Sorbonne Paris Nord)
 \small{
   \textbf{Correspondence:} \href{mailto:mathis.le-bail@polytechnique.edu}{mathis.le-bail@polytechnique.edu}
 }
}

\begin{document}
\maketitle
\begin{abstract}
Sparse Autoencoders (SAEs) have been successfully used to probe Large Language Models (LLMs) and extract interpretable concepts from their internal representations. These concepts are linear combinations of neuron activations that correspond to human-interpretable features. In this paper, we investigate the effectiveness of SAE-based explainability approaches for sentence classification, a domain where such methods have not been extensively explored. We present a novel SAE-based model \verb|ClassifSAE| tailored for text classification, leveraging a specialized classifier head and incorporating an activation rate sparsity loss. We benchmark this architecture against established methods such as ConceptShap, Independent Component Analysis, HI-Concept and a standard TopK-SAE baseline. Our evaluation covers several classification benchmarks and backbone LLMs. We further enrich our analysis with two novel metrics for measuring the precision of concept-based explanations, using an external sentence encoder. Our empirical results show that  \verb|ClassifSAE| improves both the causality and interpretability of the extracted features.\footnote{See code at: \href{https://github.com/orailix/ClassifSAE}{https://github.com/orailix/ClassifSAE}}
\end{abstract}

\section{Introduction}

\begin{figure}[ht!]
    \centering

    \begin{subfigure}{0.48 \linewidth}
        \centering
        \includegraphics[width=\linewidth]{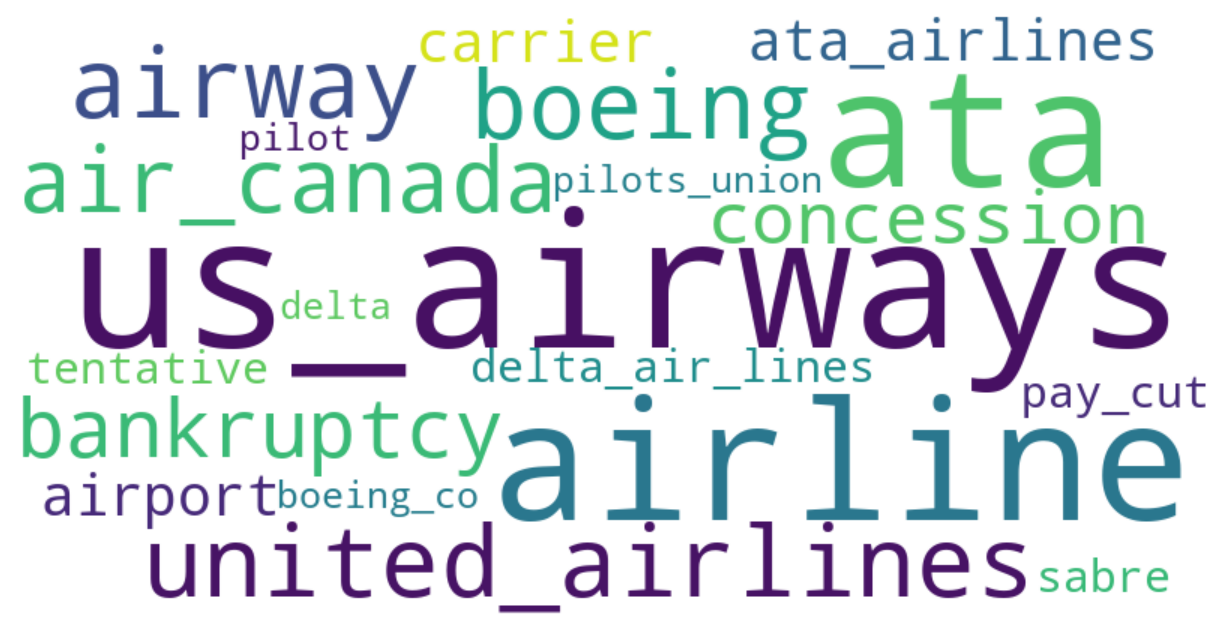}
        \caption{Business: Airlines}
        \label{fig:panel1}
    \end{subfigure}
    \hfill
    \begin{subfigure}{0.48\linewidth}
        \centering
        \includegraphics[width=\linewidth]{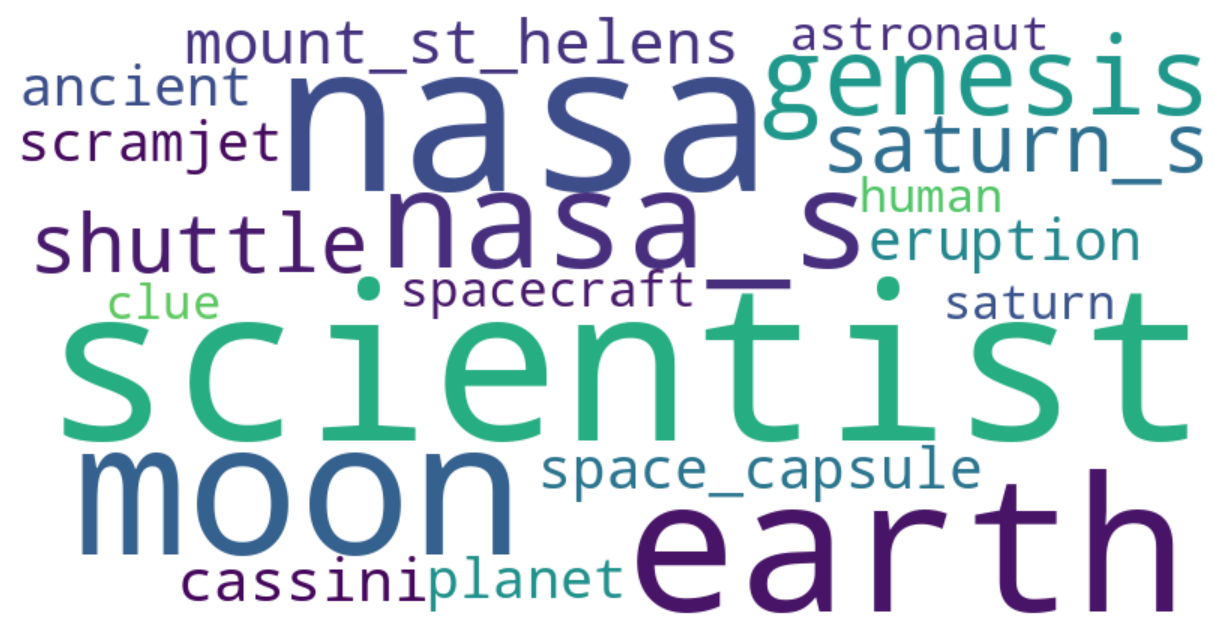}
        \caption{Sci/tech: Science–Nature }
        \label{fig:panel2}
    \end{subfigure}

    \vspace{0.7em} 

   \begin{subfigure}{0.48 \linewidth}
        \centering
        \includegraphics[width=\linewidth]{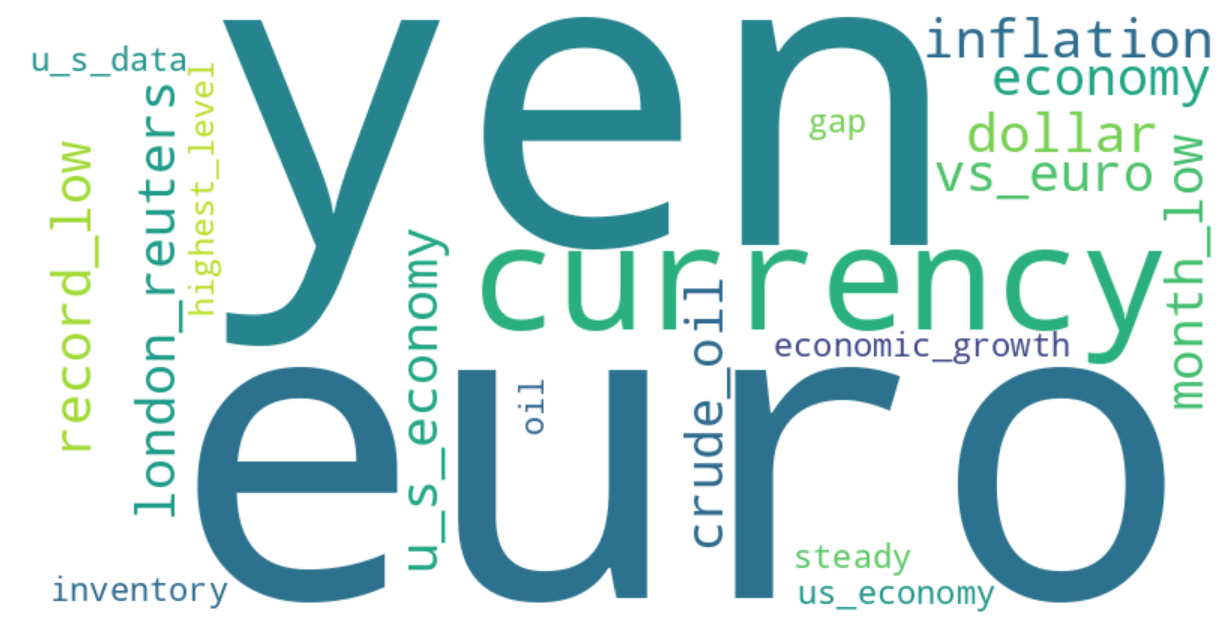}
        \caption{Business: Currency-Trade}
        \label{fig:panel3}
    \end{subfigure}
    \hfill
    \begin{subfigure}{0.48 \linewidth}
        \centering
        \includegraphics[width=\linewidth]{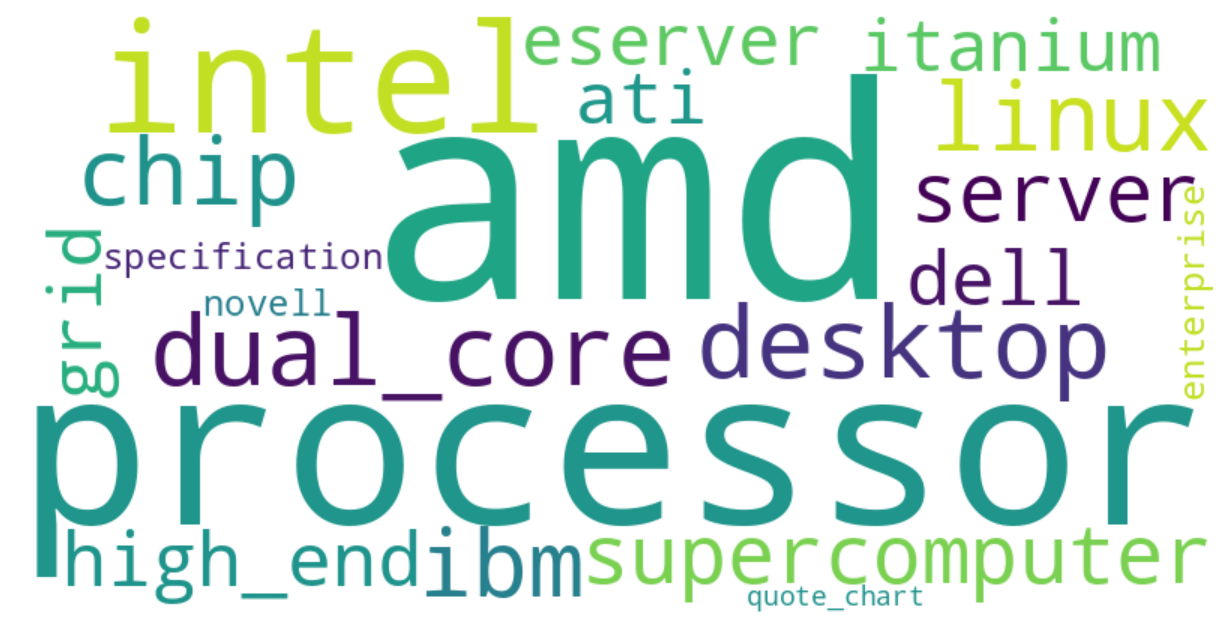}
        \caption{Sci/tech: Processors}
        \label{fig:panel4}
    \end{subfigure}

    \vspace{0.7em}

    \begin{subfigure}{0.48 \linewidth}
        \centering
        \includegraphics[width=\linewidth]{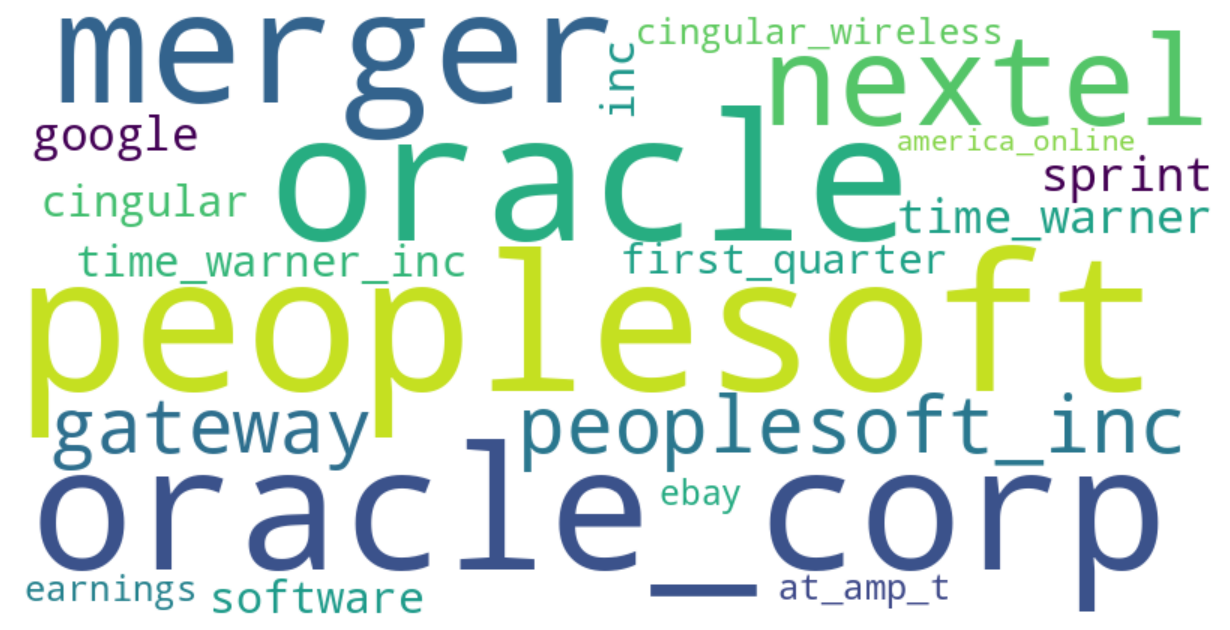}
        \caption{Business: \\ Tech–Corporations}
        \label{fig:panel5}
    \end{subfigure}
    \hfill
    \begin{subfigure}{0.48 \linewidth}
        \centering
        \includegraphics[width=\linewidth]{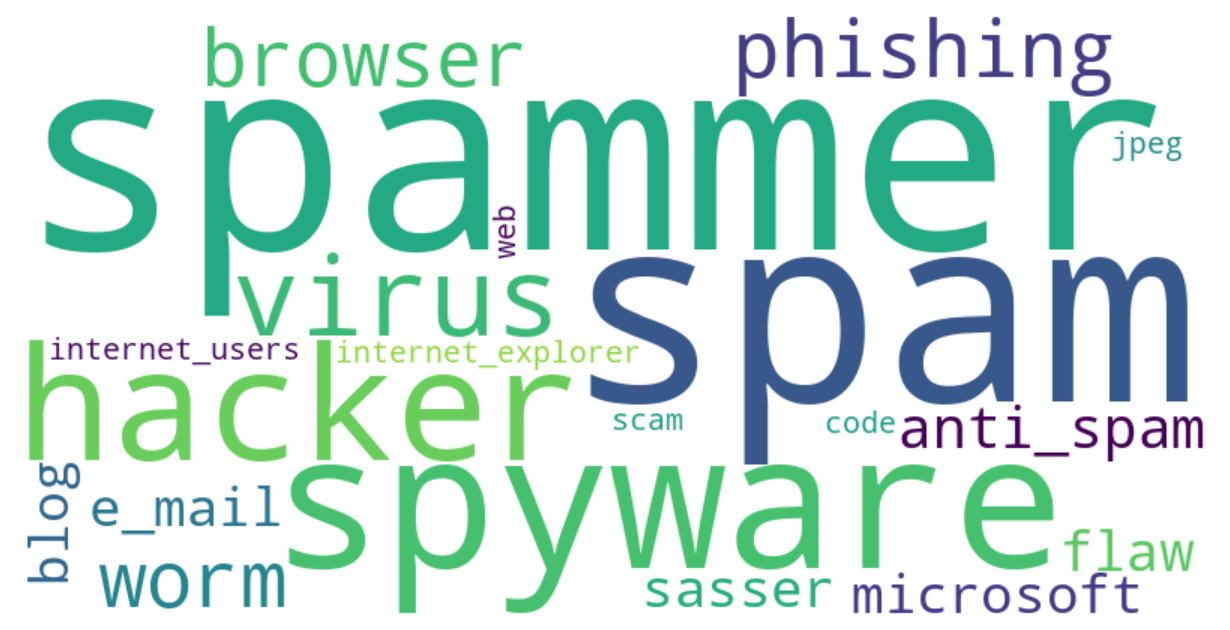}
        \caption{Sci/tech: \\ Cybersecurity-Spam}
        \label{fig:panel6}
    \end{subfigure}
    
    \cprotect\caption{Examples of concepts discovered by $\verb|ClassifSAE|$ from the internals of GPT-J fine-tuned on AG News.}
    \label{fig:fig_main}
\end{figure}

Text classification, similar to many other NLP tasks, has seen significant performance improvements with the adoption of Large Language Models (LLMs). However, compared to more self-explainable methods, the Transformer architecture used in LLMs does not readily reveal the specific concepts it actually leverages from the text to make a labeling decision. It relies on a high-dimensional latent space where vectors lack intuitive interpretability. Concept-based interpretability methods aim to extract high-level concepts from this space \cite{poeta2023conceptbasedexplainableartificialintelligence}. The latter are directions in the latent space that seek to align with human-understandable ideas, making them more meaningful than the latent vectors themselves.  In text classification, the labels may be too simplistic and various notions can result in the same decision. Therefore, extracting more nuanced concepts from the hidden states can provide deeper insights into the model’s intermediate decision-making process. 

To serve as a good proxy for the concepts learned by a model, the extracted directions should encompass three key properties. Effective concept vectors should be \textit{complete} and fully reflect the internal mechanism of the inspected neural network. This means that the original layer activations can be reliably reconstructed from the concept activations. Second, the explanations provided by the activated concepts need to be \textit{faithful} or \textit{causal} with respect to the model's final prediction \citep{lyu_etal_2024_towards}. The ablation of an identified concept vector must lead to significant variations in the inferred probabilities. Finally, the directions should align closely with well-defined and semantically meaningful human notions. This is measured by the \textit{precision} and \textit{recall} of the concepts. A high precision ensures that the activation of a direction is reflected by the presence of a well-defined concept within the input sentence. Conversely, the recall measures how reliably the ground-truth notion activates its associated direction when present in the tested sentence. Since recall depends on ground truth labels, which are unavailable for arbitrary text, we primarily focus on the precision of the extracted concepts.

Unsupervised approaches have gained in popularity to construct interpretable linear directions from LLMs latent space. Mechanistic Interpretability is a field of research that aims to automatically break down complex neural networks into simpler and interpretable parts to gain insights into the overall system \cite{explainability_survey,wang_interpretability_2022,gpt4explainsgpt2}. Recent contributions in this area built on the superposition linear representation hypothesis \cite{elhage2022toymodelssuperposition} to design scalable Sparse AutoEncoders (SAEs) for identifying meaningful directions within LLMs latent space without supervision \cite{Cunningham2023SparseAF}. They are mainly trained in the broad framework of autoregressive prediction, using as input a large number of token embeddings produced by the investigated LLM. A few studies have shown the practicality of SAE features extracted from pre-trained LLMs to obtain high accuracy scores on several text classification datasets \cite{gao_scaling_2024,DBLP:journals/corr/abs-2502-11367}. However, to the best of our knowledge, in the context of text classification by language model, no prior research has thoroughly compared SAE-extracted representations and  post-hoc methods from the field of concept discovery.

In this work, we address this question and we propose $\verb|ClassifSAE|$, a supervised variant of the SAE to reveal sentence-level features captured from LLMs' internal representations. Our goal is to identify features that both align with interpretable concepts and strongly influence classification outcomes. From a practical standpoint, we train our SAEs from scratch on the classification dataset, as in other concept-based reference methods. This differs from previous methods that pre-train SAEs on larger datasets and select few features that are relevant for the classification task. The method requires a limited number of sentence examples, on the order of 10,000 to 100,000. This allows any user with a model fine-tuned on a given classification task to quickly capture only the concepts related to that setting. Since the expansion dimension of the SAE is larger than the number of relevant concepts for most classification tasks, we enforce the concentration of key concepts within a subset of the hidden layer by training jointly the SAE and a classifier on that subset. Additionally, to prevent the collapse of diversity into a few active features, we design a sparsity mechanism that enables better control over the maximum activation rate of the concepts. We assess representation quality using proxy metrics for completeness, causality and interpretability.

\paragraph{Our paper makes the following contributions:}

\begin{itemize}
  \item We propose a new supervised SAE-based method, $\verb|ClassifSAE|$,  to extract fine-grained concepts learned by an LLM trained for sentence classification. 
  \item We introduce two novel metrics, $\verb|ConceptSim|$ and $\verb|SentenceSim|$, based on an external sentence encoder to assess the precision and interpretability property of the sentence-level concepts.
  \item We empirically compare our extracted concepts with four baselines: TopK-SAE, ICA \cite{COMON1994287}, ConceptSHAP \cite{NEURIPS2020_ecb287ff} and HI-Concept \cite{zhao-etal-2024-explaining}, across seven backbone LLMs, each fine-tuned on four distinct classification datasets.
  \item Using \verb|ClassifSAE|, we obtain sparser and more monosemantic concepts while achieving second-best causality scores and requiring up to 83\% less training time than HI-Concept, the state-of-the-art method, revealing a trade-off between causality and interpretability.
\end{itemize}

\section{Related Work}

\subsection{Concepts discovery in classification}  

Concept-based explanations provide a more robust understanding of an LLM's internal processes compared to the often unstable token attributions from gradient-based methods \cite{10.5555/3327546.3327621}. TCAV \cite{pmlr-v80-kim18d} was one of the first methods to identify activation-space vectors aligned with human-interpretable concepts and to quantify their influence on classification decisions. However, TCAV is supervised and requires example sets to define target concepts. Unsupervised approaches were later developed to automatically discover relevant concepts for explaining classification decisions. \citet{NEURIPS2020_ecb287ff} introduced a score to quantify the \textit{completeness} of a concept set in reconstructing the model’s original predictions. \citet{jourdan-etal-2023-cockatiel} proposed COCKATIEL, a method based on Non-Negative Matrix Factorization of the activation matrix, with the factorization rank controlling the number of concepts. Recently, \citet{zhao-etal-2024-explaining} introduced HI-Concept, an approach which emphasizes the causal impact of extracted concepts by training an MLP with a causal loss to reconstruct the classifier’s embedding space from the learned concepts. Notable progress has also been made in computer vision, where the discovery and hierarchical organization of interpretable concepts is particularly appealing due to their visual and intuitive nature \cite{ge_peek_2021,Fel_2023_CVPR,panousis:hal-04532065,wang_mcpnet_2024}.

\subsection{SAE-based concept extraction}

Early work in Mechanistic Interpretability aimed to associate interpretable concepts with individual neurons \cite{DBLP:conf/iclr/BauBSDDG19,10.1609/aaai.v33i01.33016309}. However, results were mixed due to \textit{polysemanticity}: single neurons often respond to multiple unrelated features, making interpretation ambiguous \cite{elhage2022toymodelssuperposition, gurnee_finding_2023}. Later studies showed that linear directions in latent space tend to be more interpretable than individual neuron activations \cite{Park2023TheLR,Marks2023TheGO,hollinsworth-etal-2024-language}. These approaches are supervised, requiring expert-defined concepts and example datasets, which is a limitation. This has motivated the development of unsupervised approaches to capture a broader and more diverse set of concepts. In particular, SAEs, inspired by the superposition hypothesis of linear representations \cite{elhage2022toymodelssuperposition}, have emerged as promising tools for disentangling superposed notions in the latent space of LLMs and for identifying interpretable directions \citep{Cunningham2023SparseAF}.

The SAE serves as a post-hoc technique to provide interpretability for a trained LLM. The large collection of token embeddings produced by the LLM serves as input to train the SAE. The columns of the decoder matrix are interpreted as concept directions, while the corresponding activations in the hidden layer represent the strength of each concept's presence in the input. For this reason, the concepts are also referred to as SAE \textit{features} or \textit{directions}. \citet{AnthropicSAE_Dictionary_learning} use an external LLM to generate explanations for each direction based on the text excerpts that most strongly activate each concept. Subsequent studies validate the interpretability of SAE-based directions in recent LLM architectures \citep{AnthropicSAE_Dictionary_learning_scale_claude,rajamanoharan2024improvingdictionarylearninggated,gao_scaling_2024}. While the original architecture enforces sparsity via an $|.|_1$ penalty, \citet{gao_scaling_2024} introduce a TopK activation function that enables direct control over $|.|_0$ sparsity by selecting a fixed number of active concepts per input. This leads to significant improvements in the sparsity–reconstruction trade-off and reduces the number of dead features at the end of training. Variants of TopK SAE have been proposed \cite{ayonrinde2024adaptivesparseallocationmutual,bussmann2024batchtopksparseautoencoders} to allow more dynamic allocation of feature capacity for tokens that are harder to reconstruct.

\section{Methodology}

\subsection{Preliminaries}\label{sec:problem_settings}

Let $f$ be a neural network trained for text classification. Similar to \citet{NEURIPS2020_ecb287ff,zhao-etal-2024-explaining}, we define concepts as vectors in $\mathbb{R}^p$. For a layer index $\ell$, let $\textbf{h}^\ell \in \mathbb{R}^d$ denote the residual-stream representation at that layer, which captures the sequence-level information used for prediction. We aim to approximate the hidden state $\textbf{h}^\ell$ using a combination of $m$ sparse concept vectors. We denote by $\hat{\textbf{h}}^{\ell}$ the resulting reconstruction of $\textbf{h}^\ell$ obtained from this sparse combination.

\subsection{Existing evaluation metrics} \label{sec:existing_metrics}

As noted in the Introduction, we evaluate three key properties of the extracted concepts: \textit{completeness}, \textit{causality} and \textit{interpretability}. For the first two, we rely on established metrics from the literature. For \textit{interpretability}, we introduce two new metrics, detailed in Section \ref{sec:new_metrics}.

\paragraph{Evaluating \textit{completeness}}

This property assesses whether the extracted concepts are sufficient to recover the model’s decisions. For concepts extracted from generative models, completeness is often measured via the reconstruction error of the original hidden state. In classification settings, however, not all directions in the hidden state necessary matter for prediction, as they can be discarded by subsequent layers if not useful for the classification decision \cite{hernandez_linearity_2023}. Therefore, we use recovery accuracy (RAcc), defined in Eq.~\ref{eq:racc}, as our completeness metric. RAcc is the proportion of inputs for which the model’s prediction remains unchanged when the original hidden state is replaced by its concept-based reconstruction. Formally, we decompose the prediction model into two blocks, $f = f_{\geq \ell} \circ f_{< \ell}$, where $\textbf{h}^{\ell}$ denotes the output of $f_{< \ell}$. Let $\mathcal{D}$ be the dataset for the sentence classification task. High RAcc suggests that the extracted concepts retain the classification-relevant information encoded at layer $\ell$, making them a reliable basis for interpreting the model’s decisions.

\begin{equation}
     \label{eq:racc}
     \text{RAcc} =\frac{1}{|\mathcal{D}|} \sum_{i \in \mathcal{D}} \mathbbm{1}
     \scalebox{0.9}{$
         \begin{bmatrix}
            \argmax \left(  f_{\geq \ell}\left( \textbf{h}_{i}^\ell \right)  \right) \\ 
            = \argmax \left(  f_{\geq \ell}\left( \hat{\textbf{h}}_{i}^\ell \right) \right)
        \end{bmatrix}
    $}
\end{equation}

\paragraph{Evaluating \textit{causality}}

This property measures the influence of the extracted concepts on the model’s prediction. Let ${\hat{\textbf{h}}} \backslash \{j\}$ denote the reconstructed hidden state obtained by ablating the $j$-th concept, by setting for instance its activation to a constant value such as $0$ across the dataset. Common metrics for evaluating the influence of concept $j$ include the shift in model accuracy, the label‑flip rate and the total variation distance (TVD) between the original and modified probability distributions measured after replacing $\hat{\textbf{h}}^\ell$ with $\hat{\textbf{h}}^\ell \backslash \{j\}$: 

\bigskip

\scalebox{0.8}{%
  \begin{minipage}{\linewidth}
    \begin{align}
        \Delta \text{Acc}_{\{j\}} &=
        \left| \text{Acc}\left( f_{\geq \ell}\left( \hat{\textbf{h}}^\ell \right) \right)
        - \text{Acc}\left( f_{\geq \ell}\left( \hat{\textbf{h}}^\ell \backslash \{j\} \right) \right) \right|
        \label{eq:causality_metrics1}\\[0.5em]
        \Delta f_{\{j\}} &=
        \mathbbm{1}\!\left[ \argmax  f_{\geq \ell}\left( \hat{\textbf{h}}^\ell \right)
        \neq \argmax f_{\geq \ell}\left( \hat{\textbf{h}}^\ell \backslash \{j\} \right)  \right]
        \label{eq:causality_metrics2}\\[0.5em]
        \text{TVD}_{\{j\}} &=
        \frac{1}{2} \left\| f_{\geq \ell}\left( \hat{\textbf{h}}^\ell \right)
        -  f_{\geq \ell}\left( \hat{\textbf{h}}^\ell \backslash \{j\} \right) \right\|_1
        \label{eq:causality_metrics3}
    \end{align}
  \end{minipage}%
}

\bigskip

Importantly, we distinguish the notions of \textit{global} and \textit{conditional} feature importance. Traditionally, these metrics are averaged across the entire dataset. However, this global averaging can undervalue sparse features, those that activate for only a small subset of inputs, even if they are highly influential when active. This distinction is drawn from the causal‑inference literature, where it is common to report both the Average Treatment Effect (ATE), the expected change in predictions if the feature were ablated for the entire population, and the Average Treatment Effect on the Treated (ATT), which quantifies the effect only among observations where the feature is present \cite{Morgan_Winship_2014}. Thus, we provide these metrics in the two configurations. $(\Delta Acc^{\text{global}}_{\{j\}}, \Delta f^{\text{global}}_{\{j\}} ,\text{TVD}^{\text{global}}_{\{j\}})$ denotes the metrics averaged over all evaluated sentences, while $(\Delta Acc^{\text{cond}}_{\{j\}},\Delta f^{\text{cond}}_{\{j\}}, \text{TVD}^{\text{cond}}_{\{j\}})$ only consider sentences which activate feature $j$.

\subsection{New metrics to evaluate \textit{Interpretability}} \label{sec:new_metrics}

Existing methods for assessing interpretability either rely on human evaluators, which is not reproducible, or LLM-as-a-judge, which is sensitive to prompting \cite{pmlr-v80-kim18d,gpt4explainsgpt2,gurnee2023finding}. We therefore introduce two new metrics: $\verb|ConceptSim|$ and $\verb|SentenceSim|$. The former measures how coherent a single concept’s meaning is across sentences, while the latter assesses how consistent meaning is between sentences sharing the same concepts. We evaluate the inspected LLM on a held-out test set and record concept activations per sentence. For each concept, this produces an activation vector over the test set, which we cluster in one dimension to distinguish activated from non-activated sentences. Let $\mathcal{S}^j$ be the set of sentences activating the $j$-th feature and $N^j$ its cardinal. We encode each sentence using Sentence-BERT \cite{reimers-gurevych-2019-sentence}. For $s_i \in \mathcal{S}^j$, let $e(s_i)$ be the sentence embedding. We evaluate the average pairwise similarity of activating sentences:

\begin{equation}
    \ConceptSimMath(j) = \dfrac{1}{\binom{N^j}{2}} \sum_ {\substack{i,i' \in \mathcal{S}^j \\ i \neq i'}}  \frac{e(s_i) \cdot e(s_{i'})}{|e(s_i)||e(s_{i'})|} 
    \label{eq:concept_sim}
\end{equation}

A higher value of $\verb|ConceptSim|$ indicates a better interpretability and monosemanticity in concept activations. If the sentences in $\mathcal{S}^j$ really share a common concept, their pairwise cosine similarity should be high. Cosine similarity has already been used by \citet{li-etal-2024-evaluating-readability} to assess concept interpretability, showing strong correlation with human annotations. The novelty of our metric lies in computing cosine similarity after performing a one-dimensional clustering to isolate the activating sentences in $\mathcal{S}^j$. Further implementation details are provided in Appendix \ref{sec:interpretability_metrics}, with an illustration of the pipeline in Figure \ref{fig:activating_sentences}. 

We also define $\verb|SentenceSim|$ to assess the similarity of sentences that share the same activating concepts. Each sentence $s_i$ is associated with the set of its $p$ most strongly activated concepts. We measure sentence similarity based on the number of shared concepts between these sets. For an integer $k$, let $\verb|SentenceSim|(s_i, k)$ denote the average cosine similarity between $e(s_i)$ and the embeddings of sentences whose concept sets share exactly $k$ elements with that of $s_i$. A higher overlap in activated features should align with closer sentences in the embedding space. To quantify this property, we average $\verb|SentenceSim|(s_i, k)$ over all sentences to obtain $\verb|SentenceSim|(k)$. We expect $\verb|SentenceSim|(k)$ to increase with $k$, since a higher $k$ implies greater overlap in top-$p$ concepts.

\subsection{Sparse AutoEncoders}

SAEs learn a higher-dimensional sparse representation of $\textbf{h}^\ell$. They consist of an encoder and a decoder that attempt to reconstruct the input embedding from this representation:

\begin{align}
    \textbf{z} = \sigma(W^{enc} \textbf{h} + b^{enc} ) \in \mathbb{R}^m \label{eq:test1} \\
    \hat{\textbf{h}} = W^{dec} \textbf{z}  + b^{dec} \in \mathbb{R}^d \label{eq:test2} 
\end{align}

where $\sigma$ is an activation function. The columns of $W^{dec} \in \mathbb{R}^{d \times m}$ can be understood as the directions of the concepts extracted in the embedding space. A popular choice for $\sigma$ is the TopK activation function as it enables to fix the number of non-zeros values in $\textbf{z}$ to $k$ per input, thereby always having the same number of allocated features per embedding to reconstruct. The training loss for TopK SAEs is often presented as: 

\begin{equation}
    \mathcal{L}^\text{TopK}_\text{SAE}(\textbf{h}) = \frac{\| \textbf{h} - \hat{\textbf{h}} \|_2^2 }{\| \textbf{h} - \textbf{h}_{\text{batch mean}} \|_2^2 } + \alpha \mathcal{L}_{\text{aux}}(\textbf{h},\hat{\textbf{h}},\textbf{z})
\end{equation}

where $\alpha > 0$. It is trained to minimize the reconstruction error between $\textbf{h}$ and $\hat{\textbf{h}}$. An auxiliary loss $\mathcal{L}_{\text{aux}}$ can be combined to mitigate the premature emergence of dead features. These are defined as the coefficients in $\textbf{z}$ that no longer activate for any input beyond a certain training threshold. A high proportion of dead features hinders the SAE representational capacity. A popular choice for $\mathcal{L}_{\text{aux}}$ is the auxiliary loss introduced in \citet{gao_scaling_2024} inspired by the "ghost grads" method \cite{ghost_grads}.

\subsection{ClassifSAE for text classification}\label{sec:our_method}

\begin{figure}[t!] 
  \centering
  \includegraphics[%
    clip,                     
    trim=80mm 20mm 30mm 100mm,   
    width=\linewidth         
  ]{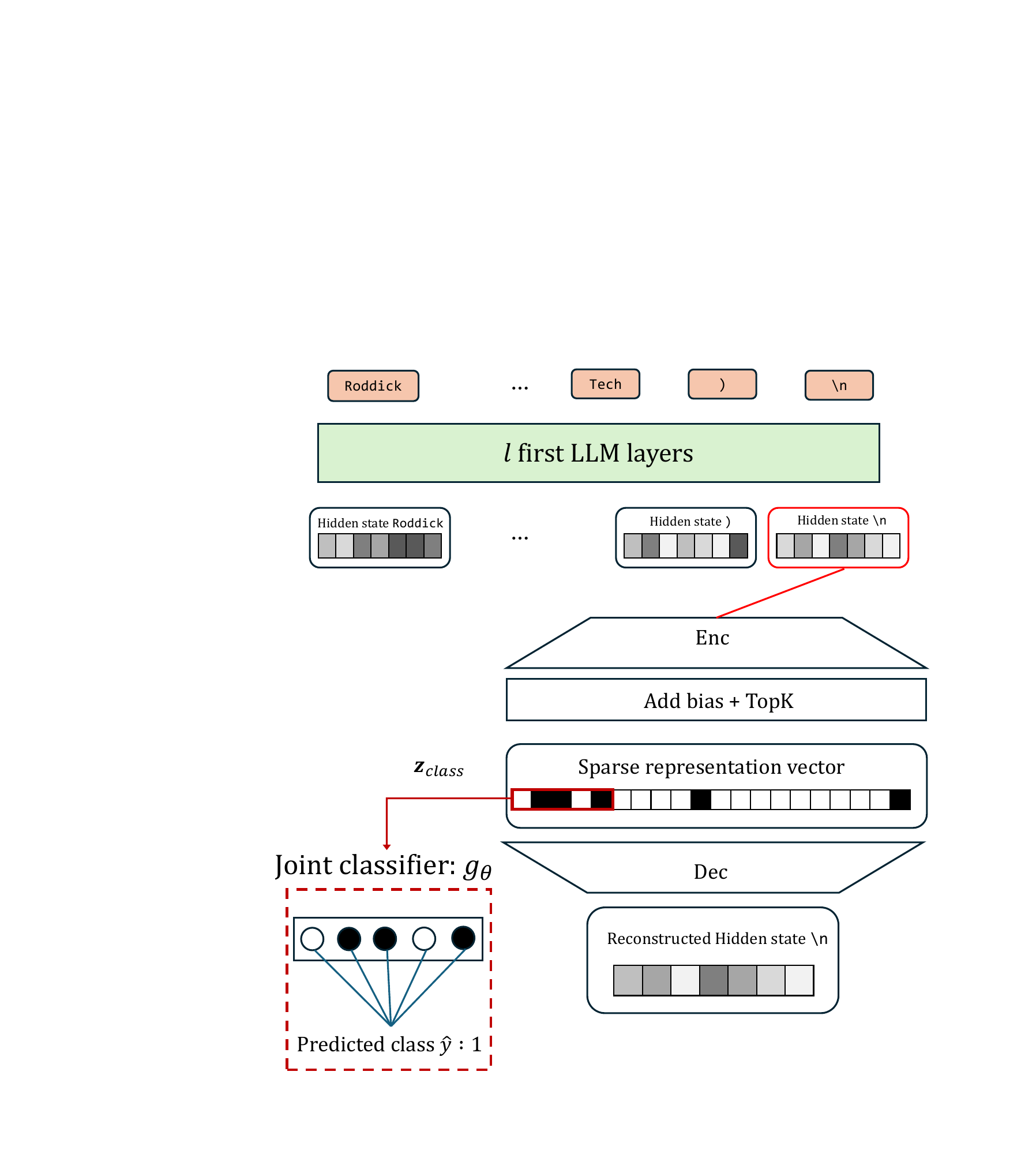}
  \cprotect\caption{Architecture of our $\verb|ClassifSAE|$ model. A classifier is trained jointly with the SAE to replicate the original LLM prediction. A low dimensionality of $\textbf{z}_{\text{class}}$ incentivizes the model to extract a small number of distinct task-relevant features.}
  \label{fig:architecture_sae_variant}
\end{figure}

We now introduce $\verb|ClassifSAE|$, our adaptation of TopK SAE for sentence classification. In generative tasks, the SAE is typically trained on a large and diverse dataset to capture the full range of \mbox{representations} encoded by the model. In contrast, for text classification, we assume that a thematically focused dataset of moderate size is used to fine-tune the model. We train the SAE on this dataset to uncover only the concepts that are relevant for the classification. To capture sentence-level features, we only train the SAE on the hidden state $\textbf{h}^\ell$ of a single token in the input sentence. For the autoregressive language models, it is the token preceding the label. For encoder‑only models like BERT, it would be the $\verb|[CLS]|$ token. This token indeed serves as an aggregate representation of the input sentence. As the model progresses through its layers, it consolidates sentence-level information into this token, whose final representation is used for classification.

\paragraph{Training a joint classifier} We perform a forward pass of the LLM on the sentence dataset $\mathcal{D} = \{s_i\}_{i=1}^N$. This produces $X = \{(\textbf{h}^{\ell}_i,\hat{y}_i)\}_{i=1}^N$, where $\hat{y}_i$ is the LLM’s predicted label for sentence $i$. These pairs serve as inputs to jointly train the SAE and a classifier $g_{\theta}$ (see Figure \ref{fig:architecture_sae_variant}). The classifier $g_{\theta}$ is trained to reproduce the LLM’s decisions using only a subset of the SAE features $\textbf{z}_{\text{class}} \subset \textbf{z}$. Consequently, the SAE is incentivized to cluster task-relevant features in $\textbf{z}_{\text{class}}$, while the remaining features support the reconstruction. We draw inspiration from \citet{DBLP:conf/cvpr/DingXXPYWT20}, this joint training enables end-to-end generation and selection of task-relevant features, while preserving the model’s capacity to encode task-irrelevant concepts in a distinct subspace. The classifier is trained with cross-entropy (CE) loss:

\begin{equation}
    \mathcal{L}_{\text{class}} = \text{CE}(g_{\theta}(\textbf{z}_{\text{class}}),\hat{y})
\end{equation}

\paragraph{Feature sparsity} 

We observed that some features of the SAE are triggered by a very large portion of the dataset. Moreover, these highly-activated features exhibit a significant degree of correlation, which is not relevant for discovering diverse concepts. To alleviate these issues, we introduced an activation rate sparsity mechanism in the training phase. We select a hyperparameter $\gamma$ that we consider as the targeted maximum activation rate for an individual feature. For a given batch of size $B$, we denote $\mathcal{I} = \llbracket 1,B \rrbracket$ and $T = \lfloor \gamma B \rfloor$. We define:

\begin{equation}
    \mathcal{I}_j =  \argmax_{\mathcal{I}' \subseteq \mathcal{I}, |\mathcal{I}'| = T} \sum_{i \in \mathcal{I}'} |z^j_i| \quad ,\forall j \in \llbracket 1,m \rrbracket
\end{equation} 

\noindent where $z_i^j$ stands for the activation value of the $j\text{-th}$ concept for input $\textbf{h}_i$. For each feature $j \in \llbracket 1,m \rrbracket$, $\mathcal{I}_j$ contains the indices of the top $T$ inputs that most strongly activate direction $j$. Features with non-zero activations in more than $T$ sentences of the batch are penalized with the following loss term:  

\begin{equation}
    \mathcal{L}_{\substack{\text{sparse} \\ \text{feature}}} = \sum_{j=1}^m \sum_{i' \notin \mathcal{I}_j} |z_{i'}^j|
\end{equation}

This incentivizes a more balanced distribution of retained information across dimensions of the hidden layer, while promoting features that are more discriminative across sentences. We enforce  sparsity in the activation rates of the learned directions via a penalty and not through zeroing out every exceeding activations to account for randomness in the distribution of the batch. The final training loss of $\verb|ClassifSAE|$ is then:

\begin{equation}
    \begin{aligned}
         \mathcal{L} = \lambda_1 \mathcal{L}^\text{TopK}_\text{SAE} + \lambda_2 \mathcal{L}_{\text{class}} + \lambda_3 \mathcal{L}_{\substack{\text{sparse} \\ \text{feature}}}  
    \end{aligned}
\end{equation}

\paragraph{Evaluation} We retain only the task-relevant features $\textbf{z}_{\text{class}}$ as final concepts. They are evaluated using metrics from Sections \ref{sec:existing_metrics} and \ref{sec:new_metrics}. For fair comparison with other methods,  we zero out \mbox{all $z$} not in $\textbf{z}_{\text{class}}$ during evaluation. The dimensionality of $\textbf{z}_{\text{class}}$ serves as a hyperparameter controlling the number of extracted concepts. Finally, each feature is assigned the class on which it exhibits the highest average activation. See Appendix~\ref{sec:feature_segmentation} for details.

\section{Experiments and Results}

We compare $\verb|ClassifSAE|$ to four concept discovery methods: ICA \cite{COMON1994287}, TopK SAE, ConceptShap (\citet{NEURIPS2020_ecb287ff}) and Hi-Concept \cite{zhao-etal-2024-explaining}, all trained on cached internal embeddings. The latter two and $\verb|ClassifSAE|$ additionally use supervision from LLM predicted labels or logits. Model descriptions and implementation details are in Appendices \ref{sec:baselines} and \ref{sec:methods_implementation_details}. We consider four text‑classification tasks: AG News \cite{10.5555/2969239.2969312}, IMDB \cite{maas-etal-2011-learning}, an offensive language identification dataset \cite{zampieri2019semeval} and a sentiment analysis dataset \cite{rosenthal2017semeval}, both from the TweetEval benchmark \cite{barbieri2020tweeteval}. TweetEval’s tasks are more challenging, with fuzzier class boundaries and skewed label distributions. We conduct experiments with seven backbone LLMs (two encoder‑only and five decoder‑only representatives). For the largest models, alignment with the classification task is performed via soft-prompt tuning rather than full fine-tuning. Dataset and training details are in Appendices \ref{sec:dataset_details} and \ref{sec:classifier_details}. In our comparative study, concepts are extracted from the penultimate transformer block in decoder-only architectures and from the final encoder layer before the classification head in encoder-only ones, capturing high-level and sentence-aware representations. Appendix \ref{sec:depth_effect} provides an example of a more targeted analysis of layer‑depth effects on concepts extraction with $\verb|ClassifSAE|$. For fairness, all methods extract $20$ concepts per configuration. Latent variables in $\textbf{z}_{\text{class}}$ with near-zero mean activation are discarded. We set $\gamma=0.1$ for sparsity and $K=10$ for TopK.

\subsection{Numerical results}

Our results are compiled in Table \ref{tab:table_metrics_compiled}, they are averaged over the 4 datasets. Individual results are reported in Appendix Tables \ref{tab:table_metrics_agnews},\ref{tab:table_metrics_te_offensive}, \ref{tab:table_metrics_te_sentiment} and \ref{tab:table_metrics_imdb}. Since $\verb|ConceptSim|$ varies significantly across datasets, we standardize the weighted average $\verb|ConceptSim|$ metric using the mean and variance of pairwise sentence-embedding cosine similarities within each dataset. On average, all methods maintain acceptable recovery accuracy RAcc (Eq. \ref{eq:racc}), ensuring the completeness of the learned concepts. 
\paragraph{Concepts Interpretability} $\verb|ClassifSAE|$ is capable of engineering features that are more interpretable according to the $\verb|ConceptSim|$ (Eq.~\ref{eq:concept_sim}) metric. The features computed by $\verb|ClassifSAE|$ consistently exhibit higher $\verb|ConceptSim|$ scores and better activation rate sparsity across all evaluated model–task pairs. $\verb|ClassifSAE|$ architecture builds on the interpretability of the SAE and further enhances it. Because the training datasets are relatively small, the SAE revisits sentences multiple times to improve reconstruction, causing certain features to activate across a large portion of the evaluation set. This behavior is mitigated by incorporating the activation rate sparsity loss into $\verb|ClassifSAE|$, improving sparsity and monosemanticity in the discovered concepts.

Figure \ref{fig:sentence_sim_gptj} and \ref{fig:sentence_sim_pythia} in Appendix show $\verb|SentenceSim|$ for the GPT-J and Pythia-1B models fine-tuned on each classification task. Sentences that share identical top-activating concepts exhibit higher similarity in the sentence embedding space when using concepts from $\verb|ClassifSAE|$ for mapping, compared to other baselines. This highlights the improved interpretability of $\verb|ClassifSAE|$'s directions.

\paragraph{Concepts Causality} We measure causality in the \textit{conditional} sense (see Section \ref{sec:existing_metrics}) so as not to artificially disadvantage sparse features. Although we define three causality metrics ($\Delta Acc$, $\Delta f$,TVD) Eqs.~\ref{eq:causality_metrics1}--\ref{eq:causality_metrics3}, we report only $\Delta f$ in Table~\ref{tab:table_metrics_compiled}, as the three are highly correlated across models and datasets and lead to the same ranking of methods. HI-Concept achieves state‑of‑the‑art performance on this metric, which aligns with expectations since its training objective explicitly includes a surrogate for this measure. However, this addition comes at the cost of reduced interpretability: HI-Concept obtains lower average \verb|ConceptSim| scores than ConceptSHAP across all models. $\verb|ClassifSAE|$ still achieves the second-best $\Delta f^{\text{cond}}$ score and remains on par with HI-Concept on the largest models. Compared to standard SAE, the joint classifier improves disentanglement by promoting $\mathbf{z}_{\text{class}}$ to specialize in task-relevant concepts. Operating under a sparsity constraint and low-dimensional input, the classifier must maximize its representational capacity using a limited set of active features per sentence. This drives the selection of less correlated features in $\mathbf{z}_{\text{class}}$, improving information capture and ultimately leading to better $\Delta f^{\text{cond}}$ metrics.

\paragraph{Computational Efficiency} Table~\ref{tab:table_durations} reports the training times of \verb|ClassifSAE| and the two most competitive baseline methods, HI-Concept and ConceptSHAP, on the AG News dataset. We use the same parameter settings as in the main comparison experiments. The post-hoc interpretability methods are branched at the output of the penultimate layer of the language model. The results show that \verb|ClassifSAE| is substantially more computationally efficient, requiring up to $83\%$ and $69\%$ less training time than the baselines for the largest model (LLaMA~3.1~8B Instruct). This efficiency gain stems from the baselines’ reliance on layers of the LLM subsequent to the one under investigation to compute reconstruction accuracy and, in the case of HI-Concept, to estimate a proxy for concept causality. As a result, their computational cost increases with both the number of downstream layers and the need to repeatedly load and process parts of the LLM. In contrast, \verb|ClassifSAE| trains only an autoencoder coupled with a lightweight classifier and does not require loading any part of the LLM during training. It learns a proxy of the model’s behavior from the predicted labels via the joint classifier. Its cost scales only with the dimensionality of the LLM’s residual stream. This computation gap further widens when concepts are extracted from earlier layers, where the baselines must process an even larger portion of the remaining model.

\begin{figure}[tbp]
  \centering
  \includegraphics[width=\linewidth,height=0.8\columnwidth]{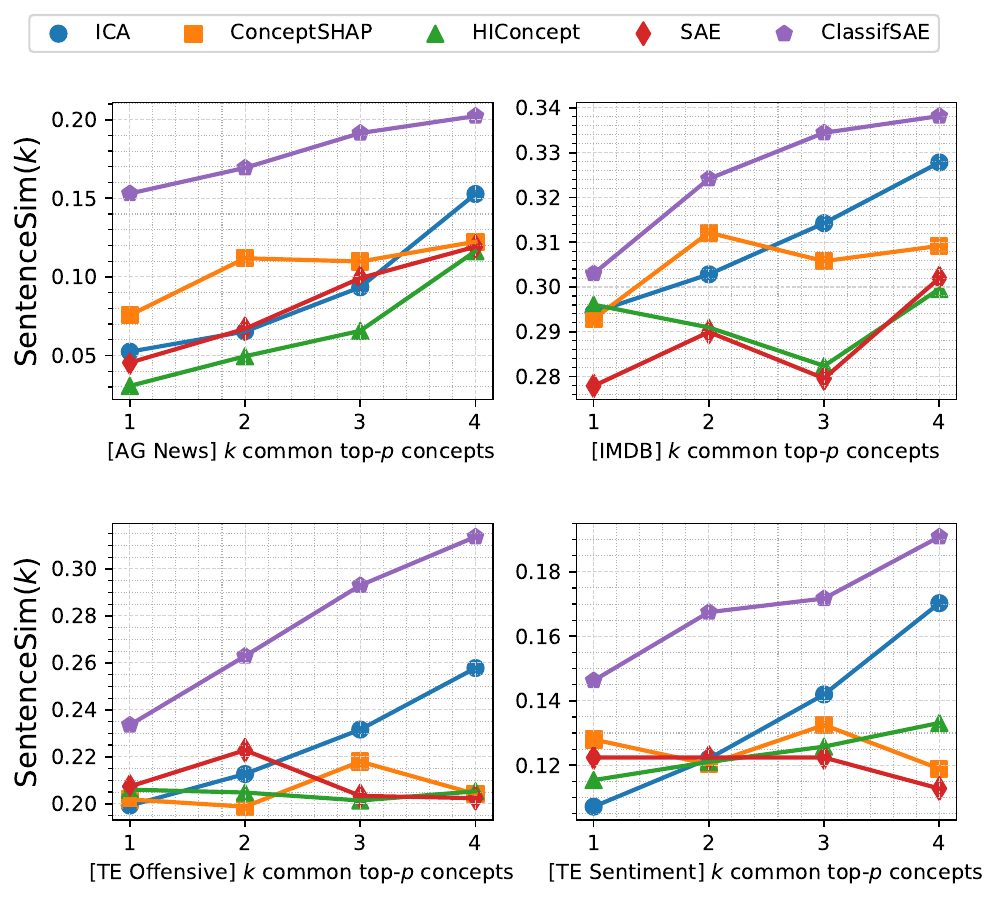}
  \cprotect\caption{$\verb|SentenceSim|(k)$ as a function of the number $k$ of shared top-activating concepts between sentence pairs. Concepts are learned from sentence-level hidden states in the penultimate transformer block of \mbox{GPT-J} fine-tuned on one of the four classification tasks. We consider $p=5$ concepts for each sentence.}
  \label{fig:sentence_sim_gptj}
\end{figure}

\begin{table*}
\centering
\scriptsize                          
\setlength{\tabcolsep}{2.8pt}        
\renewcommand{\arraystretch}{1.5}
\begin{tabular}{l rrrrrrr}
\hline\hline
\multicolumn{1}{c}{ } &
\multicolumn{7}{c}{\textbf{Classifier Model}}\\
\cline{1-8}
\multicolumn{1}{c}{} & 
BERT(110M) & DeBERTa‑v3(304M) & Pythia(410M) & Pythia(1B) & GPT‑J(6B) & Mistral‑Inst.(7B) & Llama‑Inst.(8B)  \\
\textbf{Method} &
\multicolumn{7}{c}{}\\
\hline

ConceptSHAP &  9 min 10 s & 9 min 50 s & 19 min 14 s & 23 min 39 s & 32 min 47 s & 28 min 53 s & 45 min 38 s\\
    \cline{1-8}
HI-Concept & 10 min 56 s & 11 min 50 s & 26 min 10 s & 30 min 30 s & 55 min 41 s & 49 min 52 s & 1 h 23 min 47 s \\
    \cline{1-8}
 $\verb|ClassifSAE|$ (K=10) &  2 min 34 s & 2 min 54 s & 2 min 53 s & 5 min 01 s & 13 min 47 s & 14 min 01 s & 14 min 08 s\\
        \hline\hline

\end{tabular}
\cprotect\caption{Training time comparison between \verb|ClassifSAE| and the two most competitive approaches (HI-Concept, ConceptSHAP) on the dataset AG News. All experiments were conducted using the same NVIDIA A100 GPU.}
\label{tab:table_durations}
\end{table*}

\begin{table*}[t!]
\centering
\scriptsize                          
\setlength{\tabcolsep}{2.8pt}        
\renewcommand{\arraystretch}{1.5}
\begin{tabular}{ll rrrrrrr}
\hline\hline
\multicolumn{2}{c}{ } &
\multicolumn{7}{c}{\textbf{Classifier Model}}\\
\cline{2-9}
\multicolumn{2}{c}{} & 
BERT(110M) & DeBERTa‑v3(304M) & Pythia(410M) & Pythia(1B) & GPT‑J(6B) & Mistral‑Inst.(7B) & Llama‑Inst.(8B)  \\
\textbf{Method} & \textbf{Metric} &
\multicolumn{7}{c}{}\\
\hline

ICA & RAcc (\%,$\uparrow$) & \textbf{99.52} & 99.80 & 99.29 &99.14& \textbf{99.50} & 96.74 & 93.57\\
 & $\Delta f^{\text{cond}}$ (\%,$\uparrow$) & 2.48 & 2.42 & 2.91 & 3.62 & 3.08 & 3.54 & 3.74\\
& Std $\verb|ConceptSim|$ ($\uparrow$)  & 0.0137 & 0.0073 & 0.0175 & 0.0169 & 0.0119 & 0.0561 & 0.0443\\
& Avg Act. rate (\%,$\downarrow$) & 100 & 100 & 100 & 100 & 100 & 100 & 100\\
        \cline{2-9}
ConceptSHAP & RAcc (\%,$\uparrow$) & 96.50 & 98.03 & 95.98 & 94.05 & 96.14 & 93.50 & 89.16\\
& $\Delta f^{\text{cond}}$ (\%,$\uparrow$)  & 6.63  & 0.53 & 4.01 & 3.94 & 2.89 & 3.47 & 2.38 \\
& Std $\verb|ConceptSim|$ ($\uparrow$)  & 0.2455 & 0.2135 & 0.1948 & 0.2703 & 0.2631 & 0.3694 & \textbf{0.3895}\\
& Avg Act. rate (\%,$\downarrow$) & 37.59 & 36.70 & 37.17 &  31.67 & 32.42 & 22.52 & 25.12\\
    \cline{2-9}
HI-Concept & RAcc (\%,$\uparrow$) & 99.33 & \textbf{99.66} & \textbf{99.40} & \textbf{99.47}  & 99.44 & \textbf{98.98} & \textbf{98.26}\\
& $\Delta f^{\text{cond}}$ (\%,$\uparrow$)  & \textbf{29.66} & \textbf{33.83} & \textbf{16.81} & \textbf{16.78} & \textbf{26.19} & \textbf{16.36}  & 7.30\\
& Std $\verb|ConceptSim|$ ($\uparrow$)  & 0.1799 & 0.2030 & 0.1410 & 0.1089 & 0.0624 & 0.0688 & 0.1030\\
& Avg Act. rate (\%,$\downarrow$) & 39.16 & 41.19 & 46.77 5& 44.38 & 59.11 & 61.83 & 61.33\\
    \cline{2-9}
SAE (K=10) & RAcc (\%,$\uparrow$)  & 97.39 & 98.47 & 95.54 & 98.08 &94.46 & 91.27 & 88.90\\
& $\Delta f^{\text{cond}}$ (\%,$\uparrow$)  & 2.04  & 3.43 & 1.98  & 4.28 & 8.97 & 11.60 & 16.03\\
& Std $\verb|ConceptSim|$ ($\uparrow$)  & 0.3138 & 0.2590 & 0.2742 & 0.2207 & 0.1498 & 0.0578 & 0.065\\
& Avg  Act. rate (\%,$\downarrow$) & 20.39 & 29.19 & 18.40 & 20.78 & 35.49 &  44.60 & 41.90 \\
    \cline{2-9}
 $\verb|ClassifSAE|$ (K=10) & RAcc (\%,$\uparrow$) &  97.745 & 98.35 & 96.99 & 95.94 & 97.06 & 95.93 & 96.35\\
& $\Delta f^{\text{cond}}$ (\%,$\uparrow$)  & 9.86 & 13.96 & 14.48 & 14.92 & 12.03 & 11.30 & \textbf{17.56}\\ 
& Std $\verb|ConceptSim|$ ($\uparrow$)  & \textbf{0.4512} & \textbf{0.3521} & \textbf{0.3781} & \textbf{0.3994} & \textbf{0.4657} & \textbf{0.4196} & 0.3728 \\
& Avg  Act. rate (\%,$\downarrow$) & \textbf{9.53} & \textbf{9.38}  & \textbf{8.10} & \textbf{8.48} & \textbf{9.66} & \textbf{9.85} & \textbf{10.13}\\
        \hline\hline

\end{tabular}
\cprotect\caption{Completeness, causality and interpretability metrics (see Sections \ref{sec:existing_metrics} and \ref{sec:new_metrics}) of the concepts learned from different LLM classifiers ($\uparrow$ : higher is better, $\downarrow$ : lower is better). The metrics are averaged over $4$ classification datasets. Prior to each evaluation, all models are fine-tuned except Mistral-Instruct and Llama-Instruct, which are aligned to the task via soft-prompt tuning (PT). $\Delta f^{cond}$ (Eq.~\ref{eq:causality_metrics2}) is the mean of $\Delta f_{\{j\}}^{cond}$. $\verb|ConceptSim|$ (Eq.~\ref{eq:concept_sim}) is the weighted average of individual concept scores (Appendix~\ref{sec:interpretability_metrics}). Std $\verb|ConceptSim|$ denotes its standardized version. All post-hoc methods search for 20 concepts. Concepts are computed from the sentence-level hidden state, extracted for decoder-only models from the residual stream after the penultimate block, and for encoder-only models from the layer preceding the classification head.}
\label{tab:table_metrics_compiled}
\end{table*}

\subsection{Ablation studies}\label{sec:ablation_studies_section}

To evaluate the impact of the two newly added components in $\verb|ClassifSAE|$ and the hidden-layer size $d_{sae}$, we measured the completeness, causality and interpretability of the learned concepts across different training settings. Results are shown in Figure~\ref{fig:ablation_studies} in the Appendix. The SAE with both components deactivated shows the weakest $\verb|ConceptSim|$ and $\text{TVD}^{\text{cond}}$ scores, irrespective of $d_{sae}$. The addition of the classifier for automatic variable selection significantly increases $\text{TVD}^{\text{cond}}$. Enabling the activation rate sparsity mechanism improves $\verb|ConceptSim|$ across all tested $d_{sae}$.  Applying both strategies jointly gives the best trade-off between the evaluation metrics. While the learned concepts remain competitive for $d_{sae} \in \{512,2048,4096\}$ (expansion factor of 0.25, 1 and 2), $\text{TVD}^{\text{cond}}$ shows a marked improvement with larger SAE hidden layer dimension when the joint classifier is enabled.

\begin{figure}[ht!]
  \centering
  \includegraphics[width=\columnwidth]{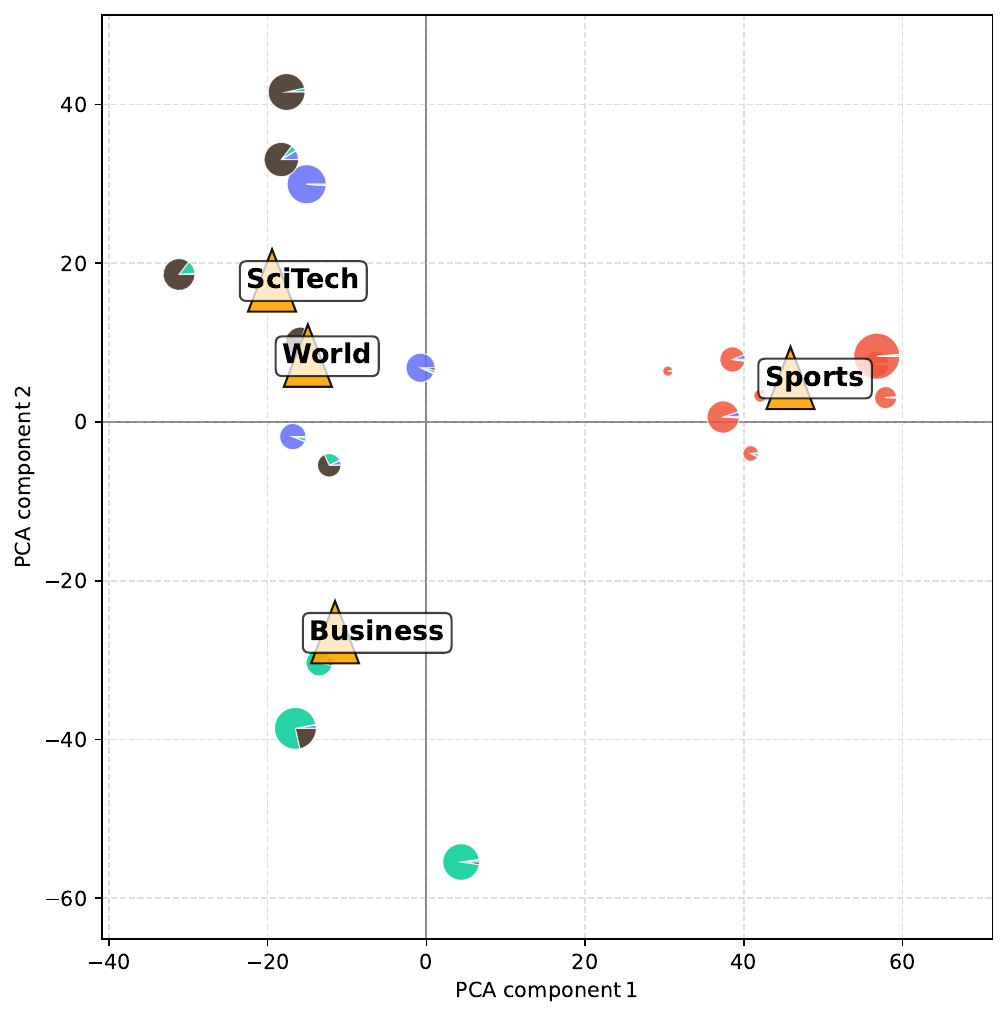}
  \includegraphics[width=\columnwidth]{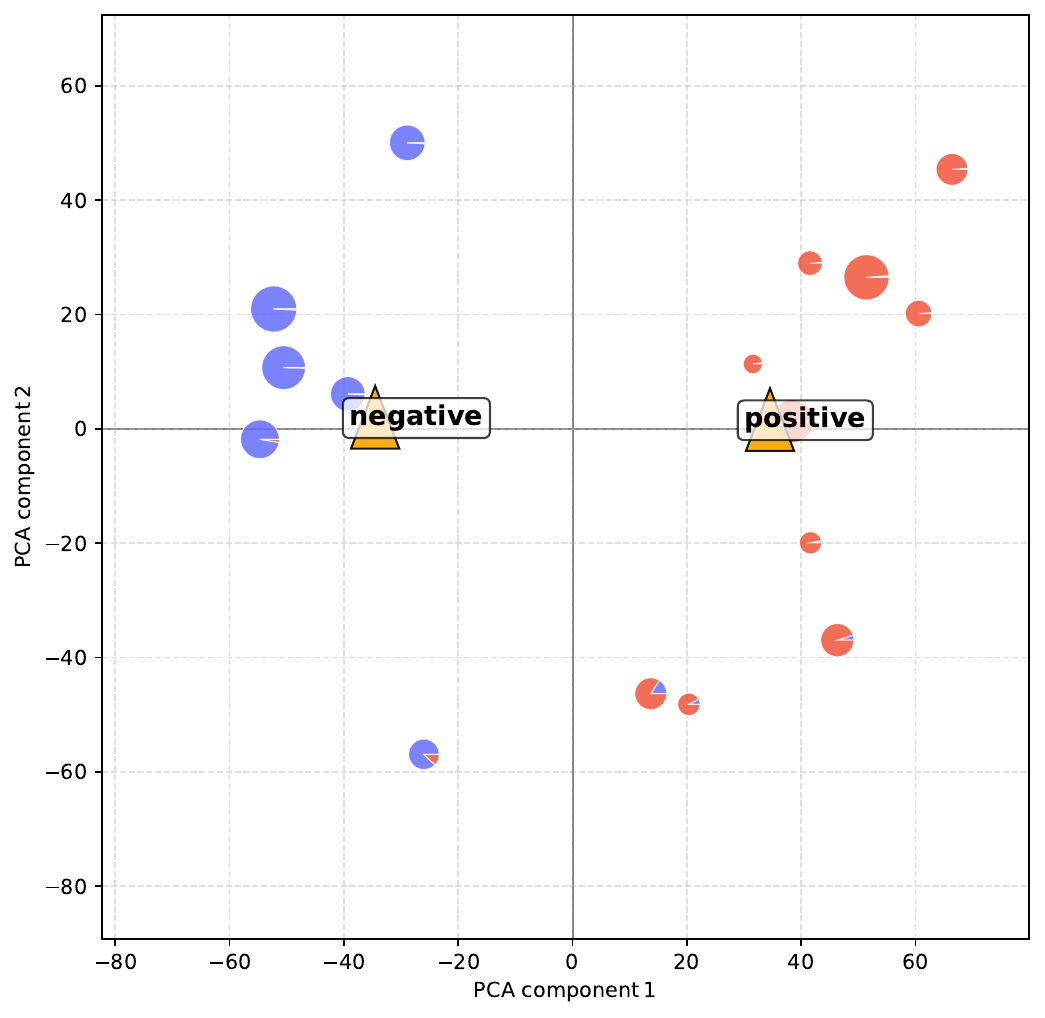}
  \cprotect\caption{2D Principal Component Analysis (PCA) fitted on sentence-level hidden-state activations extracted from the residual stream of the penultimate transformer block of LLaMA 3.1 Instruct tasked on two classification datasets: AG News (top) and IMDB (bottom). The colored circles stand for the concepts learned by $\verb|ClassifSAE|$. Their size is proportional to their mean activation over the dataset. The proportion of color is representative of the normalized class score for each concept. The triangle symbols depict the class prototypes activations.}
  \label{fig:pca_agnews_imdb_llama}
\end{figure}

\subsection{Concepts visualization}

For each dataset, we provide a simplified 2D PCA projection of the concept embeddings learned by \verb|ClassifSAE|. These projections are shown relative to the corresponding category prototypes in Figures~\ref{fig:pca_agnews_imdb_llama} and \ref{fig:pca_agnews_te_offensive_te_sentiment_llama}. The visualizations also include the normalized class scores described in Appendix~\ref{sec:feature_segmentation}. Features aligning with the same majority class naturally cluster around their prototypes, illustrating $\verb|ClassifSAE|$’s ability to capture hidden-state structures that are discriminative for classification. Figure~\ref{fig:fig_main} and Figures~\ref{fig:examples_concepts_agnews_gptj} and~\ref{fig:example_concept_te_sentiment_gptj} in Appendix~\ref{sec:concepts_illustration} present qualitative examples of concepts captured by \verb|ClassifSAE| for two datasets, illustrating that the specialized classification loss does not collapse all concepts into a single one per category. It instead supports the emergence of fine-grained and semantically coherent concepts, which remain aligned with the target categories while differentiating distinct subtopics.

\section{Conclusion}

We introduced \verb|ClassifSAE|, a supervised SAE-based method for discovering complete, causal and interpretable concepts from the internal representations of an LLM specialized for a text classification task. We compared this model in a comprehensive quantitative study to investigate what constitutes good concepts for explaining the decisions of black-box LLM classifiers, an area where evaluation has traditionally relied on qualitative analysis. In particular, we proposed two new metrics to quantify the monosemanticity and coherence of sentence-level features without requiring human annotations. Across multiple task–model pairs, \verb|ClassifSAE| outperforms ICA, ConceptSHAP and TopK-SAE, showing that task-aware architecture and losses enhance SAE feature quality for the target task. While HI-Concept achieves stronger causality scores due to its explicit objective, \verb|ClassifSAE| computes concepts up to 83\% faster and surpasses it in interpretability, underscoring that current methods do not yet fully reconcile causality and interpretability and suggesting a direction for future work.

\section{Limitations}

While $\verb|ClassifSAE|$ aims to break polysemanticity to uncover more precise concepts, quantifying this property remains challenging, as no single metric fully captures its different aspects. We include qualitative examples to illustrate the practicality of the extracted concepts and to complement our quantitative evaluation. However, end-to-end pipelines that go beyond simple integrated gradients, linking layer-level concepts to both the input sentence and the model's output, are still sparse in the literature. Developing such pipelines could be a promising direction for future work. Our analysis also focuses on comparing concepts extracted at a fixed layer. Introducing layer choice as an additional degree of freedom could provide better results, but would significantly increase computational costs, especially for large models. Finally, our study is limited to the text modality. While sparse autoencoders (SAEs) have demonstrated applicability across different modalities, investigating how our approach generalizes beyond text remains an open question.

\section{Acknowledgments}

This work received financial support from Crédit Agricole SA through the research chair “Trustworthy and responsible AI” with École Polytechnique. This work was granted access to the HPC resources of IDRIS under the allocation AD011015063R1 made by GENCI.

\bibliography{main}

\appendix

\newpage

\section{Features segmentation strategy}\label{sec:feature_segmentation}

To enrich our framework, we rely on a segmentation scheme to cluster learned features in segments whose number matches the total available categories. The aim is to facilitate post-analysis in proposing an automatic normalized score of each class for every feature. 

We denote $C$ the set of possible categories and we remind that $\textbf{z}$ corresponds to the projection of the sentence hidden state in the new concepts space. Therefore $z^j$ is a scalar and the activation strength of the $j$-th feature. For the SAE-based methods, since we only conserve a subset of the learned latent variables, the inspected part is restricted to $\textbf{z}_{\text{class}}$. The \textit{mean activation} of each concept \mbox{ $j \in \llbracket 1,m \rrbracket$ } on the dataset $\mathcal{D}$ is defined as :

\begin{equation}\label{eq:mean_activation_feature}
    \bar{z}^j  = \frac{1}{|\mathcal{D}|} \displaystyle \sum_{i \in \mathcal{D}} \, | z^j_i |  \qquad ,\forall j \in \llbracket 1,m \rrbracket
\end{equation}

and the \textit{normalized score of each class} for every feature is computed as : 

\begin{equation}\label{eq:normalised_score_feature}
    \begin{aligned}
        s_c(j) =& \frac{\bar{z}^j_c}{\bar{z}^j}  \quad ,\forall j \in \llbracket 1,m \rrbracket \quad ,\forall c \in C \\ =& \frac{1}{ \bar{z}^j } \frac{1}{|\mathcal{D}_c|} \displaystyle \sum_{i \in \mathcal{D}_c} \, | z^j_i | 
     \end{aligned}
\end{equation}

where $\mathcal{D}_c$ stands for the subset of $\mathcal{D}$ which only comprises the sentences categorized as belonging to the class $c$. Based on these quantities, we segment features according to the class maximizing their normalized score.

\begin{equation}
    \mathcal{F}_c := \{j : c = \argmax_{c' \in C} s_{c'}(j)\} \, ,\forall c \in C
\end{equation}

\begin{figure}[htbp]
  \centering
  \includegraphics[width=\linewidth]{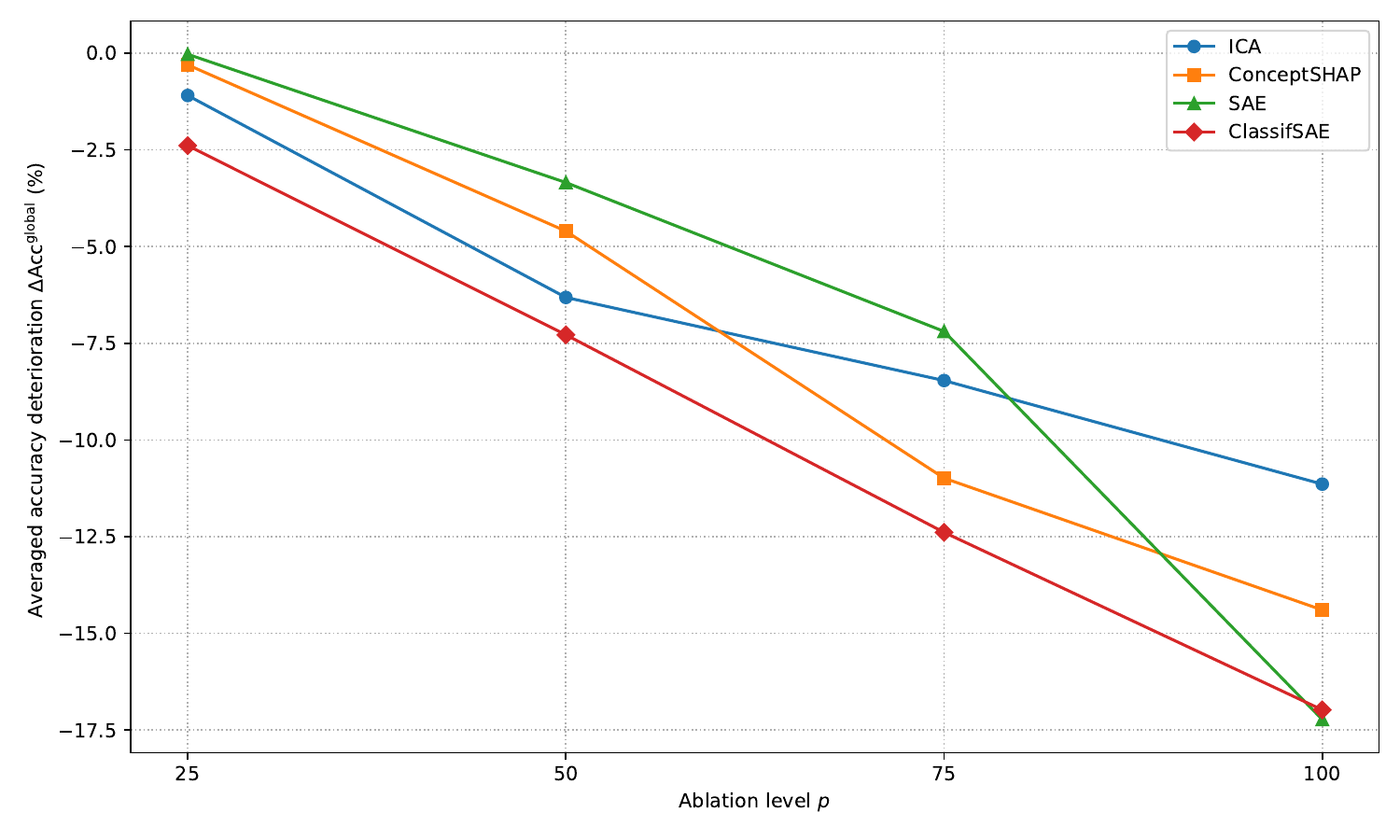}
  \caption{Averaged accuracy deterioration $\Delta \text{Acc}^{\text{global}}$ as a function of the ablation level $p$ of class-specific features segments $(\mathcal{F}_c)_{c \in C}$. Results are reported for concepts computed from hidden states extracted at the residual stream exiting the penultimate transformer block of Pythia-1B fine-tuned on AG News.}
  \label{fig:global_causality_on_classes}
\end{figure}

We leverage the knowledge of class-specific features segments $(\mathcal{F}_c)_{c \in C}$ to account for joint global causal effect. For each label $c$ and each ablation level $p \in \{25,50,70,100\}\%$, we successively ablate the top $p\%$ features in $\mathcal{F}_c$ ranked by their mean absolute activation, leaving all other segments untouched. We then measure accuracy deterioration for each $(\mathcal{F}_c,p)$ pair and average over all segments to yield a mean impact score at ablation rate $p$. We perform this procedure for most of the concept-based post-hoc methods under comparison for Pythia-1B on AG News and we report $\Delta \text{Acc}^{\text{global}}$ averaged across the class-specific features segments in Figure \ref{fig:global_causality_on_classes}. We observe that for concepts computed by $\verb|ClassifSAE|$, the decline in averaged accuracy deterioration is both more pronounced and more consistent as $p$ increases, in contrast to the other methods. Although SAE initially lags behind the others, its averaged $\Delta \text{Acc}^{\text{global}}$ converges with that of $\verb|ClassifSAE|$ once a whole class-specifc features segment is ablated. This pattern reflects better decorrelation among the $\verb|ClassifSAE|$ concepts allocated to the same class-specific segment. When features assigned to the same segment lack sufficient precision for their associated fine-grained notion, they tend to activate with similar magnitudes whenever a sentence from that class is processed, thereby negating the nuanced distinctions the concepts were meant to capture. By contrast, concepts that are less correlated and more precise each contribute incrementally to a monotonic decrease of the model's accuracy deterioration as they are sequentially removed. 

\section{Baselines}\label{sec:baselines}

\paragraph{Independent Component Analysis (ICA)} \citet{COMON1994287} introduced this technique to separate a multivariate signal into additive, independent non-Gaussian signals, resulting in components that are more interpretable to humans than individual neurons. It is a recognized non-parametric clustering method that can be applied on neural networks activations.

\paragraph{ConceptShap} It extracts interpretable concept vectors from neural networks by maximizing their contribution to the classifier completeness. While the term originally refers to a metric measuring the marginal contribution of each concept, it is often used to describe the full pipeline for identifying human-aligned features proposed in \citet{NEURIPS2020_ecb287ff}. The semantic meaning of the concepts is emphasized by encouraging them to align closely with specific input examples. However, this clustering approach incurs a high computational cost, as it requires calculating a distance measure between all features and every sample in the training set.

\paragraph{HI-Concept}
Building on the ConceptSHAP framework, HI-Concept \cite{zhao-etal-2024-explaining} introduces two innovations. It jointly minimizes a reconstruction loss to preserve the original hidden representations and explicitly maximizes the causal impact of each concept on the classifier’s predictions  via a causal loss that estimates treatment effects through randomized ablations of concept subsets. This causal objective, however, introduces additional computational overhead, as it requires repeated forward passes through the downstream layers of the model. As in ConceptSHAP, the semantic interpretability of concepts is encouraged through the same pair of regularization losses that promote alignment with specific input examples.

\paragraph{TopK SAE} We compare $\verb|ClassifSAE|$ to TopK SAE to measure the benefits of the two additional components we introduced. The training loss is identical to that described in Section \ref{sec:problem_settings}. Since there is no trained classifier clustering task-relevant features, we imitate previous approaches and train a logistic probe in post-processing to extract $\textbf{z}_{\text{class}}$.

\section{Implementation details}\label{sec:methods_implementation_details}

In this section, we detail the hyperparameter settings specific to each of the compared methods. The expected number of extracted concepts is fixed and kept identical across all approaches for fair comparison. In selecting hyperparameters, our primary objective is to ensure a decent recovery accuracy, as it reflects the completeness of the learned concepts and thus their relevance as proxy of the model’s internal representations. Within this constraint, we tuned parameters to jointly optimize concepts causality and interpretability, while ensuring that the recovery accuracy remained above 80\%. The seed is set to $42$ for all trainings.

\paragraph{ICA} 
We implement it using the FastICA method from scikit-learn \cite{scikit-learn}, with whitening set to unit-variance, the extraction algorithm set to parallel and the maximum number of iterations fixed at 1000. Unlike the methods described below, ICA does not have a natural sparsity mechanism for the activations of learned components, such as a threshold like ConceptSHAP. Hence, the absolute values of the components' activations are never zero.
 
\paragraph{ConceptSHAP} 
The method uses two auxiliary losses in addition of the completeness loss. Let denote $(\textbf{c}_j)_{j \in \llbracket 1,m \rrbracket}$ the concepts learned by the method. In the first regularizer term, the interpretability of each $\textbf{c}_k$ is enhanced by maximizing $\textbf{h} . \textbf{c}_k$ for all sentence embeddings $\textbf{h}$ belonging to the set of top-K nearest neighbors of $\textbf{c}_k$. The second term enforces diversity among the learned concepts by minimizing $\textbf{c}_k . \textbf{c}_q$ for all pairs $(\textbf{c}_k,\textbf{c}_q)$. For all experiments, we set the regularizer weights to $\lambda_1 = 0.1$ and  $\lambda_2 = 0.5$. The activation of $\textbf{c}_k$ given the input $\textbf{h}$ is computed with the formula $\text{TH}(\textbf{h} . \textbf{c}_k,\beta)$ where $\beta \geq 0$ acts as a threshold value below which the activation of the $k$-th concept is set to zero. Following guidance from  \cite{NEURIPS2020_ecb287ff}, we set the batch size to $128$ and $\beta$ is chosen within $\{0.1, 0.2, 0.3 \}$. Larger values of $\beta$ impose a higher activation threshold, producing sparser concepts responses and making them more likely to be interpretable, though this may come at the cost of lower recovery accuracy as some information can be discarded. Therefore, we start with $\beta=0.3$ for each pair model-dataset and decrease its value if the recovery accuracy in the validation phase does not match the requirement. The top-K value is set to $32$, a fourth of the batch size. The model is trained with an Adam optimizer and a learning rate of $3e-4$. We fixed the number of epochs at $100$. The reconstruction of the original hidden state from the concepts activations is handled by a 2-layer MLP with a hidden dimension of $512$.

\paragraph{HI-Concept}  For the architectural components shared with ConceptSHAP, we reuse the same hyperparameter settings, with the exception of the threshold parameter. Following the authors’ recommendation, we automatically set the threshold to $\beta = \frac{1}{n}$, where $n$ is the target number of extracted concepts. For the remaining parameters, we follow the setup of \citet{zhao-etal-2024-explaining}: all loss components are weighted equally, and the random concept masking strategy is used for computing the causal loss. To stabilize training, the causal loss is frozen during the first half of the learning.

\paragraph{SAE} Building on the open-source library SAE-Lens codebase \cite{bloom2024saetrainingcodebase} for SAE implementation in language models, we adapt the method to extract concepts from a single hidden state encoding the LLM's sentence classification decision. We reuse their implementation of the ghost grad method as auxiliary loss. In all experiments, we set the training batch size to $500$, we select $K=10$ for the TopK activation function, we use the Adam optimizer with an initial learning rate of $5e-5$ and a cosine annealing schedule down to $5e-7$. For the inspected LLMs, the SAE hidden layer size $d_{sae} \in \mathbb{N}$ is set to twice the dimensionality of the input residual stream. The influence of this parameter is examined in the ablation studies in Section \ref{sec:ablation_studies_section}. As our datasets are of moderate size, the model benefits from repeated exposure to each sentence’s embedding. We use a total of $10,000,000$ training tokens for all experiments. Lastly, we also take advantage of several SAE-Lens built-in utilities for encoder–decoder tied initialization, fixed-norm decoder columns, and hidden‐layer activation normalization.

\paragraph{ClassifSAE} The SAE component of the architecture is trained using the same procedure as outlined in the previous section. The main difference is the integration of $\mathcal{L}_{\text{class}}$ and $\mathcal{L}_{\substack{\text{sparse} \\ \text{feature}}}$ in the training loss. We implement the joint classifier head as a single linear layer with a bias term.  For deep layers, this simple architecture suffices to achieve strong reconstruction accuracy, as category-specific features are well separated in the SAE representation at that stage. We set the weights for the losses $\mathcal{L}^\text{TopK}_\text{SAE}$, $\mathcal{L}_{\substack{\text{sparse} \\ \text{feature}}}$ and $\mathcal{L}_{\text{class}}$ to $0.01$, $0.01$ and $1$, respectively.

\paragraph{Source Code} We release the Python source code and SLURM scripts to reproduce the concepts analysis experiments presented in this paper. The repository includes the license and project documentation. The code is intended for research use only. 

\begin{center}
    \href{https://github.com/orailix/ClassifSAE}{https://github.com/orailix/ClassifSAE}
\end{center}

\section{Interpretability metrics}\label{sec:interpretability_metrics}

\paragraph{Sentence Encoder} In all our experiments, we employ the \texttt{all-MiniLM-L6-v2} sentence-embedding model \cite{reimers-gurevych-2019-sentence} to obtain vector representations that reflect the semantic distribution of sentences in our dataset. 

\paragraph{1D Clustering} Concept activation vectors often contain small nonzero values due to noise or artifacts from SAE representations. These low-magnitude activations can blur the true firing threshold of a concept, making a fixed threshold at zero unreliable. To address this, we apply a one-dimensional clustering procedure to estimate a more semantically meaningful activation cutoff. Specifically, we use the Jenks natural breaks optimization \cite{Jenks1967TheDM}, which partitions the activation distribution into two clusters by minimizing intra-cluster variance. The resulting breakpoint defines the activation threshold. Sentences associated with values above it are labeled as "activating sentences" for the inspected feature. We implement this procedure using the \texttt{jenkspy} library \cite{jenkspy_library}.

\begin{figure*}[ht]
  \centering
  \includegraphics[width=\linewidth]{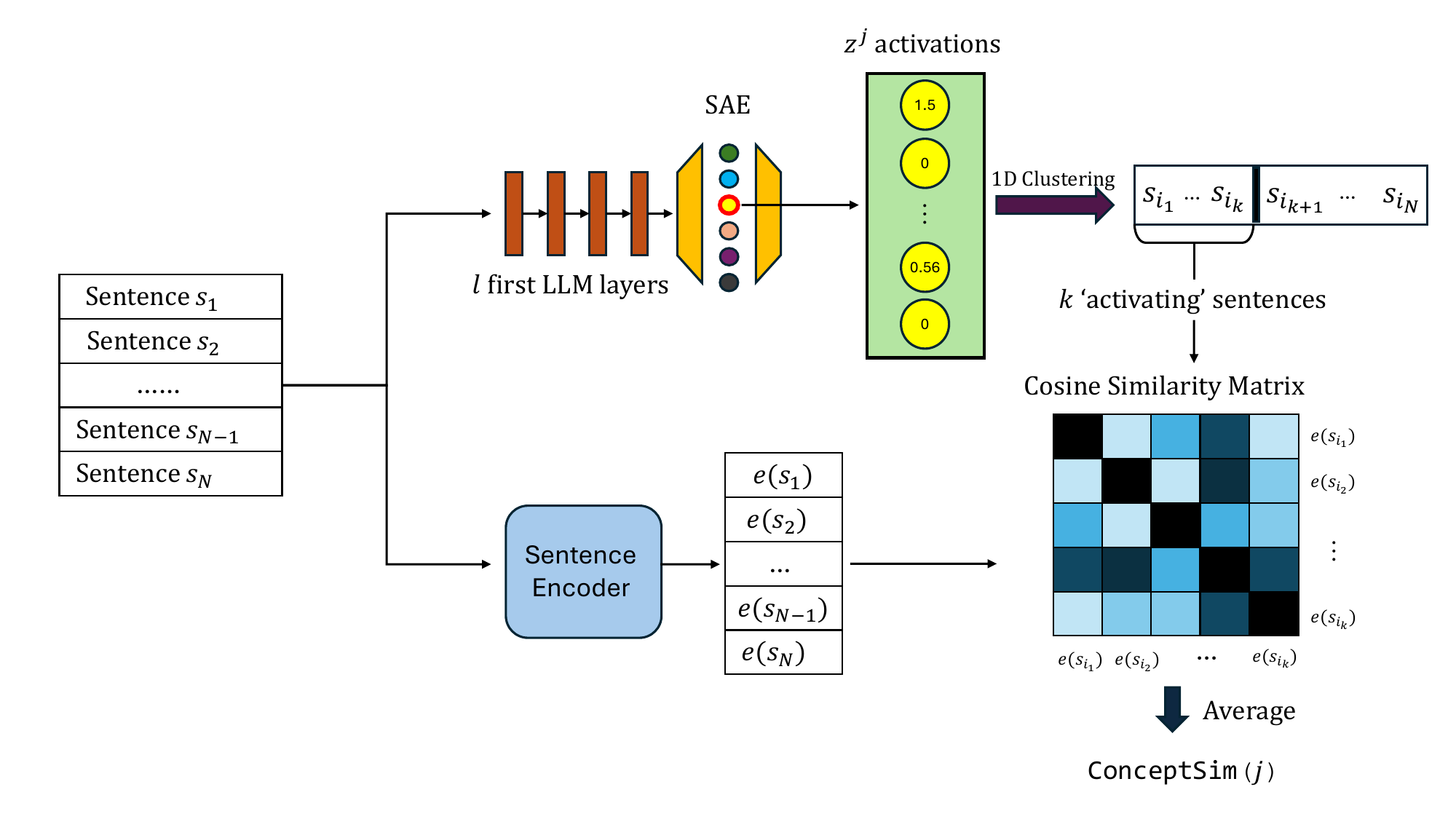}
  \cprotect\caption{Illustrative explanation procedure to compute $\verb|ConceptSim|(j)$ for the $j$-th evaluated concept.}
  \label{fig:activating_sentences}
\end{figure*}

\paragraph{Single metric for ConceptSim} Since we want to evaluate the overall interpretability of the extracted concepts, we derive a single metric from the concepts scores $(\verb|ConceptSim|(j))_{j \in \llbracket 1,m \rrbracket}$, simply referred to as  $\verb|ConceptSim|$. Each concept \mbox{$j \in  \llbracket 1,m \rrbracket$} is associated with a set of activating sentences and $\verb|ConceptSim|(j)$ approximates the expected cosine similarity between two sentences randomly drawn from this set. Therefore, we report $\verb|ConceptSim|$ as a weighted average of the individual scores, with weights given by the number of sentence pairs $\binom{N^j}{2}$ per concept. The pair-weighted average prevents inflated scores caused by rarely active concepts that fire on a few semantically similar sentences and appear unrealistically coherent, overshadowing more frequent, low-coherence concepts

\section{Datasets}\label{sec:dataset_details}

Dataset statistics are summarized in Table \ref{tab:datasets_details}. AG News \cite{10.5555/2969239.2969312} and IMDB \cite{maas-etal-2011-learning} are class-balanced, unlike TweetEval Offensive and TweetEval Sentiment~\cite{barbieri2020tweeteval}.  AG News is a topic classification dataset containing news articles labeled across four categories: World, Sports, Business and Sci/Tech. IMDB is a binary sentiment analysis dataset composed of movie reviews labeled as positive or negative. TweetEval Offensive is a tweet classification dataset for detecting offensive language in social media posts and TweetEval Sentiment is a sentiment analysis dataset of tweets labeled as positive, negative or neutral.

For each task, the train split is used to train the investigated concept-based methods and we rely on the test split to report the concepts metrics. 

\begin{table}[ht!]
\centering
\renewcommand{\arraystretch}{1.5} 
\resizebox{\columnwidth}{!}{%
\begin{tabular}{c|c c c c }
    \hline
    Dataset & Train size & Test size & Nb. labels & Avg nb. words \\
    \hline
    IMDB & 25k & 25k & 2 & 234 \\
    AG News & 120k & 7.6k & 4 & 38  \\
    TweetEval Offensive & 12k & 2k & 2 & 23\\
    TweetEval Sentiment & 45.6k & 14.3k & 3 & 19\\
    \hline
\end{tabular}
}
\caption{Summary statistics of the datasets used in our experiments}
\label{tab:datasets_details}
\end{table}

\section{Classification 
models}\label{sec:classifier_details}

To investigate models that are effective in classification settings, we fine-tune five LLM backbones independently on the four classification datasets. For each backbone–dataset pair, we train a distinct fine-tuned model. 
We have selected BERT-Base \cite{devlin-etal-2019-bert} and DeBERTa-v3-Large \cite{he2021debertadecodingenhancedbertdisentangled} as representatives of encoder-only architectures and Pythia-410M, Pythia-1B \cite{biderman2023pythia} and GPT-J-6B \cite{gpt-j} for auto-regressive architectures. Figure \ref{fig:example_prompt_template} shows an example of prompt template used to cast the auto-regressive LLMs into a classification setting by formatting each sentence accordingly. Training is performed with the \texttt{Trainer} method from HuggingFace Transformers \cite{wolf-etal-2020-transformers}. Additionally, we include two larger models: Mistral-7B v0.1 Instruct \cite{jiang2023mistral7b} and LLaMA 3.1 8B Instruct \cite{grattafiori2024llama3herdmodels}. They are not fine-tuned, instead we compute a task-specific prompt embedding for each architecture-dataset pair and concatenate it to the input token embeddings. This allows us to align them with the classification task without updating their weights, using soft prompt tuning. We report the individual performance of each aligned model on its corresponding dataset in Table~\ref{tab:table_performance_classifiers}.

\begin{table}[ht!]
\centering
\scriptsize                          
\setlength{\tabcolsep}{2.8pt}        
\label{tab:rot_results}
\renewcommand{\arraystretch}{1.5}
\begin{tabular}{ll rrrr}
\hline\hline
\multicolumn{2}{c}{ } &
\multicolumn{4}{c}{\textbf{Classifier Model}}\\
\cline{3-6}
\multicolumn{2}{c}{} 
& AG News & IMDB & TE-Off. & TE-Sent.  \\
\textbf{Method} & \textbf{Metric} &
\multicolumn{4}{c}{}\\
\hline

BERT & $M_{train}$ & 120000 & 120000 & 60000 & 180000\\
        &  Acc. (\%) & 93.00 & 93.50 & 82.91 & 70.63\\
        & Macro-F1 & 0.935 & 0.935 & 0.800 & 0.708\\
        \cline{1-6}
DeBERTa-v3 &  $M_{train}$  & 120000 & 20000 & 40000 & 80000 \\
        & Acc. (\%)  & 94.60 & 96.65 & 81.90 & 73.01\\
        & Macro-F1 & 0.945 & 0.965 & 0.790 & 0.734\\
        \cline{1-6}
Pythia-410M & $M_{train}$  & 60000 & 60000 & 60000 & 80000\\
        & Acc. (\%) & 94.03 &95.12 &80.84 &71.73\\
        & Macro-F1  & 0.940 & 0.951 & 0.770 & 0.718\\
        \cline{1-6}
 Pythia-1B & $M_{train}$  & 60000 & 60000 & 60000 & 80000\\
        & Acc. (\%) & 94.07 & 95.60 & 81.12 & 71.72\\
        & Macro-F1 & 0.940 & 0.956 & 0.773 & 0.718\\
        \cline{1-6}
 GPT-J & $M_{train}$  & 10000 & 25000 & 30000 & 24000\\
        & Acc. (\%) & 93.04 & 96.50 & 81.71 & 73.22\\ 
        & Macro-F1  & 0.930 & 0.965 & 0.787 & 0.730\\
        \cline{1-6}
 Mistral-Instruct (PT) &  $M_{train}$ & 4000 &4000 &4000& 4000\\
        & Acc. (\%) & 90.20 & 80.84 & 76.62 & 66.27\\ 
        & Macro-F1  & 0.901 & 0.808 & 0.691 & 0.660\\
         \cline{1-6}
 Lama-Instruct (PT) &  $M_{train}$ & 4000 &4000 &4000& 4000\\
        & Acc. (\%) & 89.20 & 95.55 & 80.06 & 65.22\\ 
        & Macro-F1  & 0.892 & 0.955 & 0.750 & 0.648\\
\hline \hline
\end{tabular}
\cprotect\caption{Performance metrics of the classifiers (Accuracy and Macro-F1) across the different datasets. $M_{\text{train}}$ denotes the total number of training sentence instances each model was exposed to during fine-tuning on the corresponding dataset. When the dataset contains fewer than $M_{\text{train}}$ unique sentences, samples are reused across epochs and repetitions are counted towards $M_{\text{train}}$. For Mistral-Instruct and LLaMA-Instruct, $M_{\text{train}}$ instead indicates the number of sentences used to compute the dataset-specific prompt embedding. Our objective is not to achieve state-of-the-art classification performance, but rather to tune models to a level of predictive reliability sufficient to enable the extraction of classification-relevant concepts.}
\label{tab:table_performance_classifiers}
\end{table}

\begin{figure}[ht!]
\centering

\begin{minipage}{\linewidth}

\tcbset{
  colback=gray!5!white, 
  colframe=black, 
  width=\linewidth, 
  boxrule=1pt, 
  arc=1mm, 
  fontupper=\small\ttfamily 
}

\begin{tcolorbox}
\textbf{\textsf{Prompt}}: Roddick and Williams to Star on Sunday at U.S. Open  NEW YORK (Reuters) - The first week of the U.S. Open  concludes Sunday with Andy Roddick and Serena Williams the star  attractions at Flushing Meadows.\textbackslash n\textbackslash nOPTIONS:\textbackslash n0(World)\textbackslash n1(Sports) \textbackslash n2(Business)\textbackslash n3(Sci/Tech)\textbackslash n1
\end{tcolorbox}

\end{minipage}

\caption{Example from our training set, based on the AG News dataset \cite{10.5555/2969239.2969312}. At the end of the sentence to be classified, the possible categories are listed along with corresponding integers, enabling the model to respond with a single token. The ground-truth label is appended at the end of the prompt: in this example, the integer 1 for the category Sports}
\label{fig:example_prompt_template}
\end{figure}

\section{Computational Budget} All experiments were conducted on an HPC cluster, reaching a total of $420$ hours of computation on NVIDIA A100 GPUs.

\begin{figure}[ht!]
  \centering
  \includegraphics[width=\linewidth]{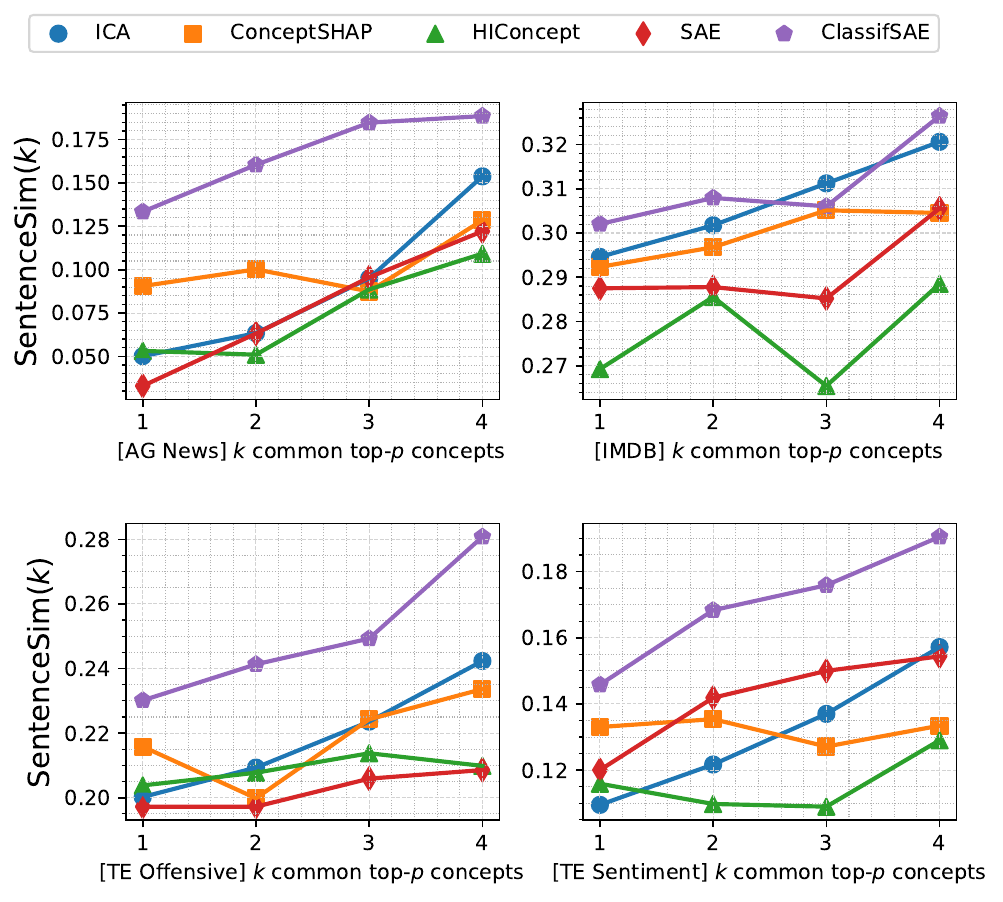}
  \cprotect\caption{$\verb|SentenceSim|(k)$ as a function of the number $k$ of shared top-activating concepts between sentence pairs. Concepts are learned from the sentence-level hidden states of the penultimate transformer block of \mbox{Pythia-1B} fine-tuned on one of the 4 classification tasks. We consider $p=5$ concepts for each sentence.}
  \label{fig:sentence_sim_pythia}
\end{figure}

\begin{figure}[ht!]
  \centering
  \includegraphics[width=\columnwidth]{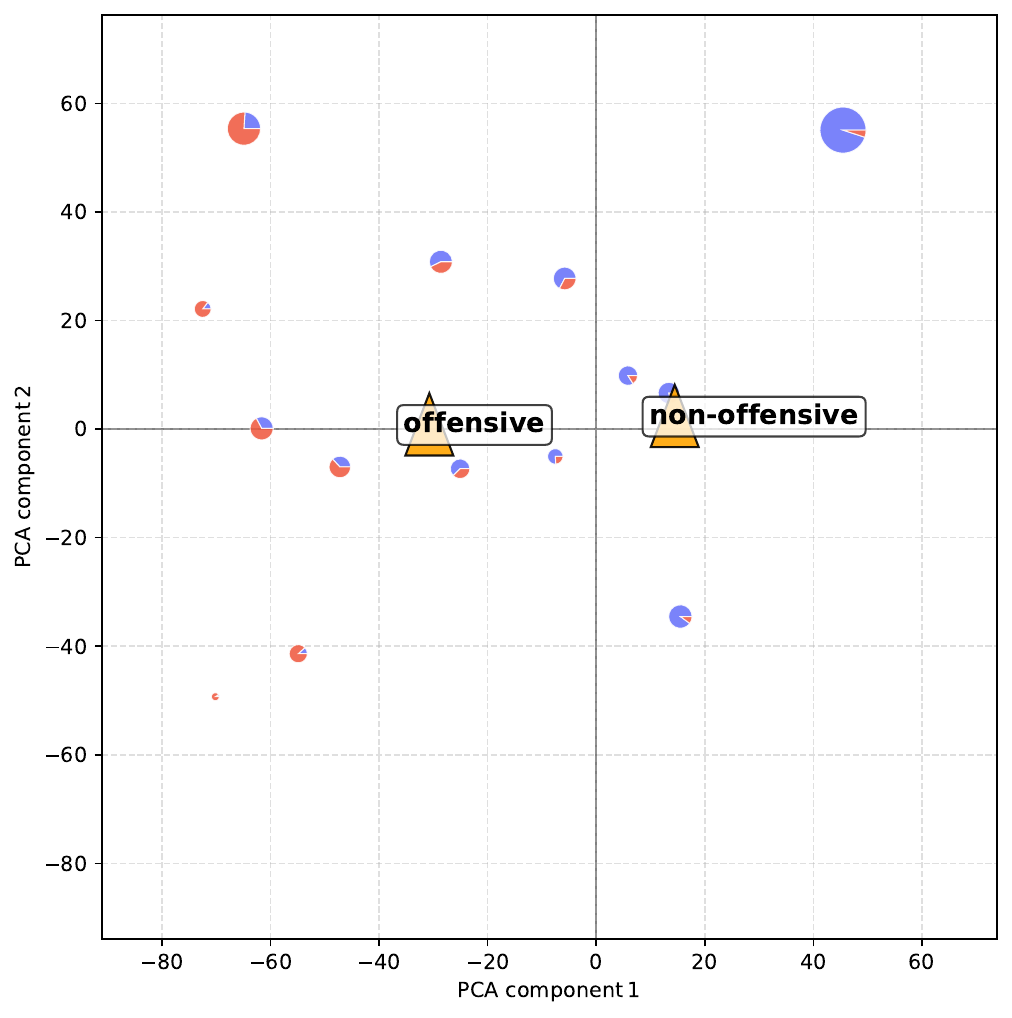}
  \includegraphics[width=\columnwidth]{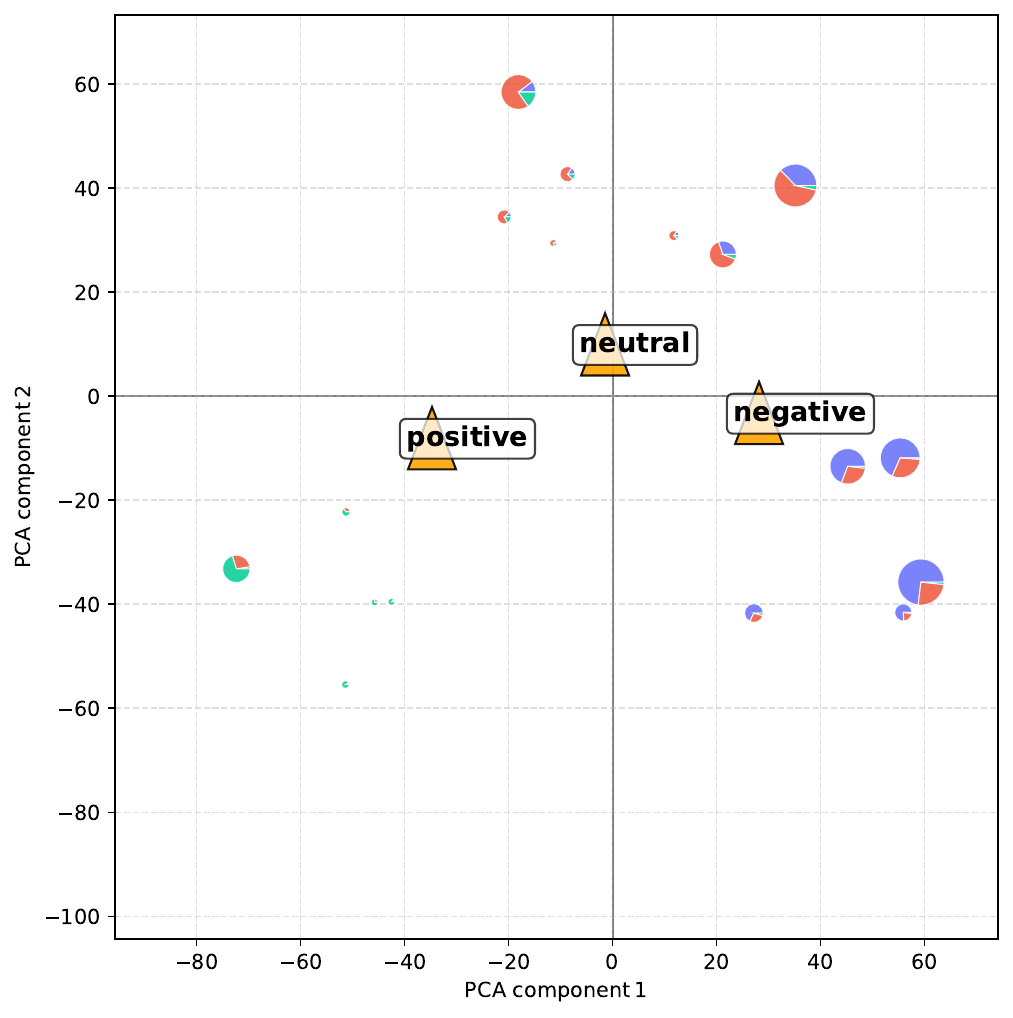}
  \cprotect\caption{2D Principal Component Analysis (PCA) fitted on sentence-level hidden-state activations extracted from the residual stream of the penultimate transformer block of LLaMA 3.1 Instruct tasked on two classification datasets: TweetEval Offensive (top) and TweetEval Sentiment (bottom). Depending from which dataset the sentence to classify come from, a prompt embedding previously computed is appended at the beginning of the sentence to align the model with the task. The colored circles stand for the concepts learned by $\verb|ClassifSAE|$. Their size is proportional to their mean activation over the dataset. The proportion of color is representative of the normalized class score for each concept. The triangle symbols depict the class prototypes activations.}
  \label{fig:pca_agnews_te_offensive_te_sentiment_llama}
\end{figure}

\section{Concepts illustrations}\label{sec:concepts_illustration}

We provide visualizations of some concepts discovered by \verb|ClassifSAE|, when trained on the internal activations of two distinct fine-tuned versions of GPT-J. Since displaying the top activating sentences per concept is not visually intuitive, we follow the approach of \citet{zhao-etal-2024-explaining} and represent concepts as word clouds. For each concept, we treat the top 500 activating sentences (or fewer, if the activating sentences set does not contain that many sentences) as a single document and compute word importance using TF-IDF. Word sizes in the resulting word cloud reflect their corresponding TF-IDF scores. For each displayed concept, we include two defining keywords in the caption. These keywords are generated by GPT-4, which is prompted with the top 20 activating sentences associated with the concept. We also display the category that each concept is predominantly associated with, based on our feature segmentation strategy detailed in Section~\ref{sec:feature_segmentation}. For the two inspected dataset, AG News and TweetEval Sentiment, we observe that $\verb|ClassifSAE|$ is capable of identifying finer-grained concepts beyond the coarse label categories. Moreover, features associated with the same majority class often exhibit nuanced preferences for distinct subtopics. This illustrates that the additional loss components introduced in \verb|ClassifSAE| do not hinder the SAE’s ability to capture fine-grained and semantically precise representations

\section{Inputs interpretability}\label{sec:input_interpretability}

\begin{figure*}[!htbp]
  \centering
  \begin{subfigure}[b]{\textwidth}
    \centering
    \includegraphics[width=\textwidth]{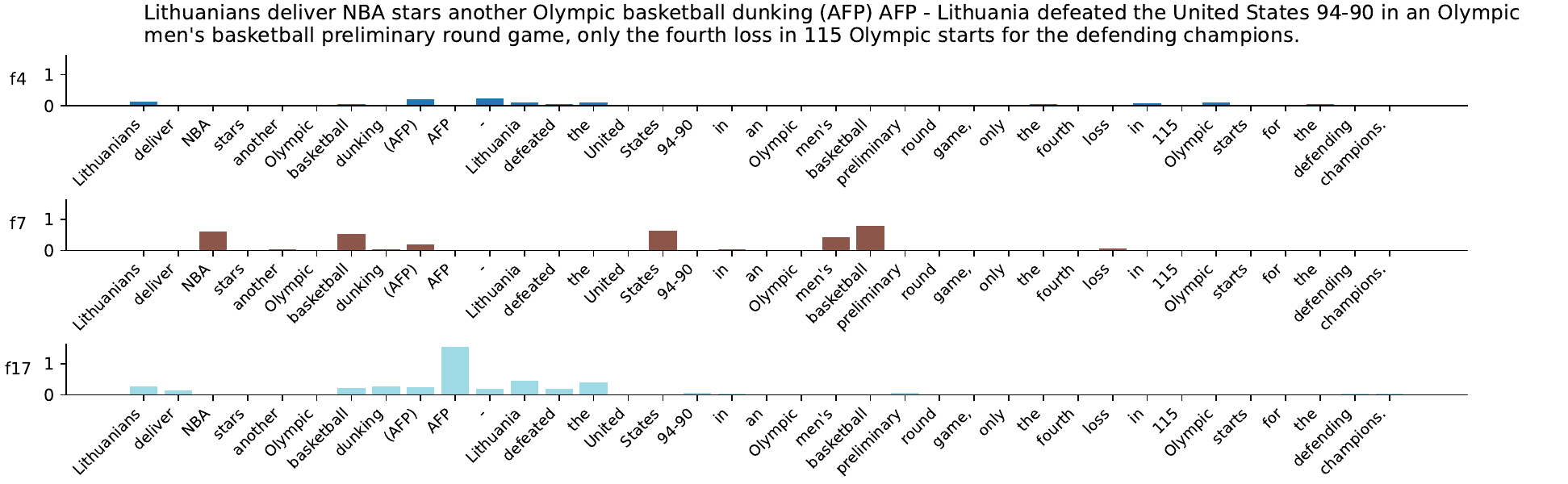}
    \subcaption{Attribution of each word with regard to the activation of the $3$ active feature in $\textbf{z}_{\text{class}}$ for this sentence. Attributions are computed via Integrated Gradients method as implemented in the Captum library \cite{kokhlikyan2020captum}}
    \label{fig:input1_classifsae}
  \end{subfigure}
  \\[1em] 
  \begin{subfigure}[b]{0.32\textwidth}
    \centering
    \includegraphics[width=\textwidth]{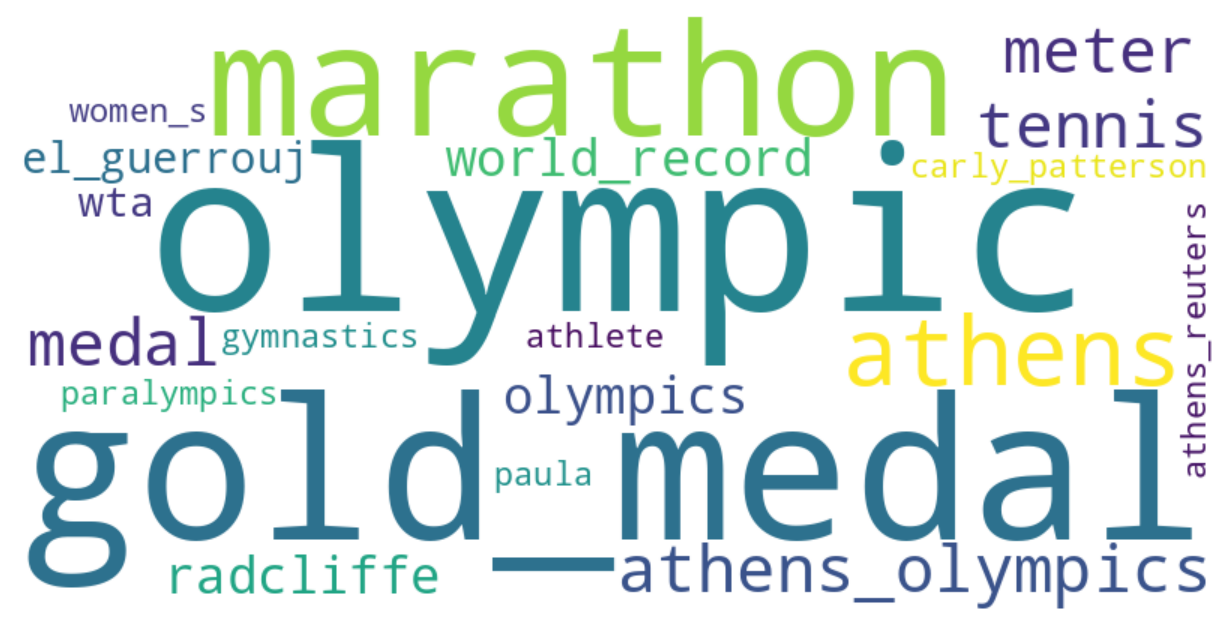}
    \subcaption{  \parbox{0.8\linewidth}{\centering
      Feature 4 - Category Sport\\
      Concept: Olympic}}
    \label{fig:feat4}
  \end{subfigure}%
  \hfill
  \begin{subfigure}[b]{0.32\textwidth}
    \centering
    \includegraphics[width=\textwidth]{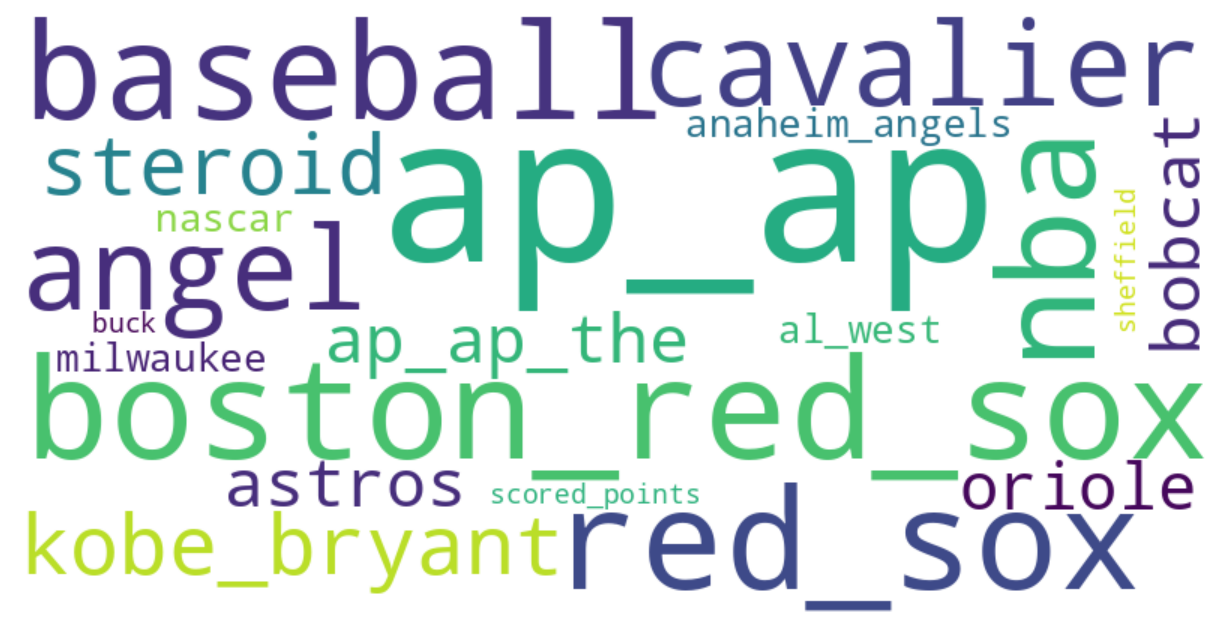}
    \subcaption{ \parbox{0.8\linewidth}{\centering
      Feature 7 - Category Sport\\
      Concept: American Sport}}
    \label{fig:feat7}
  \end{subfigure}%
  \hfill
  \begin{subfigure}[b]{0.32\textwidth}
    \centering
    \includegraphics[width=\textwidth]{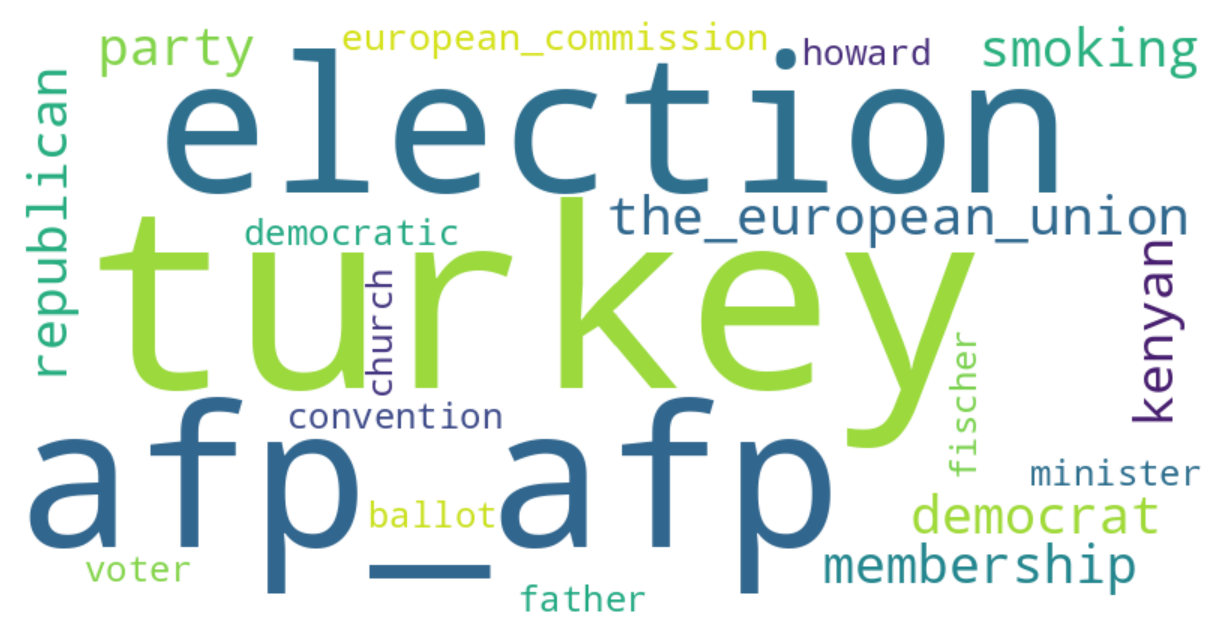}
    \subcaption{ \parbox{0.8\linewidth}{\centering
      Feature 17 - Category World\\
      Concept: AFP}}
    \label{fig:feat17}
  \end{subfigure}
  \cprotect\caption{Example of a sentence misclassified by our fine-tuned Pythia-1B model on the AG News dataset. The true label is Sport, but the model predicted the World category. The activated concepts computed by \verb|ClassifSAE| are shown, along with their respective attributions over the words in the sentence}
  \label{fig:input_1_classifsae_main}
\end{figure*}

\begin{figure*}[!htbp]
  \centering
  \begin{subfigure}[b]{\textwidth}
    \centering
    \includegraphics[width=\textwidth]{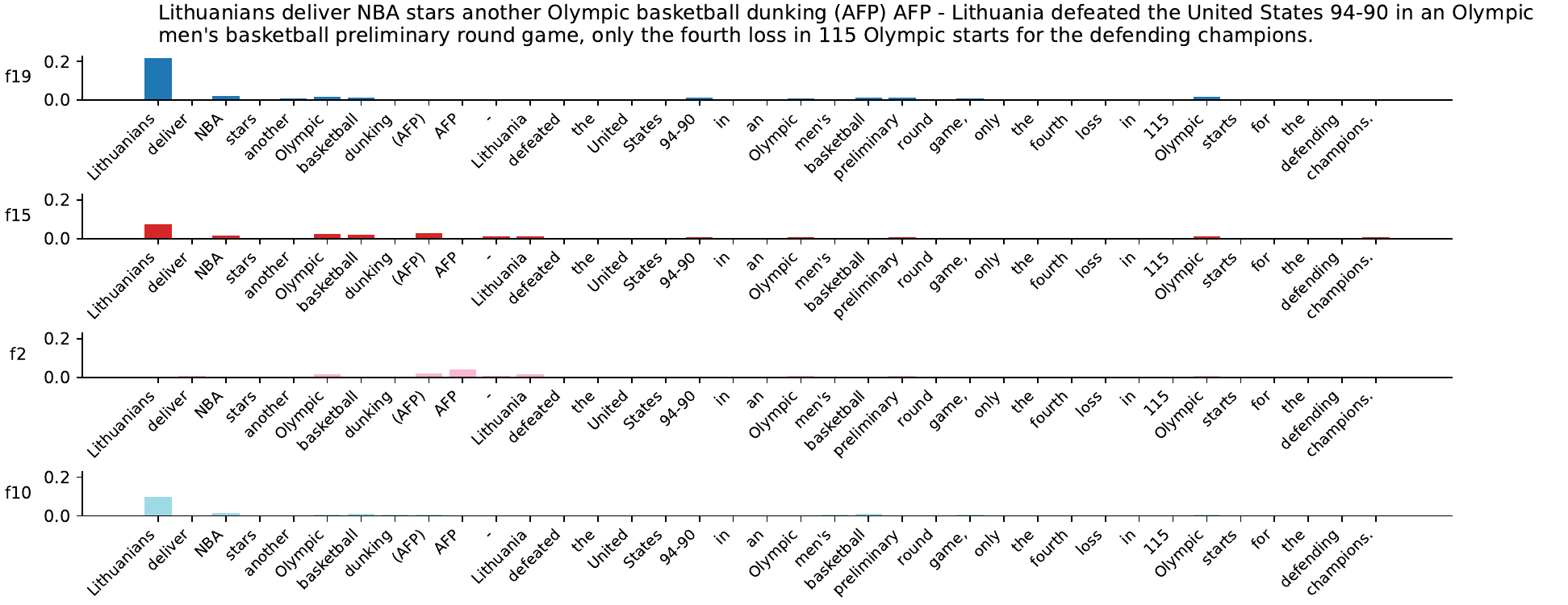}
    \subcaption{Attribution of each word with regard to the activation of the top $4$ most active features in $\textbf{z}_{\text{class}}$ for this sentence. Attributions are computed via Integrated Gradients method as implemented in the Captum library \cite{kokhlikyan2020captum}.}
    \label{fig:input_HI-Concept}
  \end{subfigure}
  \\[1em] 
  
  \begin{subfigure}[b]{0.24\textwidth}
    \centering
    \includegraphics[width=\textwidth]{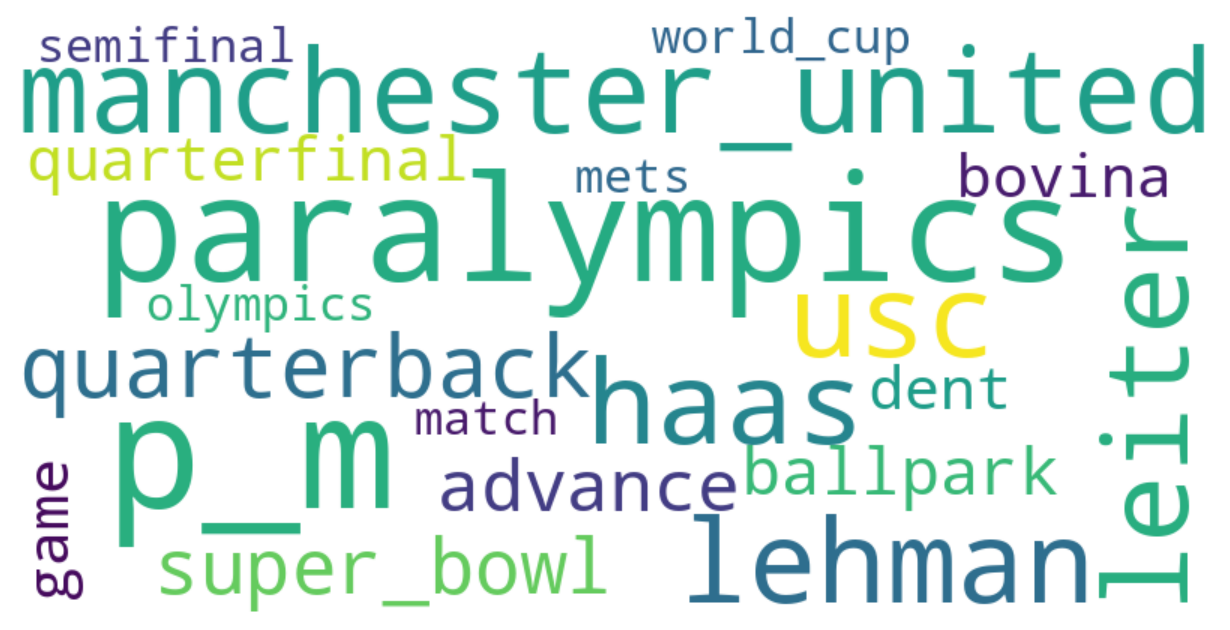}
    \subcaption{ 
      Feature 19 - Category Sport}
    \label{fig:feat19}
  \end{subfigure}%
  \hfill
  \begin{subfigure}[b]{0.24\textwidth}
    \centering
    \includegraphics[width=\textwidth]{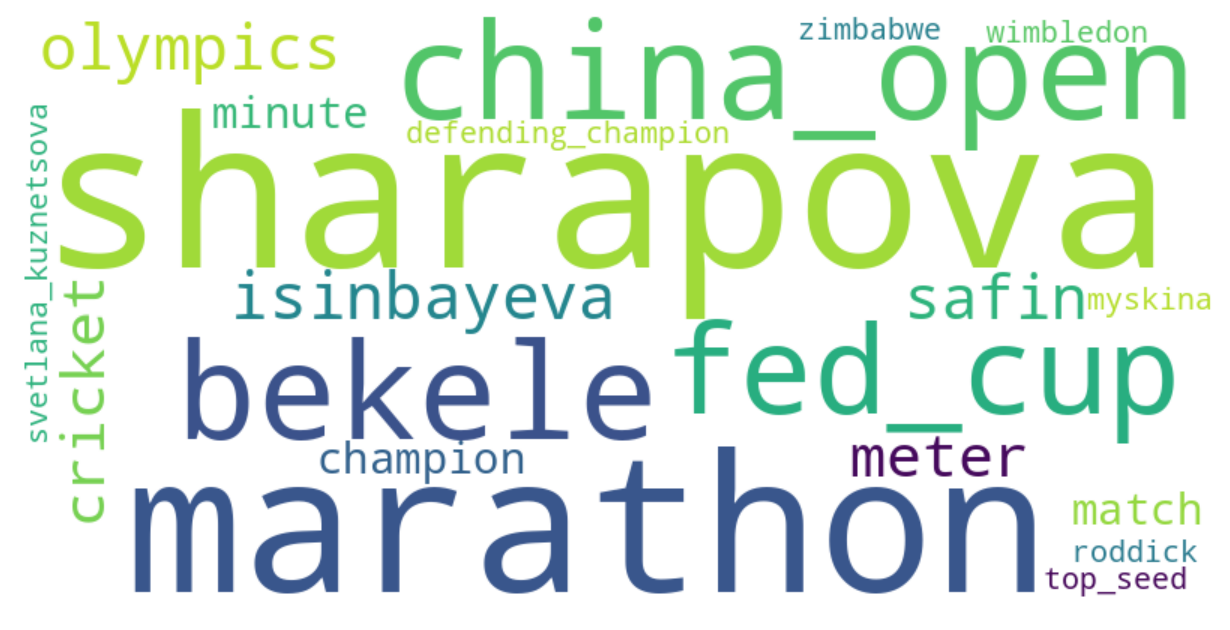}
    \subcaption{ 
      Feature 15 - Category Sport}
    \label{fig:feat15}
  \end{subfigure}%
  \hfill
  \begin{subfigure}[b]{0.24\textwidth}
    \centering
    \includegraphics[width=\textwidth]{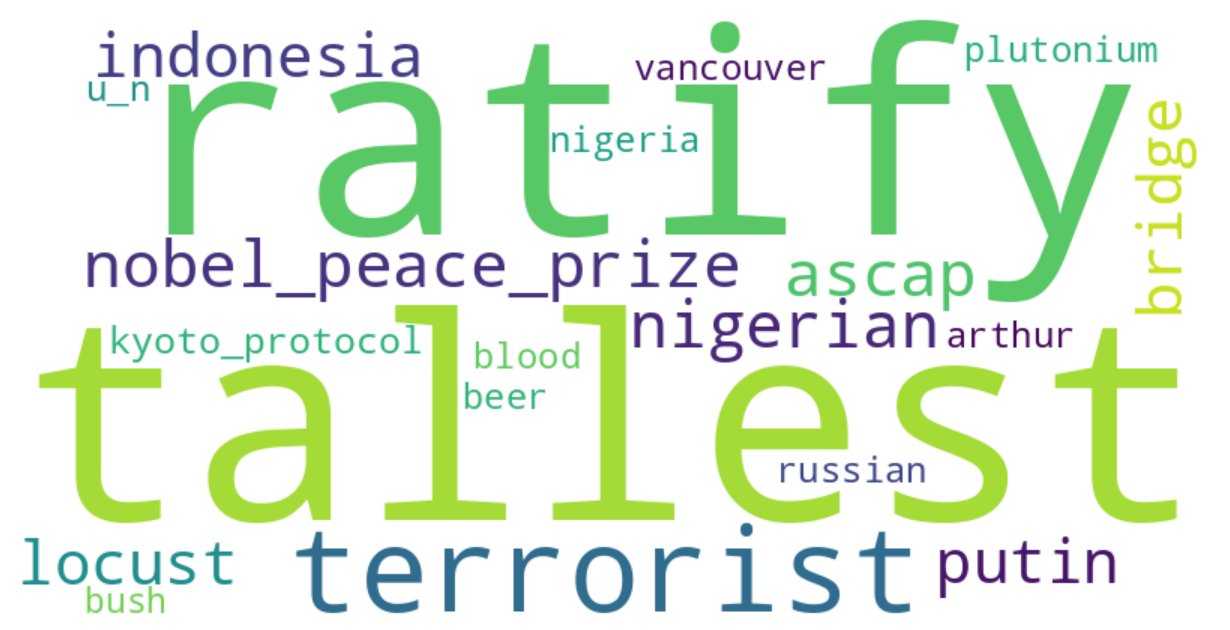}
    \subcaption{ Feature 2 - Category World}
    \label{fig:feat2}
  \end{subfigure}
  \hfill
  \begin{subfigure}[b]{0.24\textwidth}
    \centering
    \includegraphics[width=\textwidth]{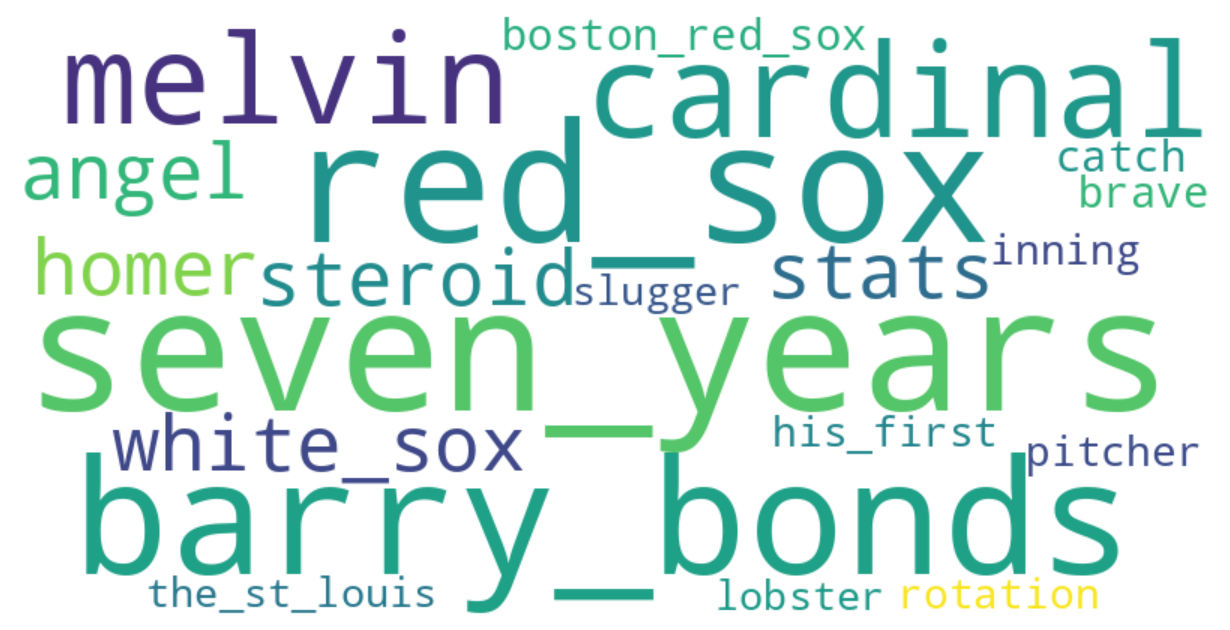}
    \subcaption{Feature 10 - Category World}
    \label{fig:feat10}
  \end{subfigure}
  \caption{Example of a sentence misclassified by our fine-tuned Pythia-1B model on the AG News dataset. The true label is Sport, but the model predicted the World category. The activated concepts computed by HI-Concept are shown, along with their respective attributions over the words in the sentence.}
  \label{fig:input_1_HI-Concept}
\end{figure*}

To illustrate one practicality of the computed concepts, we provide an example centered on input explainability. For each sentence, we first identify the neurons in $\textbf{z}_{\text{class}}$ that are activated. Then, for each such feature, we compute token-level attributions with respect to its activation. This procedure reveals which words contribute to the capture of the associated concepts. In practice, we use the Integrated Gradients method as implemented in the Captum library \cite{kokhlikyan2020captum}. Since Captum does not natively support attribution at the token level, we compute attributions with respect to the token embeddings and then sum the attributions across embedding dimensions. Although gradient-based methods are noisy and often yield diffuse attributions, they nevertheless provide useful intuition about which parts of a sentence contribute most to the emergence of a given concept. Figure \ref{fig:input_1_classifsae_main} and Figure \ref{fig:input_1_HI-Concept} compare the attributions of concepts found by the interpretability methods HI-Concept and $\verb|ClassifSAE|$. The inspected sentence was labeled as Sport in the AG News dataset, but our fine-tuned Pythia-1B model classified it under the World category. Analyzing the attributions from the 3 concepts activated by \verb|ClassifSAE|, we observe that, despite two features associated with the Sport class being activated, the presence of the pattern '(AFP) AFP' strongly triggered a concept linked to World events, likely due to its frequent occurrence in that context. The high activation of this concept ultimately led the model to predict the World category. In contrast, the concept attributions derived from the HI-Concept method did not provide meaningful insight for this example. Furthermore, the word clouds of the activated HI-Concept features appear less precise, offering less interpretable fine-grained concepts.

\section{Depth Effect on Concept Extraction}\label{sec:depth_effect}

\begin{figure}[htbp]
    \centering
    \begin{subfigure}[b]{0.45\textwidth}
        \centering
        \includegraphics[width=\textwidth]{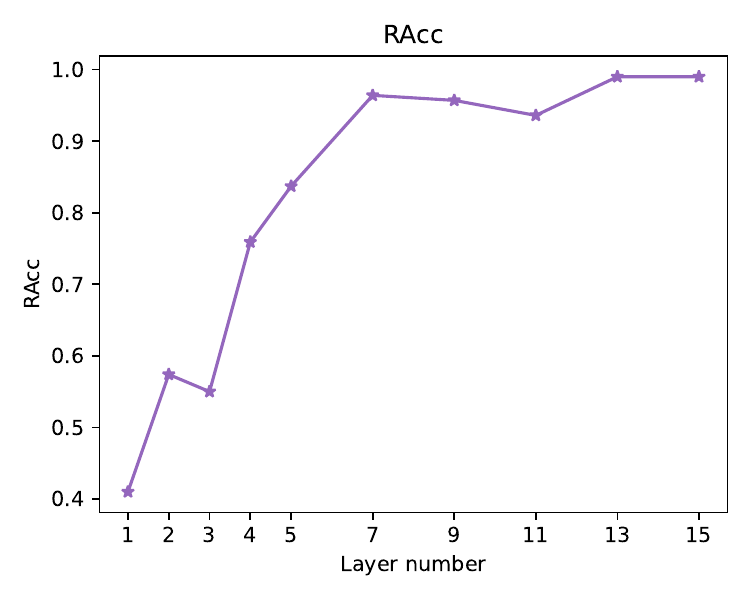}
        \cprotect\caption{Recovery accuracy (RAcc) of the concepts extracted by $\verb|ClassifSAE|$ across the layers of Pythia-1B fine-tuned on AG News}
        \label{fig:RAcc_pythia_agnews}
    \end{subfigure}
    \hfill
    \begin{subfigure}[b]{0.45\textwidth}
        \centering
        \includegraphics[width=\textwidth]{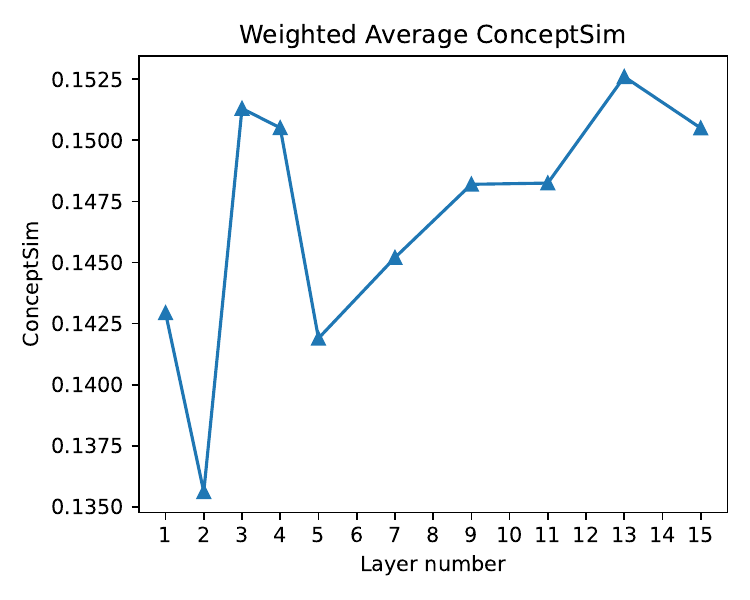}
         \cprotect\caption{Weighted Average $\verb|ConceptSim|$ of the concepts extracted by $\verb|ClassifSAE|$ across the layers of Pythia-1B fine-tuned on AG News}
        \label{fig:WA_Conceptsim_pythia_agnews}
    \end{subfigure}
    \caption{Comparison of the extracted concepts properties across layers of Pythia-1B fine-tuned on AG News.}
    \label{fig:layerwise_pythia_ag_news}
\end{figure}

\begin{figure}[htbp]
    \centering
    \begin{subfigure}[b]{0.45\textwidth}
        \centering
        \includegraphics[width=\textwidth]{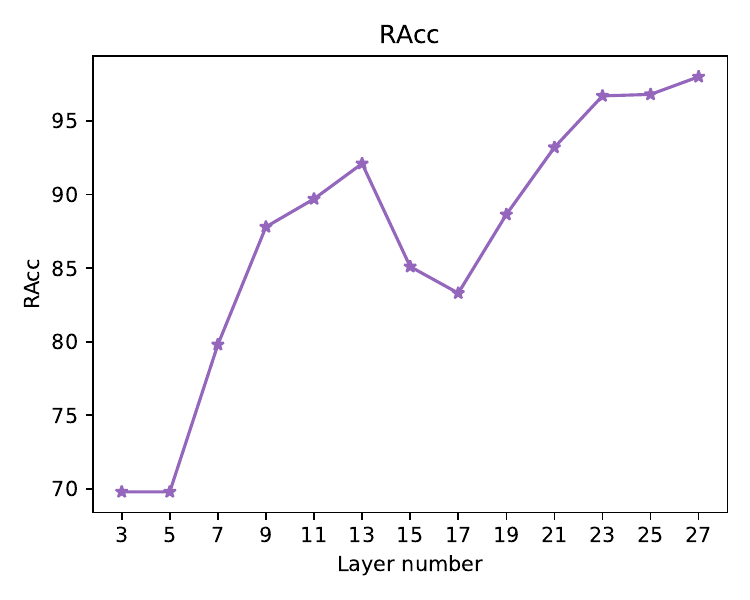}
        \cprotect\caption{Recovery accuracy (RAcc) of the concepts extracted by $\verb|ClassifSAE|$ across the layers of GPT-J fine-tuned on TweetEval Offensive}
        \label{fig:RAcc_gpt_te_offensive}
    \end{subfigure}
    \hfill
    \begin{subfigure}[b]{0.45\textwidth}
        \centering
        \includegraphics[width=\textwidth]{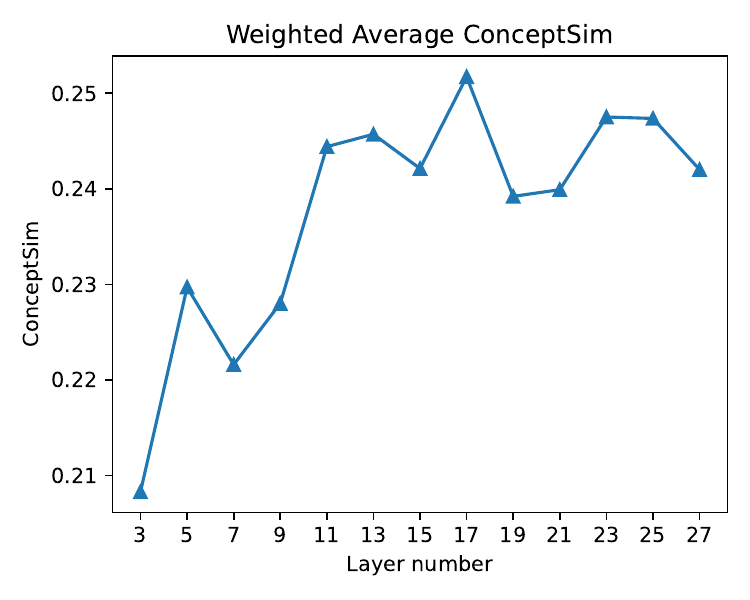}
        \cprotect\caption{Weighted Average $\verb|ConceptSim|$ of the concepts extracted by $\verb|ClassifSAE|$ across the layers of GPT-J fine-tuned on TweetEval Offensive}
        \label{fig:WA_Conceptsim_gpt_te_offensive}
    \end{subfigure}
    \caption{Comparison of the extracted concepts properties across layers of GPT-J fine-tuned on TweetEval Offensive.}
    \label{fig:layerwise_gpt_te_offensive}
\end{figure}

We analyze how the layer depth at which concepts are extracted by \verb|ClassifSAE| affects the metrics of recovery accuracy (RAcc) and weighted average $\verb|ConceptSim|$. Figures \ref{fig:layerwise_pythia_ag_news} and \ref{fig:layerwise_gpt_te_offensive} report these metrics across layers for two configurations: Pythia-1B fine-tuned on AG News and GPT-J fine-tuned on TweetEval Offensive. In each case, the method was trained and evaluated independently at every layer to isolate the representational contribution of depth. Across both settings, we observe a steady decline in recovery accuracy toward earlier layers, with a small dip though in the mid-layers for GPT-J before improvement in the deeper ones. This trend is expected, since the representations in lower layers have not yet integrated complete sentence-level information and the embedding space at these stages is not yet specialized for the downstream classification task. Consequently, our jointly trained classifier has greater difficulty identifying discriminative sparse features to encode in $\textbf{z}_{\text{class}}$. 

For Pythia-1B tuned on AG News, the $\verb|ConceptSim|$ values show a more irregular trend but generally improve in the upper layers, indicating that extracted concepts become more coherent and semantically consistent as abstraction deepens. The slight rise observed in the earliest layers may correspond to stable lexical or syntactic regularities that emerge before semantic specialization. For GPT-J tuned on TweetEval Offensive, the $\verb|ConceptSim|$ trajectory shows greater variability, with local peaks at intermediate depths. These peaks may correspond to a representational transition stage where the model organizes information into semantically coherent concept spaces that are not yet tightly aligned with the output representation, accounting for the temporary decrease in recovery accuracy at similar depths. Overall though, $\verb|ConceptSim|$ values also improve in the upper layers.

\begin{figure*}[htbp]
  \centering

  \begin{subfigure}{\textwidth}
    \centering
    \begin{subfigure}[b]{0.32\textwidth}
      \includegraphics[width=\linewidth]{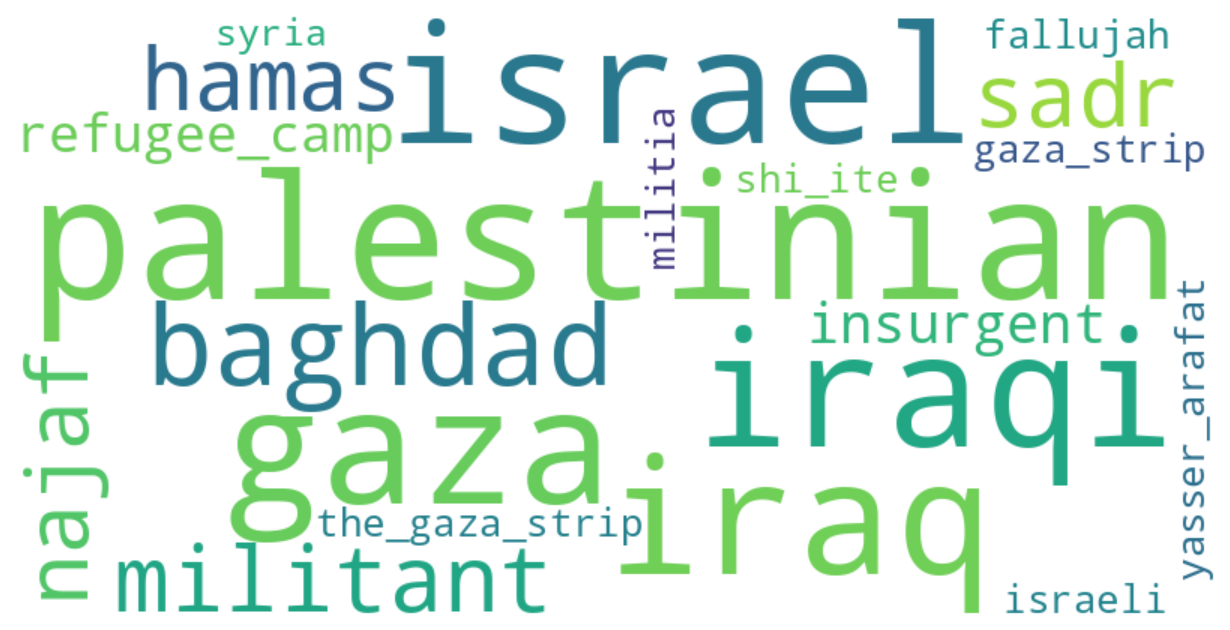}
      \subcaption{World \\ Conflict – Middle East}
    \end{subfigure}
    \hfill
    \begin{subfigure}[b]{0.32\textwidth}
      \includegraphics[width=\linewidth]{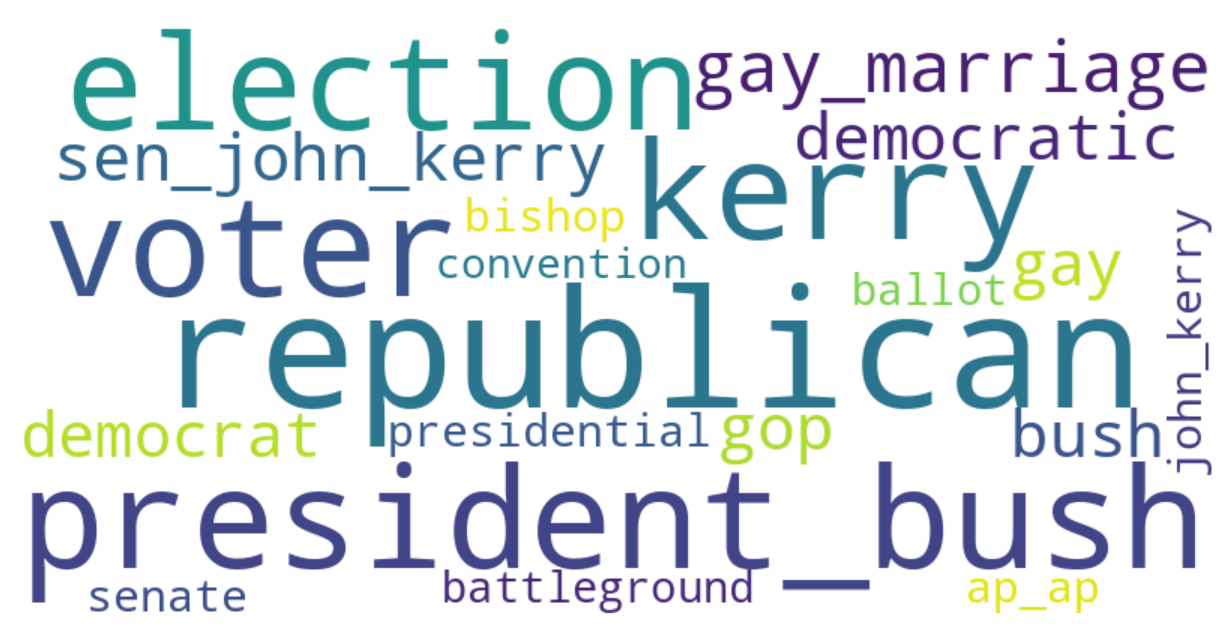}
      \subcaption{World \\ Election - Social issues}
    \end{subfigure}
    \hfill
    \begin{subfigure}[b]{0.32\textwidth}
      \includegraphics[width=\linewidth]{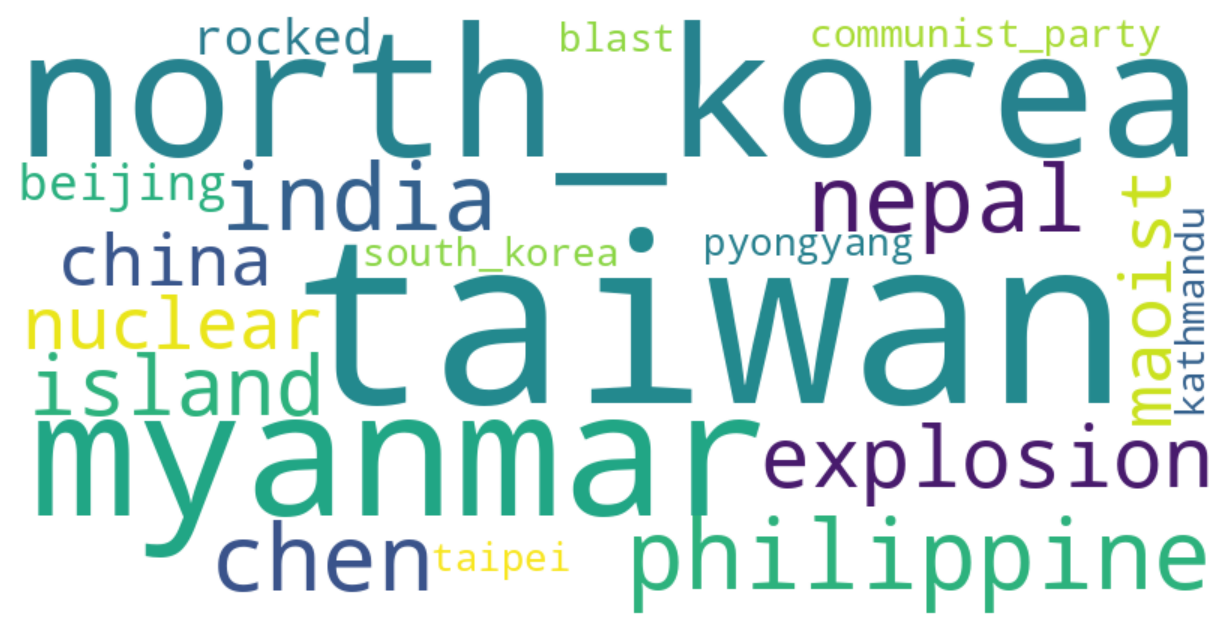}
      \subcaption{World \\  Asia – Tensions}
    \end{subfigure}
  \end{subfigure}

  \vspace{2em}

  \begin{subfigure}{\textwidth}
    \centering
    \begin{subfigure}[b]{0.32\textwidth}
      \includegraphics[width=\linewidth]{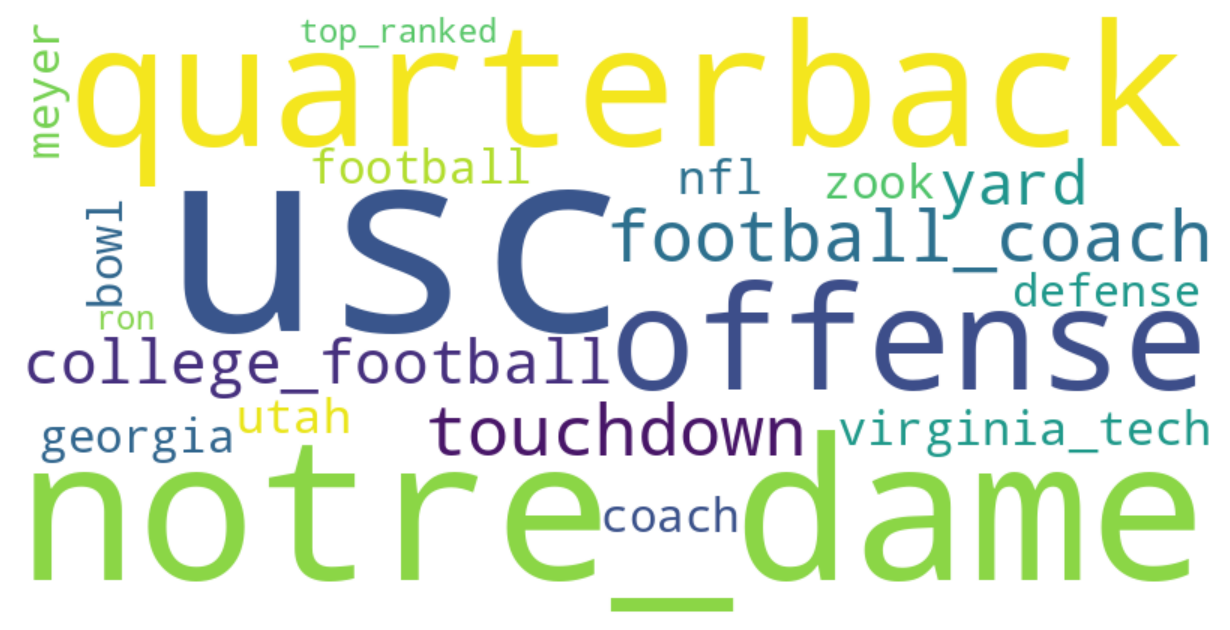}
      \subcaption{Sport \\ College – American Football}
    \end{subfigure}
    \hfill
    \begin{subfigure}[b]{0.32\textwidth}
      \includegraphics[width=\linewidth]{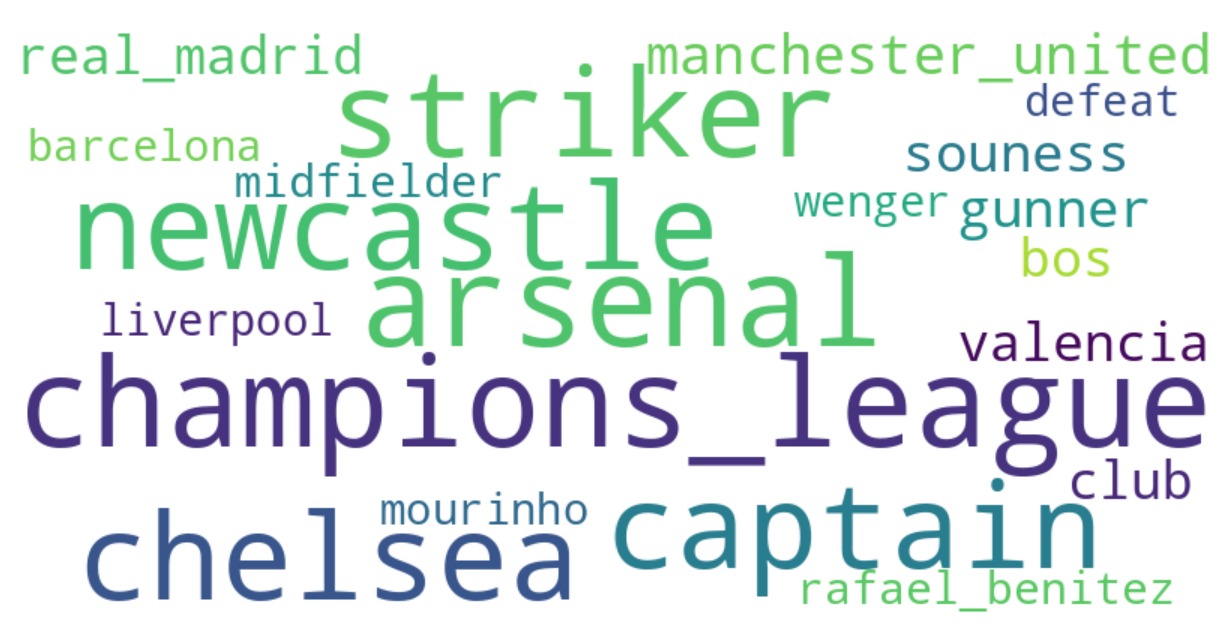}
      \subcaption{Sport \\  Football – Europe}
    \end{subfigure}
    \hfill
    \begin{subfigure}[b]{0.32\textwidth}
      \includegraphics[width=\linewidth]{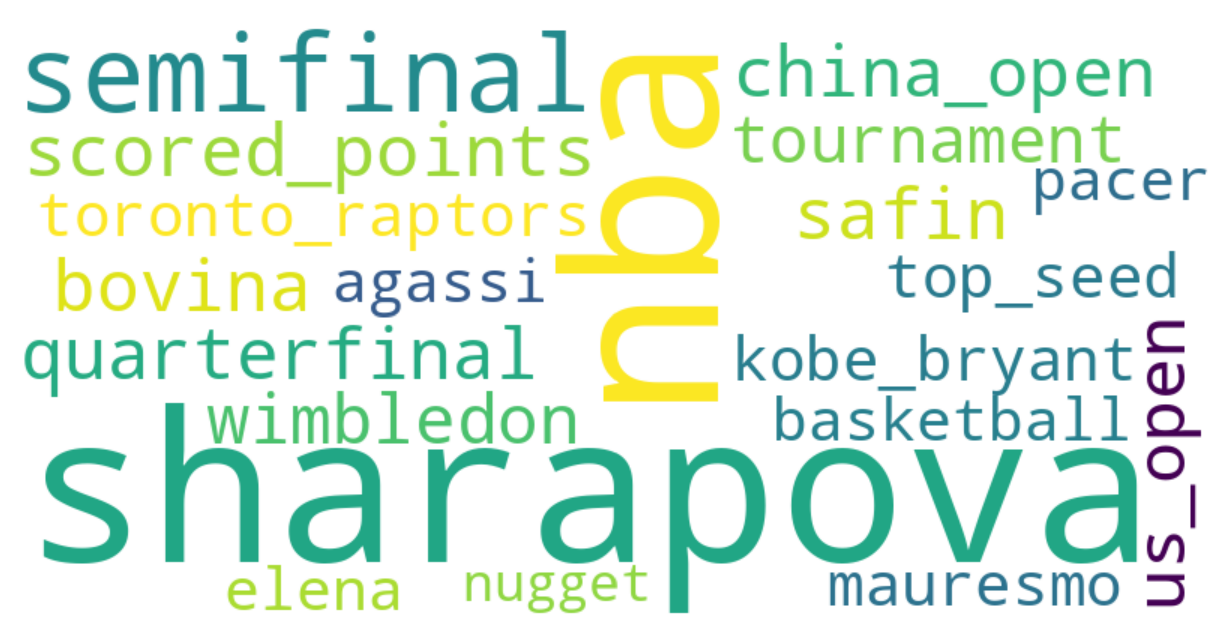}
      \subcaption{Sport \\ Tennis - Basketball}
    \end{subfigure}
  \end{subfigure}

  \vspace{2em}

\begin{subfigure}{\textwidth}
    \centering
    \begin{subfigure}[b]{0.24\textwidth}
      \includegraphics[width=\linewidth]{Figures/concepts_wordclouds/concepts_agnews/Concept_7_Category_Business.pdf}
      \subcaption{Business \\ Airlines - bankruptcy}
    \end{subfigure}
    \hfill
    \begin{subfigure}[b]{0.24\textwidth}
      \includegraphics[width=\linewidth]{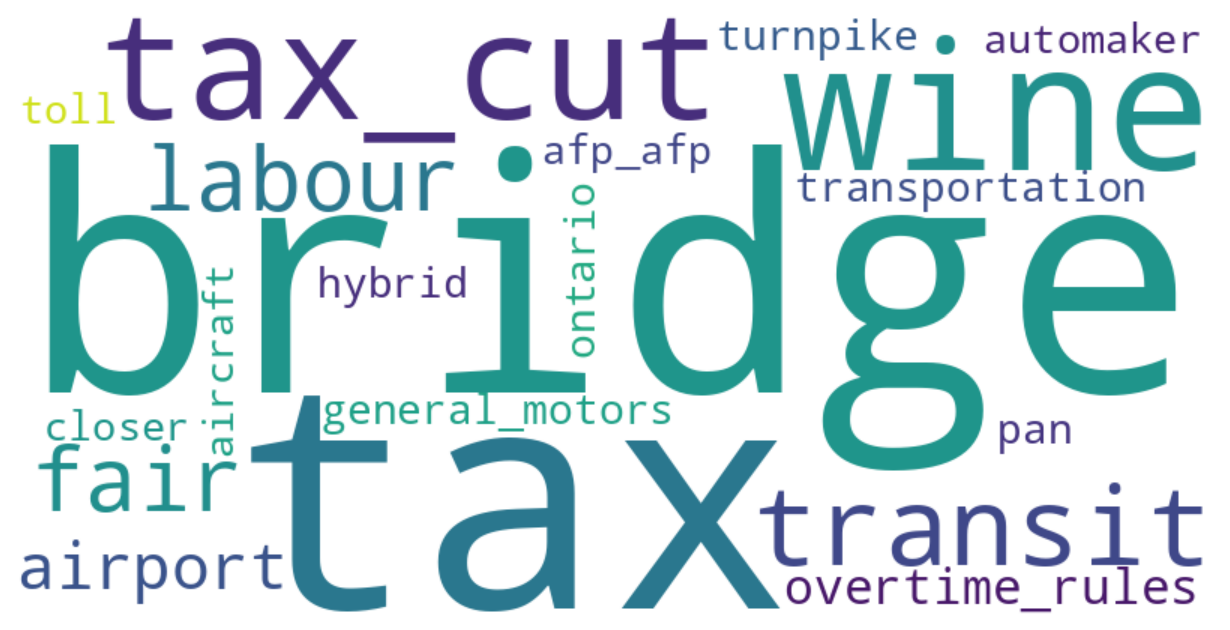}
      \subcaption{Business \\ Infrastructure – Policy}
    \end{subfigure}
    \hfill
    \begin{subfigure}[b]{0.24\textwidth}
      \includegraphics[width=\linewidth]{Figures/concepts_wordclouds/concepts_agnews/Concept_15_Category_Business.pdf}
      \subcaption{Business \\ Currency  - Trade}
    \end{subfigure}
    \hfill
    \begin{subfigure}[b]{0.24\textwidth}
      \includegraphics[width=\linewidth]{Figures/concepts_wordclouds/concepts_agnews/Concept_16_Category_Business.pdf}
      \subcaption{Business \\ Tech – Corporations}
      \end{subfigure}
    
  \end{subfigure}

  \vspace{2em}

  \begin{subfigure}{\textwidth}
    \centering
    \begin{subfigure}[b]{0.19\textwidth}
      \includegraphics[width=\linewidth]{Figures/concepts_wordclouds/concepts_agnews/Concept_2_Category_SciTech.pdf}
      \subcaption{Sci/tech \\ Science – Nature }
    \end{subfigure}
    \hfill
    \begin{subfigure}[b]{0.19\textwidth}
      \includegraphics[width=\linewidth]{Figures/concepts_wordclouds/concepts_agnews/Concept_3_Category_SciTech.pdf}
      \subcaption{Sci/tech \\ Processors – Servers}
    \end{subfigure}
    \hfill
    \begin{subfigure}[b]{0.19\textwidth}
      \includegraphics[width=\linewidth]{Figures/concepts_wordclouds/concepts_agnews/Concept_9_Category_SciTech.pdf}
      \subcaption{Sci/tech \\ Cybersecurity - Spam}
    \end{subfigure}
    \hfill
    \begin{subfigure}[b]{0.19\textwidth}
      \includegraphics[width=\linewidth]{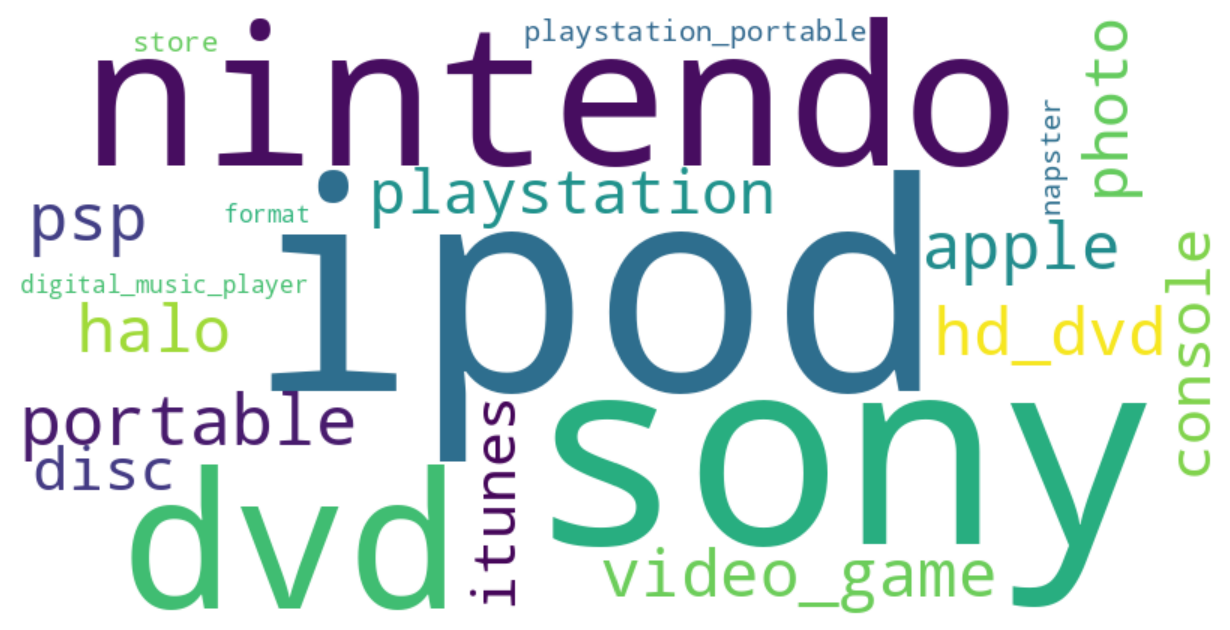}
      \subcaption{Sci/tech \\ Gadgets – Gaming }
    \end{subfigure}
    \hfill
    \begin{subfigure}[b]{0.19\textwidth}
      \includegraphics[width=\linewidth]{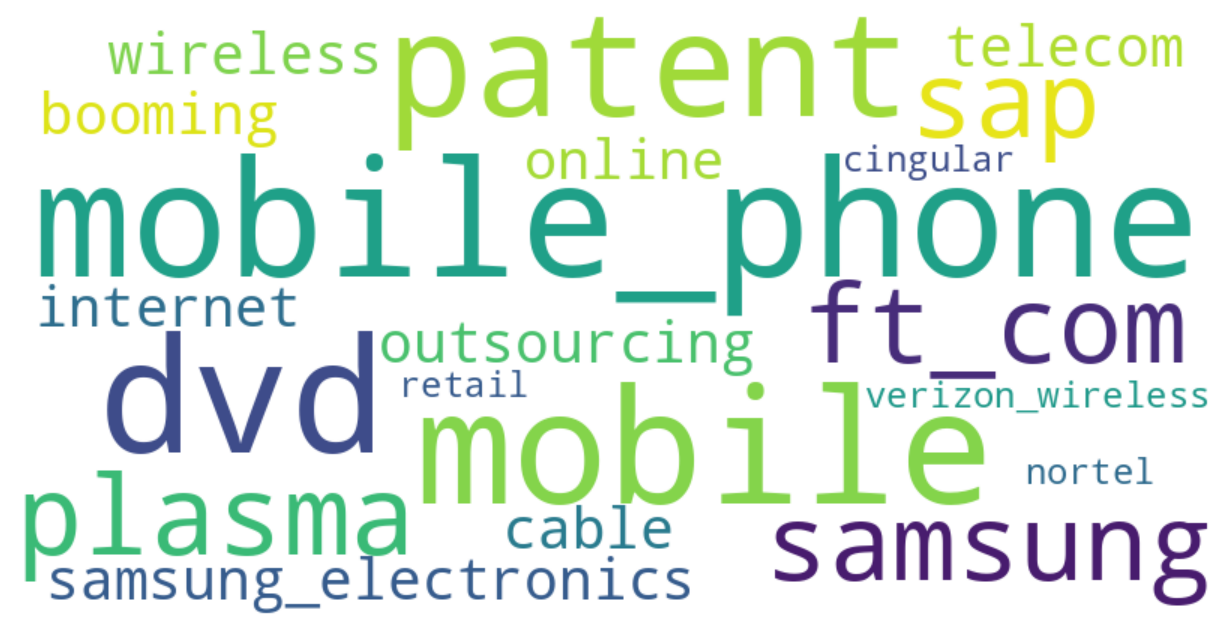}
      \subcaption{Sci/tech \\ Mobile - Telecom}
    \end{subfigure}
  \end{subfigure}

  \cprotect\caption{Examples of concepts discovered by $\verb|ClassifSAE|$ from the internals of GPT-J fine-tuned on AG News.}
  \label{fig:examples_concepts_agnews_gptj}

\end{figure*}

\begin{figure*}[htbp]
  \centering

  \begin{subfigure}{\textwidth}
    \centering
    \begin{subfigure}[b]{0.32\textwidth}
      \includegraphics[width=\linewidth]{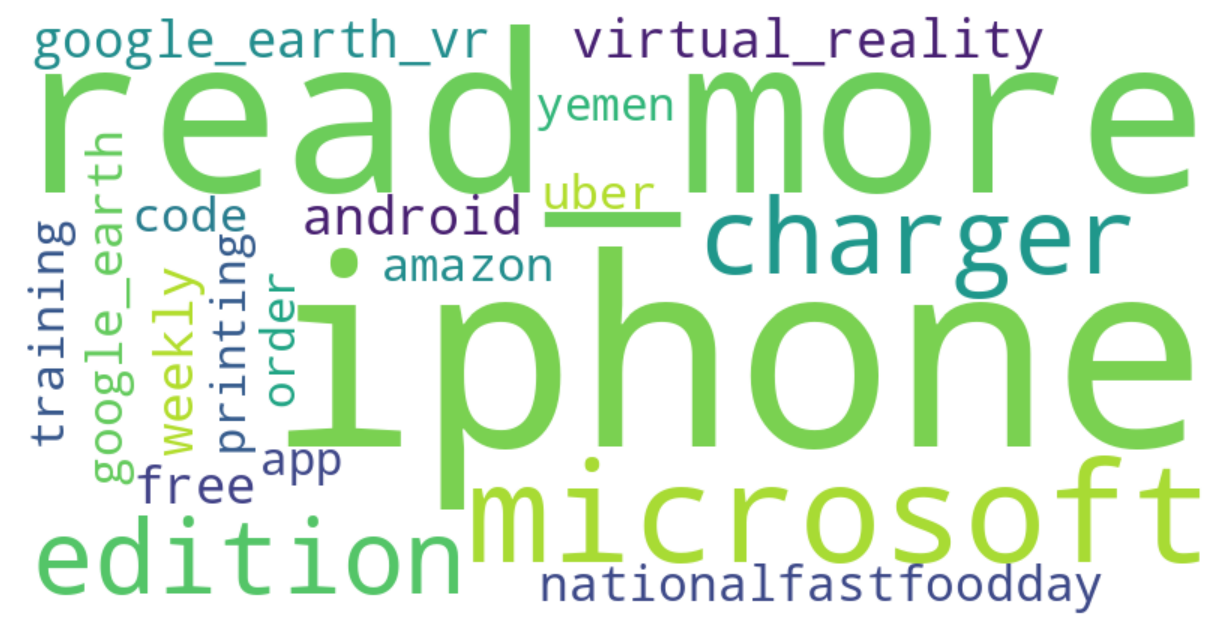}
      \subcaption{Neutral \\ Tech - Innovation }
    \end{subfigure}
    \hfill
    \begin{subfigure}[b]{0.32\textwidth}
      \includegraphics[width=\linewidth]{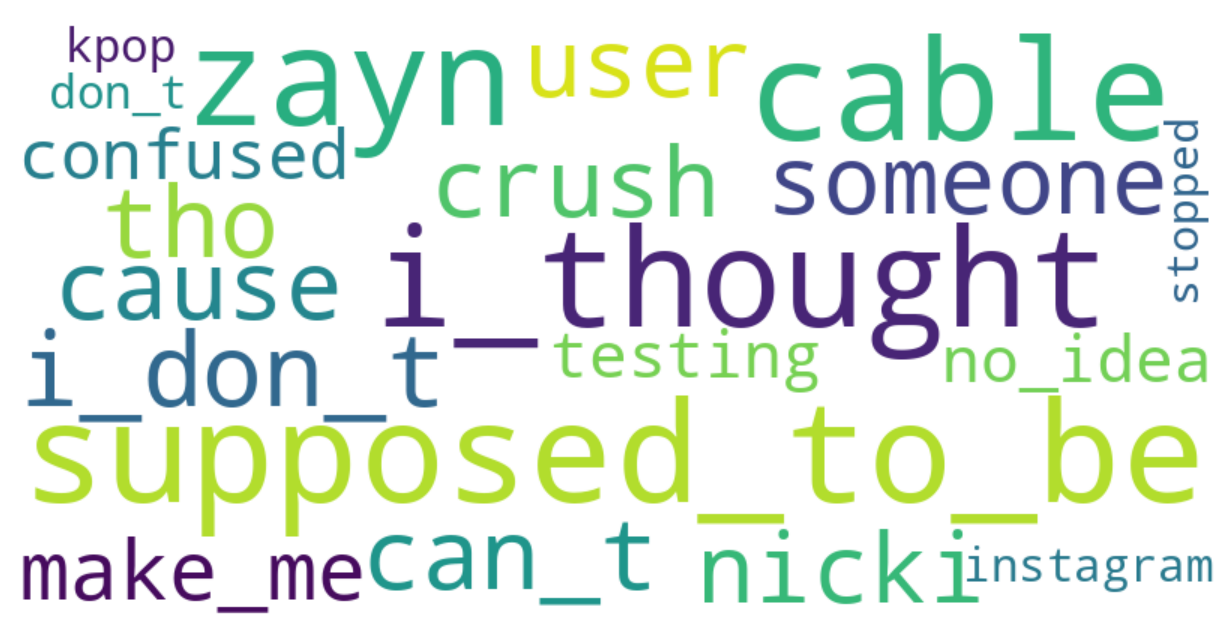}
      \subcaption{Neutral \\ Confused - Unsure}
    \end{subfigure}
    \hfill
    \begin{subfigure}[b]{0.32\textwidth}
      \includegraphics[width=\linewidth]{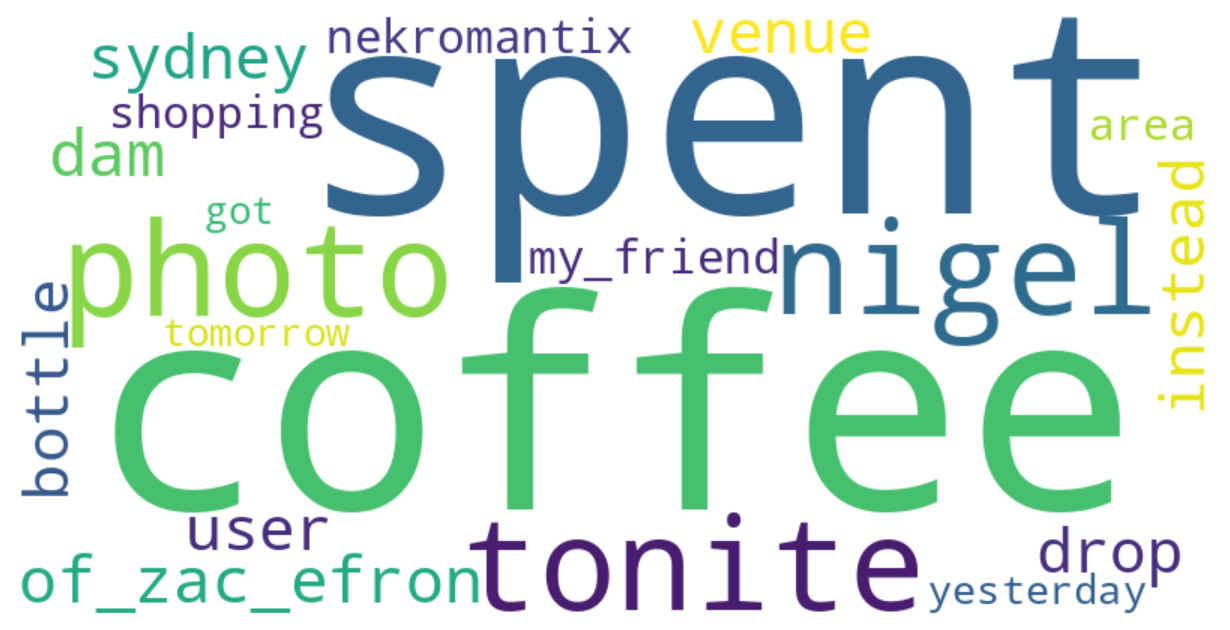}
      \subcaption{World \\ Moments – Daily life}
    \end{subfigure}
  \end{subfigure}

  \vspace{2em}

  \begin{subfigure}{\textwidth}
    \centering
    \begin{subfigure}[b]{0.24\textwidth}
      \includegraphics[width=\linewidth]{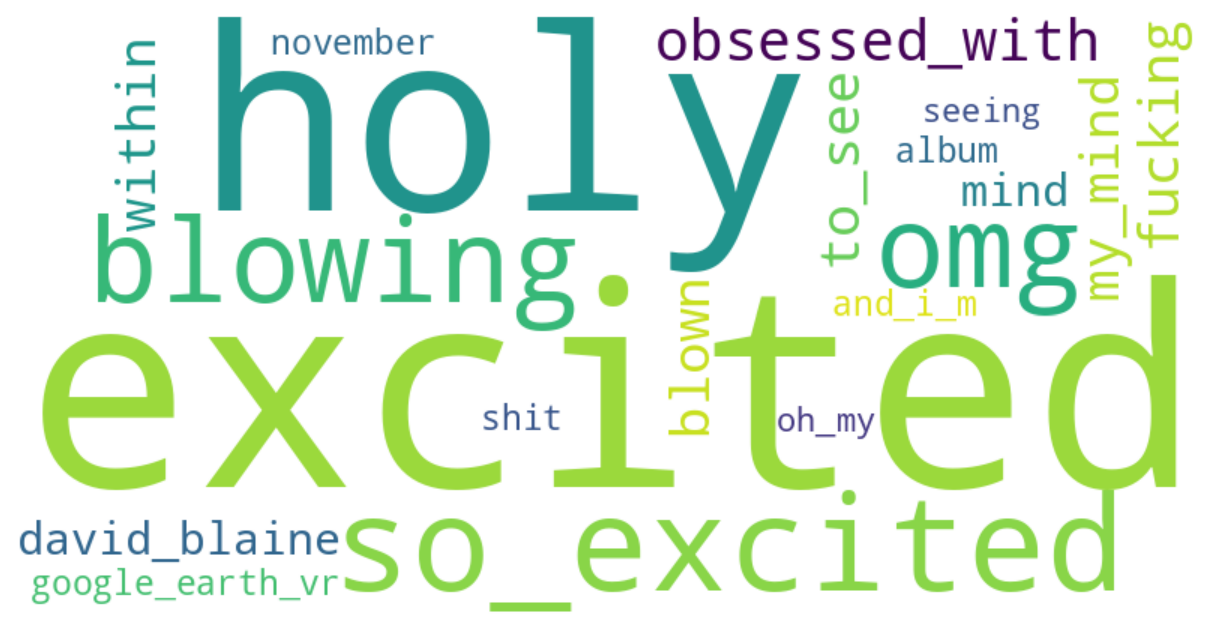}
      \subcaption{Positive \\ Awe – Obsession }
    \end{subfigure}
    \hfill
    \begin{subfigure}[b]{0.24\textwidth}
      \includegraphics[width=\linewidth]{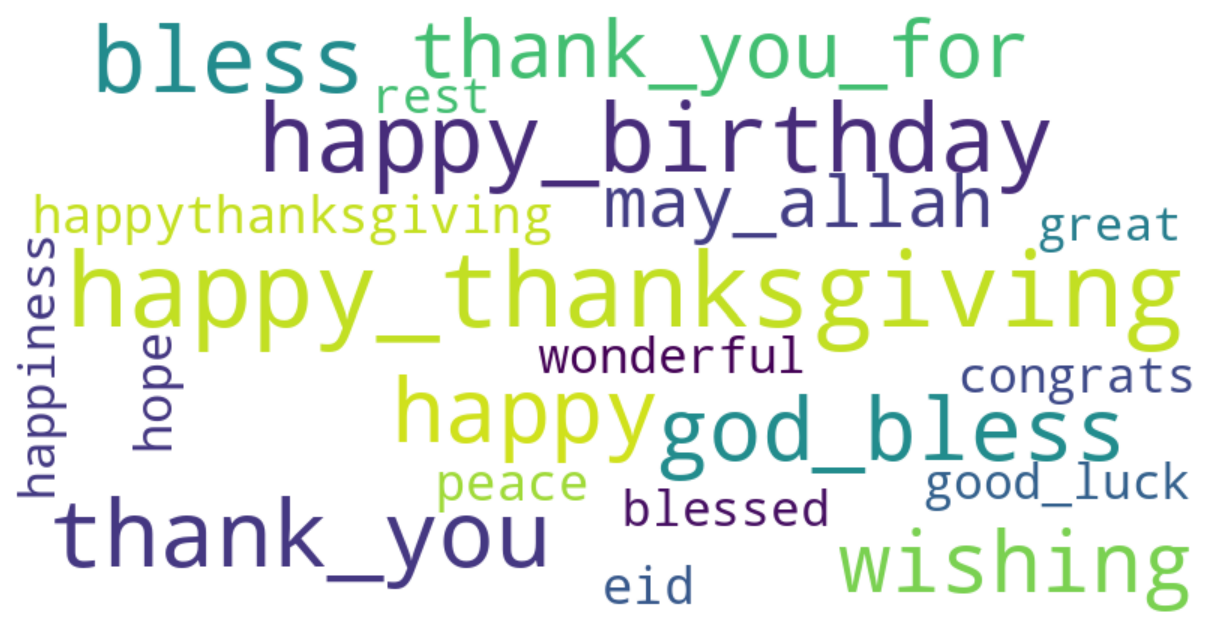}
      \subcaption{Positive \\ Blessings – Gratitude}
    \end{subfigure}
    \hfill
    \begin{subfigure}[b]{0.24\textwidth}
      \includegraphics[width=\linewidth]{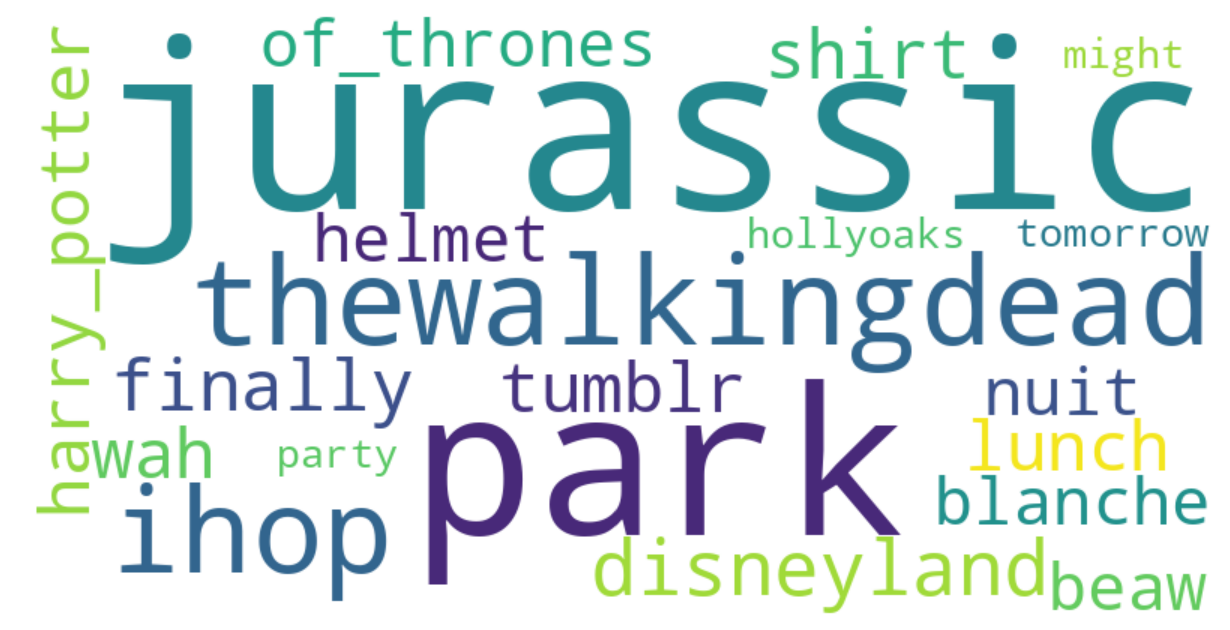}
      \subcaption{Positive \\ Fandom – Anticipation}
    \end{subfigure}
    \hfill
    \begin{subfigure}[b]{0.24\textwidth}
      \includegraphics[width=\linewidth]{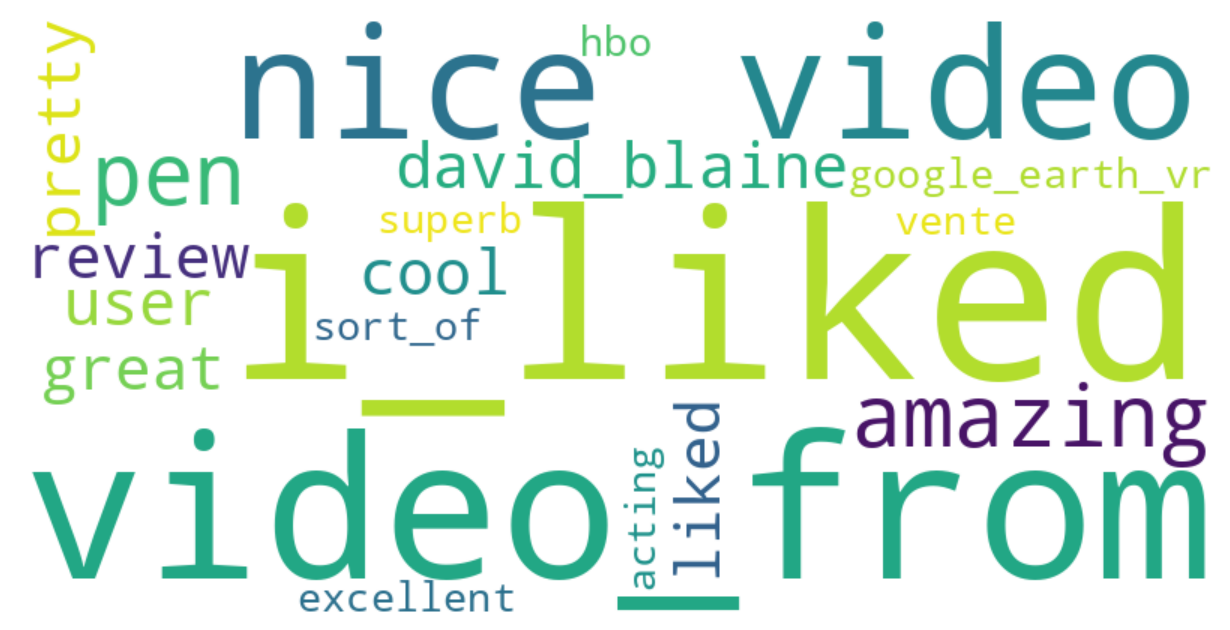}
      \subcaption{Positive \\ Reaction – Content}
    \end{subfigure}
  \end{subfigure}

  \vspace{2em}

\begin{subfigure}{\textwidth}
    \centering
    \begin{subfigure}[b]{0.48\textwidth}
      \includegraphics[width=\linewidth]{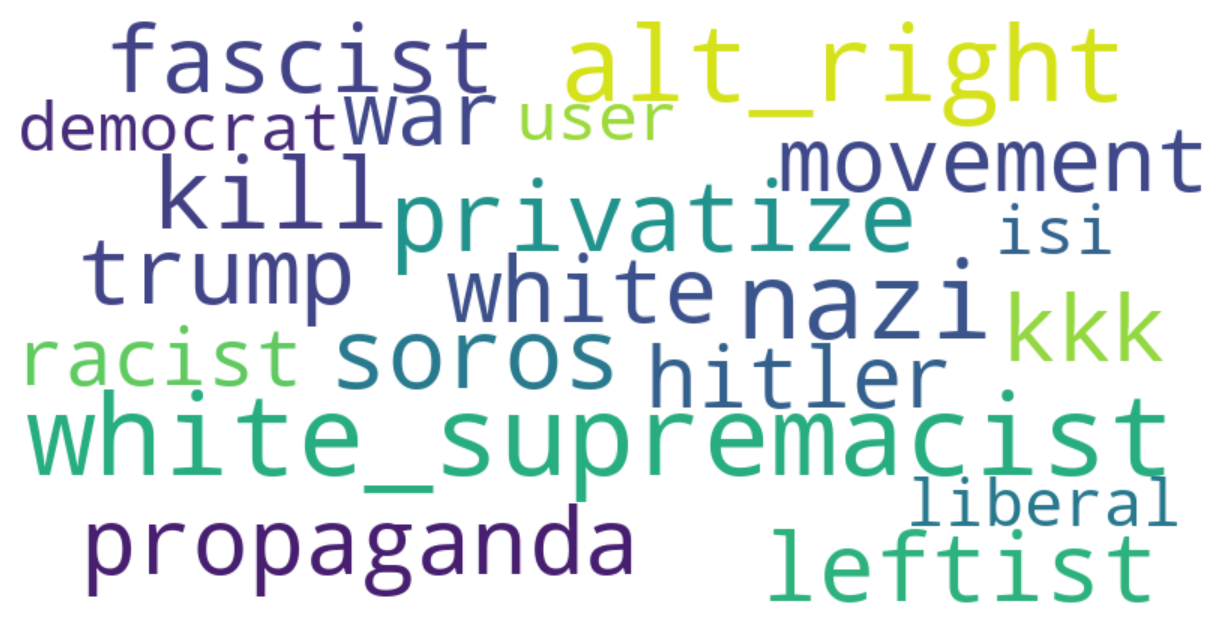}
      \subcaption{Negative \\ Extremism – Ideology}
    \end{subfigure}
    \hfill
    \begin{subfigure}[b]{0.48\textwidth}
      \includegraphics[width=\linewidth]{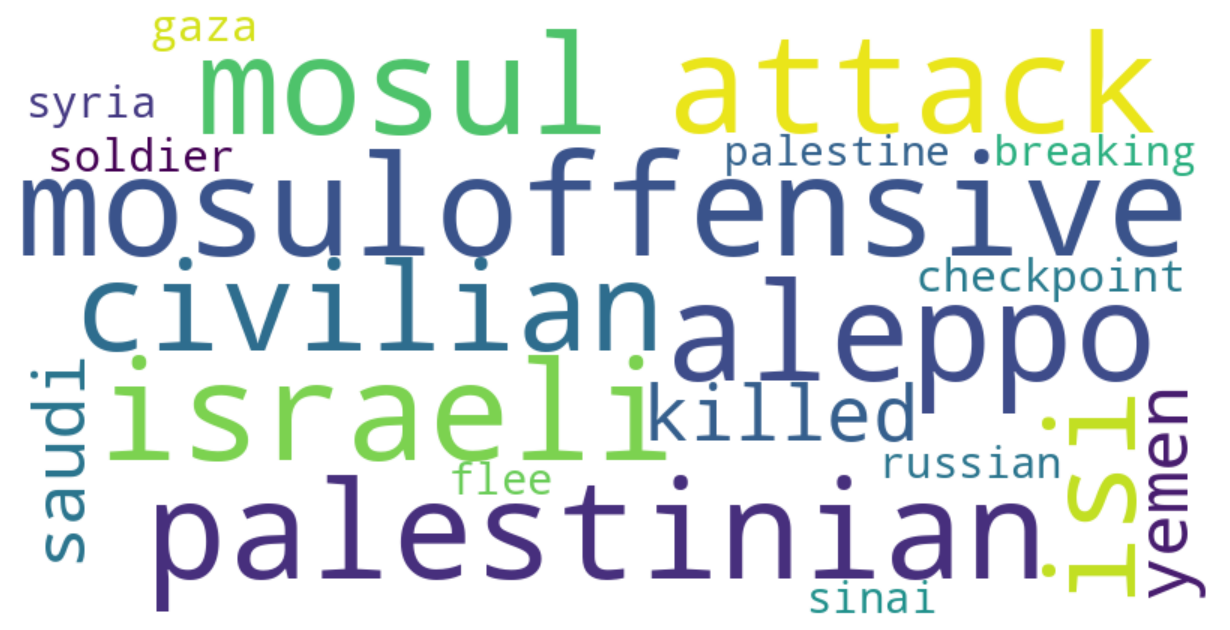}
      \subcaption{Negative \\War – Crisis }
    \end{subfigure}

  \end{subfigure}

  \cprotect\caption{Examples of concepts discovered by $\verb|ClassifSAE|$ from the internals of GPT-J fine-tuned on TweetEval Sentiment}
  \label{fig:example_concept_te_sentiment_gptj}

\end{figure*}

\begin{figure*}[ht!]
  \centering
  \includegraphics[width=\linewidth]{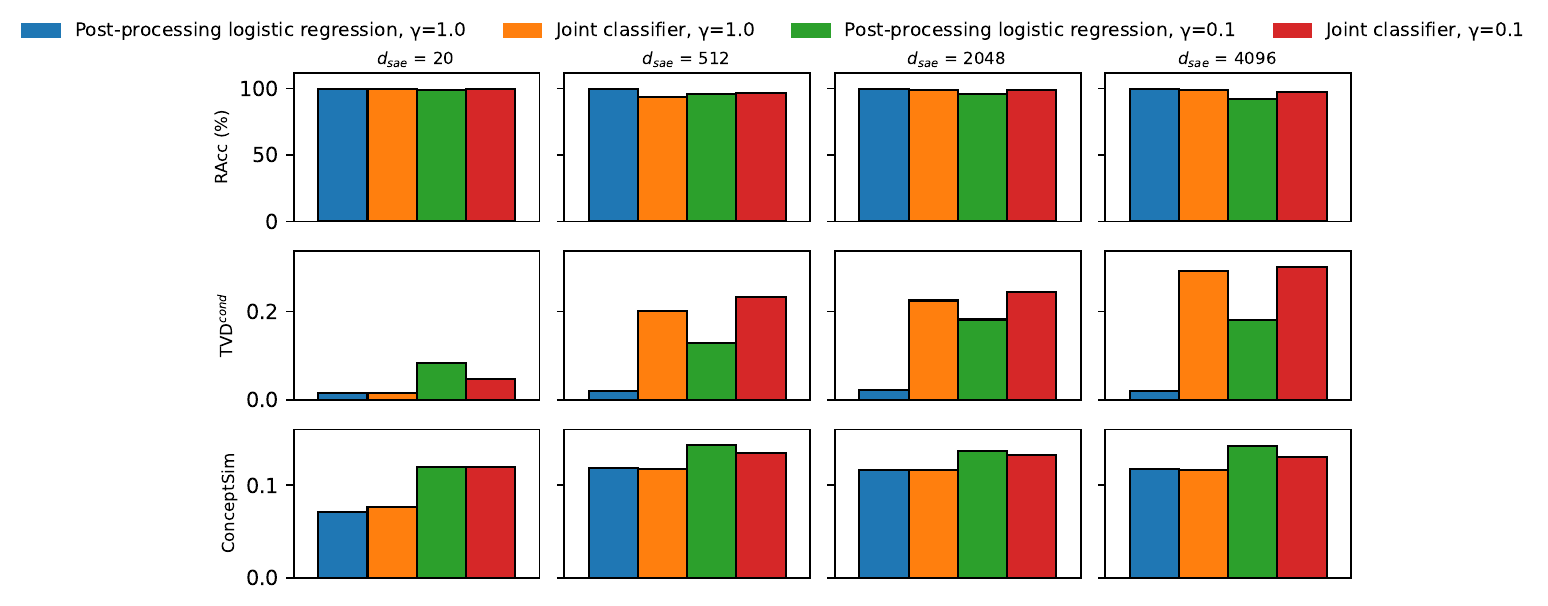}
  \cprotect\caption{Report of the recovery accuracy $\text{RAcc}$ for completeness, $\text{TVD}^{\text{cond}}$ for individual causality and  $\verb|ConceptSim|$ for interpretability (see Sections \ref{sec:existing_metrics} and \ref{sec:new_metrics}) under ablations of the joint classifier and activation rate sparsity mechanism across varying SAE hidden layer sizes. For $d_{sae}$ fixed, we test four configurations: Regular SAE without either component ($\gamma=1$, Logistic regression for the selection of $\textbf{z}_{\text{class}}$ ), SAE with only the activation rate sparsity mechanism ($\gamma=0.1$, Logistic regression for the selection of $\textbf{z}_{\text{class}}$ ), SAE trained with the joint classifier but no activation rate sparsity loss ($\gamma=1$, Joint classifier) and $\verb|ClassifSAE|$ ($\gamma=0.1$, Joint classifier). $\text{TVD}^{\text{cond}}$ increases significantly with the inclusion of a learned classifier while $\verb|ConceptSim|$ mostly benefits from the activation rate sparsity loss. We note that while the activation rate sparsity loss enforcement alone maximizes $\verb|ConceptSim|$, the simultaneous integration of the two components produces the best overall trade-off across the investigated three metrics. Experiments are carried out on residual-stream activations taken at the penultimate layer of Pythia-1B fined-tuned on AG News.}
  \label{fig:ablation_studies}
\end{figure*}

\begin{table*}[ht!]
\centering
\scriptsize                          
\setlength{\tabcolsep}{2.8pt}        
\renewcommand{\arraystretch}{1.5}
\begin{tabular}{lll rrrrrrr}
\hline\hline
\multicolumn{3}{c}{ } &
\multicolumn{7}{c}{\textbf{Classifier Model}}\\
\cline{3-10}
\multicolumn{3}{c}{} & 
BERT & DeBERTa‑v3 & Pythia-410M & Pythia-1B & GPT‑J & Mistral‑Inst. & Llama‑Inst.  \\
\textbf{Dataset} & \textbf{Method} & \textbf{Metric} &
\multicolumn{7}{c}{}\\
\hline

& ICA & RAcc (\%,$\uparrow$) & 99.80  & 99.98 & 99.43 & 99.60 & 99.68 & 99.05 & 96.34\\
        & & $\Delta f^{\text{cond}}$ (\%,$\uparrow$) & 0.57 & 1.98 & 2.47 & 2.63 & 2.24 & 1.76 & 3.33\\
        & & $\verb|ConceptSim|$ ($\uparrow$)  & 0.0627 & 0.0612 & 0.0609 & 0.0607 & 0.0609 & 0.0662 & 0.0616\\
        & & Avg Act. rate (\%,$\downarrow$) & 100 & 100 & 100 & 100 & 100 & 100 & 100\\
        \cline{2-10}
& ConceptSHAP & RAcc (\%,$\uparrow$)   & 99.58  & 99.84 & 97.27 & 98.55 & 98.17 & 96.18 & 88.68\\
        & & $\Delta f^{\text{cond}}$ (\%,$\uparrow$)  &  11.35 & 0.76  & 9.98 & 13.34 & 10.06 & 5.54 & 1.19\\
        & & $\verb|ConceptSim|$ ($\uparrow$)  & 0.1226 & 0.1250 & 0.1259 & 0.1260 & 0.1323 & 0.1312 & 0.1430\\
        & & Avg Act. rate (\%,$\downarrow$) & 26.32   &  25.71 & 24.38 & 24.56 & 24.06 & 23.28 & 20.60\\
        \cline{2-10}
AG News & HI-Concept & RAcc (\%,$\uparrow$)  & 99.72  & 99.91 & 99.63 & 99.72 & 99.82 & 99.67 & 98.97\\
        & & $\Delta f^{\text{cond}}$ (\%,$\uparrow$)  & 53.03 & 81.32 & 7.09 & 20.47 & 33.56 & 31.05 & 12.81\\
        & & $\verb|ConceptSim|$ ($\uparrow$)  & 0.1112 & 0.1239 & 0.0949 & 0.0907 & 0.0823 & 0.075 & 0.0852\\
        & & Avg Act. rate (\%,$\downarrow$) & 33.51  & 25.93 & 37.15 & 47.53 & 50.97 & 63.56 & 58.73\\
        \cline{2-10}
& SAE (K=10) & RAcc (\%,$\uparrow$)  & 98.63  & 99.67 & 98.17 & 98.89 & 99.00 & 96.24 & 92.63\\
        & & $\Delta f^{\text{cond}}$ (\%,$\uparrow$)  & 0.92 & 0.27 & 2.32 & 2.87 & 4.54 & 7.76 & 10.59\\
        & & $\verb|ConceptSim|$ ($\uparrow$)  & 0.1297 & 0.1241 & 0.1185 & 0.1264 & 0.099 & 0.0696 & 0.0699\\
        & & Avg  Act. rate (\%,$\downarrow$) & 24.20 & 26.53 & 18.54 & 26.40 & 30.72 & 49.93 & 42.77\\
        \cline{2-10}
 & $\verb|ClassifSAE|$ (K=10) & RAcc (\%,$\uparrow$) & 99.14 &  96.30 & 98.84 & 99.43 & 99.39 & 99.02 & 96.57\\
        & & $\Delta f^{\text{cond}}$ (\%,$\uparrow$)  & 14.27 & 23.63 & 18.67 & 22.10 & 24.32 & 17.86 & 17.53\\ 
        & & $\verb|ConceptSim|$ ($\uparrow$)  & 0.1566 & 0.1371 & 0.1562 & 0.1412 & 0.1648 & 0.1431 & 0.1444\\
        & & Avg  Act. rate (\%,$\downarrow$) & 9.59  & 9.65 & 8.41 & 8.11 & 8.28 & 10.28 & 9.38\\
        \hline\hline

\end{tabular}
\cprotect\caption{Completeness, causality and interpretability metrics (see Sections \ref{sec:existing_metrics} and \ref{sec:new_metrics}) of the concepts learned from different LLM classifiers for the dataset AG News. Prior each task evaluation, all models are fine-tuned at the exception of Mistral-Instruct and Llama-Instruct, which are aligned with the task via soft‑prompt tuning. $\Delta f^{cond}$ is simply the average of $\Delta f_{\{j\}}^{cond}$. $\verb|ConceptSim|$ is the average of the individual concept scores, weighted as detailed in the Appendix \ref{sec:interpretability_metrics}. All post-hoc methods were configured to search for 20 concepts. Concepts are computed from the sentence‑level hidden state: for decoder‑only models, from the residual stream after the penultimate transformer block and for encoder‑only models, from the layer preceding the classification head. Results are obtained with seed equals to $42$}
\label{tab:table_metrics_agnews}
\end{table*}

\begin{table*}[ht!]
\centering
\scriptsize                          
\setlength{\tabcolsep}{2.8pt}        
\renewcommand{\arraystretch}{1.5}
\begin{tabular}{lll rrrrrrr}
\hline\hline
\multicolumn{3}{c}{ } &
\multicolumn{7}{c}{\textbf{Classifier Model}}\\
\cline{3-10}
\multicolumn{3}{c}{} & 
BERT & DeBERTa‑v3 & Pythia-410M & Pythia-1B & GPT‑J & Mistral‑Inst. & Llama‑Inst.  \\
\textbf{Dataset} & \textbf{Method} & \textbf{Metric} &
\multicolumn{7}{c}{}\\
\hline

& ICA & RAcc (\%,$\uparrow$) & 99.77 & 99.86 & 99.22 & 99.08 & 99.63 & 98.58 & 94.05\\
        & & $\Delta f^{\text{cond}}$ (\%,$\uparrow$) & 1.81 & 0.84 & 3.10 & 4.14 & 2.72 & 3.22 & 1.93\\
        & & $\verb|ConceptSim|$ ($\uparrow$)  & 0.2059 & 0.2076 & 0.2091 & 0.2044 & 0.2044 & 0.2130 & 0.2127\\
        & & Avg Act. rate (\%,$\downarrow$) & 100 &  100 & 100 & 100 & 100 & 100 & 100\\
        \cline{2-10}
& ConceptSHAP & RAcc (\%,$\uparrow$)   & 97.00 & 97.80 & 95.55 & 82.73 & 95.19 & 89.88 & 92.08\\
        & & $\Delta f^{\text{cond}}$ (\%,$\uparrow$)  &  0.06 & 0.29 & 0.26 & 1.03 & 0.23 & 0.02 & 0.03\\
        & & $\verb|ConceptSim|$ ($\uparrow$)  & 0.2049 & 0.2015 & 0.2062 & 0.2241 & 0.2128 & 0.2433 & 0.2279\\
        & & Avg Act. rate (\%,$\downarrow$) & 44.78 & 46.83 & 44.04 & 24.95 & 36.46 & 10.77 & 20.12\\
        \cline{2-10}
TE Offensive & HI-Concept & RAcc (\%,$\uparrow$)  & 99.54 & 99.77 & 99.13 & 99.63 & 98.99 & 99.16 & 98.90\\
        & & $\Delta f^{\text{cond}}$ (\%,$\uparrow$) & 1.66 & 11.28 & 8.47 & 12.20 & 2.67 & 8.45 & 1.79\\
        & & $\verb|ConceptSim|$ ($\uparrow$)  & 0.2073 & 0.2027 & 0.2048 & 0.2051 & 0.2049 & 0.2062 & 0.2045\\
        & & Avg Act. rate (\%,$\downarrow$) & 38.92 & 53.10 & 53.00 & 36.24 & 57.98 & 64.37 & 64.24\\
        \cline{2-10}
 & SAE (K=10) & RAcc (\%,$\uparrow$)  & 98.07 & 99.08 & 97.53 & 97.57 & 85.89 & 88.28 & 94.30\\
        & & $\Delta f^{\text{cond}}$ (\%,$\uparrow$)  & 0.72 & 0.49 & 1.98 & 8.41 & 17.45 & 6.65 & 18.93\\
        & & $\verb|ConceptSim|$ ($\uparrow$)  & 0.2199 & 0.2079 & 0.2071 & 0.2024 & 0.2034 & 0.2097 & 0.2073\\
        & & Avg  Act. rate (\%,$\downarrow$) & 13.22 & 34.79 & 12.52 & 10.14 & 43.05 & 34.40 & 35.96\\
        \cline{2-10}
 & $\verb|ClassifSAE|$ (K=10) & RAcc (\%,$\uparrow$) & 97.76 & 99.35 & 97.66 & 94.10 & 93.17 & 96.75 & 96.79\\
        & & $\Delta f^{\text{cond}}$ (\%,$\uparrow$)  & 10.34 & 12.40 & 9.70 & 4.69 & 9.60 & 7.34 & 10.75\\ 
        & & $\verb|ConceptSim|$ ($\uparrow$)  & 0.2488 & 0.2342 & 0.2264 & 0.2328 & 0.2379 & 0.2489 & 0.2224\\
        & & Avg  Act. rate (\%,$\downarrow$) & 9.13 & 9.04 & 6.77 & 9.88 & 10.19 & 6.85 & 8.77\\
        \hline\hline

\end{tabular}
\cprotect\caption{Completeness, causality and interpretability metrics (see Sections \ref{sec:existing_metrics} and \ref{sec:new_metrics}) of the concepts learned from different LLM classifiers for the dataset TweetEval Offensive. Prior each task evaluation, all models are fine-tuned at the exception of Mistral-Instruct and Llama-Instruct, which are aligned with the task via soft‑prompt tuning. $\Delta f^{cond}$ is simply the averages of $\Delta f_{\{j\}}^{cond}$. $\verb|ConceptSim|$ is the average of the individual concept scores, weighted as detailed in the Appendix \ref{sec:interpretability_metrics}. All post-hoc methods were configured to search for 20 concepts. Concepts are computed from the sentence‑level hidden state: for decoder‑only models, from the residual stream after the penultimate transformer block and for encoder‑only models, from the layer preceding the classification head. Results are obtained with seed equals to $42$.}
\label{tab:table_metrics_te_offensive}
\end{table*}

\begin{table*}[ht!]
\centering
\scriptsize                          
\setlength{\tabcolsep}{2.8pt}        
\renewcommand{\arraystretch}{1.5}
\begin{tabular}{lll rrrrrrr}
\hline\hline
\multicolumn{3}{c}{ } &
\multicolumn{7}{c}{\textbf{Classifier Model}}\\
\cline{3-10}
\multicolumn{3}{c}{} & 
BERT & DeBERTa‑v3 & Pythia-410M & Pythia-1B & GPT‑J & Mistral‑Inst. & Llama‑Inst.  \\
\textbf{Dataset} & \textbf{Method} & \textbf{Metric} &
\multicolumn{7}{c}{}\\
\hline

& ICA & RAcc (\%,$\uparrow$) & 98.58 & 99.37 & 98.51 & 97.99 & 98.73 & 92.12 & 84.40\\
        & & $\Delta f^{\text{cond}}$ (\%,$\uparrow$) & 4.78 & 4.31 & 4.03 & 5.36 & 5.08 & 5.46 & 7.22\\
        & & $\verb|ConceptSim|$ ($\uparrow$)  & 0.1146 & 0.1154 & 0.1182 & 0.1170 & 0.1161 & 0.1216 & 0.1219\\
        & & Avg Act. rate (\%,$\downarrow$) & 100 &  100 & 100 & 100 & 100 & 100 & 100\\
        \cline{2-10}
& ConceptSHAP & RAcc (\%,$\uparrow$)   & 89.70 & 95.10 & 91.65 & 95.20 & 91.47 & 94.62 & 81.39\\
        & & $\Delta f^{\text{cond}}$ (\%,$\uparrow$)  & 11.81 & 1.10 & 5.81 & 1.21 & 1.30 & 7.91 & 7.14\\
        & & $\verb|ConceptSim|$ ($\uparrow$)  & 0.1453 & 0.1297 & 0.1171 & 0.1332 & 0.1324 & 0.1516 & 0.1565\\
        & & Avg Act. rate (\%,$\downarrow$) & 33.57 & 31.39 & 30.85 & 28.45 & 25.19 & 28.01 & 17.11\\
        \cline{2-10}
TE Sentiment & HI-Concept & RAcc (\%,$\uparrow$)  & 98.32 & 99.00& 98.91 & 98.63 & 99.03 & 98.41 & 95.62\\
        & & $\Delta f^{\text{cond}}$ (\%,$\uparrow$) & 25.43 & 18.69 & 12.56 & 9.02 & 9.23 & 8.80 & 14.12\\
        & & $\verb|ConceptSim|$ ($\uparrow$)  & 0.1225 & 0.1237  & 0.1246 & 0.12137 & 0.1206 & 0.1217 & 0.1259\\
        & & Avg Act. rate (\%,$\downarrow$) & 39.53 & 38.04 & 46.79 & 46.31 & 53.62 & 62.34 & 50.48\\
        \cline{2-10}
 & SAE (K=10) & RAcc (\%,$\uparrow$)  & 93.85 & 95.45 & 87.05 & 96.26 & 94.11 & 88.80 & 70.56\\
        & & $\Delta f^{\text{cond}}$ (\%,$\uparrow$)  & 6.05 & 3.04 & 3.38 & 5.42 & 7.80 & 16.48 & 28.84\\
        & & $\verb|ConceptSim|$ ($\uparrow$)  & 0.1465 & 0.1344 & 0.1518 & 0.1293& 0.1229 & 0.1174 & 0.1263\\
        & & Avg  Act. rate (\%,$\downarrow$) & 14.03 & 27.62 & 15.49 & 15.38 & 39.18 & 44.56 & 45.13\\
        \cline{2-10}
 & $\verb|ClassifSAE|$ (K=10) & RAcc (\%,$\uparrow$) & 94.20 & 97.89 & 95.90 & 96.21 & 95.91 & 91.44 & 93.23\\
        & & $\Delta f^{\text{cond}}$ (\%,$\uparrow$)  & 10.09 & 15.90 & 13.04 & 15.06 & 12.16 & 14.26 & 19.72\\ 
        & & $\verb|ConceptSim|$ ($\uparrow$)  & 0.1577 & 0.1494 & 0.1458 & 0.1646 & 0.1574 & 0.1550 & 0.1500\\
        & & Avg  Act. rate (\%,$\downarrow$) & 9.62 & 8.89 & 8.96 & 8.30 & 8.69 & 8.95 & 15.27\\
        \hline\hline

\end{tabular}
\cprotect\caption{Completeness, causality and interpretability metrics (see Sections \ref{sec:existing_metrics} and \ref{sec:new_metrics}) of the concepts learned from different LLM classifiers for the dataset TweetEval Sentiment. Prior each task evaluation, all models are fine-tuned at the exception of Mistral-Instruct and Llama-Instruct, which are aligned with the task via soft‑prompt tuning. $\Delta f^{cond}$ is simply the averages of $\Delta f_{\{j\}}^{cond}$. $\verb|ConceptSim|$ is the average of the individual concept scores, weighted as detailed in the Appendix \ref{sec:interpretability_metrics}. All post-hoc methods were configured to search for 20 concepts. Concepts are computed from the sentence‑level hidden state: for decoder‑only models, from the residual stream after the penultimate transformer block and for encoder‑only models, from the layer preceding the classification head. Results are obtained with seed equals to $42$.}
\label{tab:table_metrics_te_sentiment}
\end{table*}

\begin{table*}[ht!]
\centering
\scriptsize                          
\setlength{\tabcolsep}{2.8pt}        

\renewcommand{\arraystretch}{1.5}
\begin{tabular}{lll rrrrrrr}
\hline\hline
\multicolumn{3}{c}{ } &
\multicolumn{7}{c}{\textbf{Classifier Model}}\\
\cline{3-10}
\multicolumn{3}{c}{} & 
BERT & DeBERTa‑v3 & Pythia-410M & Pythia-1B & GPT‑J & Mistral‑Inst. & Llama‑Inst.  \\
\textbf{Dataset} & \textbf{Method} & \textbf{Metric} &
\multicolumn{7}{c}{}\\
\hline

& ICA & RAcc (\%,$\uparrow$) & 99.94 & 99.98 & 99.98 & 99.92 & 99.96 & 97.21 & 99.50 \\
        & & $\Delta f^{\text{cond}}$ (\%,$\uparrow$) & 2.78 & 2.57  & 2.06 & 2.37 & 2.29 & 3.72 & 2.51 \\
        & & $\verb|ConceptSim|$ ($\uparrow$)  & 0.2988 & 0.2956 & 0.2963 & 0.3012 & 0.2997 & 0.3005 &0.3005\\
        & & Avg Act. rate (\%,$\downarrow$) & 100 & 100 & 100 & 100 & 100 & 100 & 100\\
        \cline{2-10}
& ConceptSHAP & RAcc (\%,$\uparrow$)   & 99.74 & 99.40 & 99.48 & 99.75 & 99.74 & 93.34 & 94.52\\
        & & $\Delta f^{\text{cond}}$ (\%,$\uparrow$)  & 3.30 & 0.00 & 0.00 & 0.21 & 0.00 & 0.43 & 1.16\\
        & & $\verb|ConceptSim|$ ($\uparrow$)  & 0.3012 & 0.3031 & 0.3030 & 0.3051 & 0.3046 & 0.3083 & 0.3109\\
        & & Avg Act. rate (\%,$\downarrow$) & 45.70 & 42.85 & 49.39 & 48.71 & 44.00 & 28.02 & 42.67\\
        \cline{2-10}
IMDB & HI-Concept & RAcc (\%,$\uparrow$)  & 99.77 & 99.97 & 99.94 & 99.90 & 99.94 & 98.69 & 99.58\\
        & & $\Delta f^{\text{cond}}$ (\%,$\uparrow$) & 38.53 & 24.05 & 39.14 & 25.45 & 59.33 & 17.143 & 0.48\\
        & & $\verb|ConceptSim|$ ($\uparrow$)  & 0.3068 & 0.3049 & 0.3082 & 0.3021 & 0.2925 & 0.3013 & 0.3017\\
        & & Avg Act. rate (\%,$\downarrow$) & 44.69 & 47.71 & 50.15 & 47.46 & 73.88 & 57.05 & 71.85\\
        \cline{2-10}
& SAE (K=10) & RAcc (\%,$\uparrow$)  & 99.01 & 99.68 & 99.42 & 99.60 & 98.85 & 91.79 & 98.12\\
        & & $\Delta f^{\text{cond}}$ (\%,$\uparrow$)  & 0.47 & 9.94 & 0.24 & 0.44 & 6.12 & 15.54 & 6.18\\
        & & $\verb|ConceptSim|$ ($\uparrow$)  & 0.3096 & 0.3139 & 0.3099 & 0.3043 & 0.3101 & 0.3042 & 0.3003\\
        & & Avg  Act. rate (\%,$\downarrow$) & 30.12 & 27.84 & 27.06 & 31.24 & 29.02 & 49.52 & 43.76\\
        \cline{2-10}
 & $\verb|ClassifSAE|$ (K=10) & RAcc (\%,$\uparrow$) & 98.88 & 99.87 & 95.57 & 94.02 & 99.78 & 96.51 & 98.84\\
        & & $\Delta f^{\text{cond}}$ (\%,$\uparrow$)  & 4.76 & 3.92 & 16.53 & 17.85 & 2.05 & 5.76 & 22.26 \\ 
        & & $\verb|ConceptSim|$ ($\uparrow$)  & 0.3045 & 0.3036 & 0.3031 & 0.3052 & 0.3104 & 0.3087 & 0.3130\\
        & & Avg  Act. rate (\%,$\downarrow$) & 9.76 & 9.96 & 8.27 & 7.66 & 11.50 & 13.34 & 7.13\\
        \hline\hline

\end{tabular}
\cprotect\caption{Completeness, causality and interpretability metrics (see Sections \ref{sec:existing_metrics} and \ref{sec:new_metrics}) of the concepts learned from different LLM classifiers for the dataset IMDB. Prior each task evaluation, all models are fine-tuned at the exception of Mistral-Instruct and Llama-Instruct, which are aligned with the task via soft‑prompt tuning. $\Delta f^{cond}$ is simply the averages of $\Delta f_{\{j\}}^{cond}$. $\verb|ConceptSim|$ is the average of the individual concept scores, weighted as detailed in the Appendix \ref{sec:interpretability_metrics}. All post-hoc methods were configured to search for 20 concepts. Concepts are computed from the sentence‑level hidden state: for decoder‑only models, from the residual stream after the penultimate transformer block and for encoder‑only models, from the layer preceding the classification head. Results are obtained with seed equals to $42$}
\label{tab:table_metrics_imdb}
\end{table*}

\end{document}